\documentclass[mnsc,nonblindrev]{ResGate} 

\OneAndAHalfSpacedXI 

\usepackage{natbib}
 \bibpunct[, ]{(}{)}{,}{a}{}{,}%
 \def\bibfont{\small}%
\usepackage{algorithm,algorithmicx}
\usepackage[noend]{algpseudocode}
\usepackage{xspace}
\usepackage{url,hyperref}
\usepackage{multirow, multicol, array}
\usepackage{graphicx}
\usepackage[flushleft]{threeparttable}
\usepackage{diagbox}
\usepackage{caption,subcaption}
\usepackage{tikz}

\TheoremsNumberedThrough     
\ECRepeatTheorems

\EquationsNumberedThrough    

\MANUSCRIPTNO{TBD}

\newtheorem{observation}{Observation}

\newcommand{\icml}{\color{black}}

\newcommand{\bitem}{\begin{itemize}}
\newcommand{\eitem}{\end{itemize}}
\newcommand{\benum}{\begin{enumerate}}
\newcommand{\eenum}{\end{enumerate}}
\newcommand{\bdefn}{\begin{definition}}
\newcommand{\edefn}{\end{definition}}
\newcommand{\bprop}{\begin{proposition}}
\newcommand{\eprop}{\end{proposition}}
\newcommand{\bque}{\begin{question}}
\newcommand{\eque}{\end{question}}
\newcommand{\bobsv}{\begin{observation}}
\newcommand{\eobsv}{\end{observation}}

\newcommand{\ps}{\begin{proof}[Sketch]}
\newcommand{\brmk}{\begin{remark}}
\newcommand{\ermk}{\end{remark}}
\newcommand{\bduiqi}{\begin{aligned}}
\newcommand{\eduiqi}{\end{aligned}}
\newcommand{\bcoro}{\begin{corollary}}
\newcommand{\ecoro}{\end{corollary}}
\newcommand{\bcom}{}

\newcommand{\asym}{asymptotic}

\newcommand{\asymly}{asymptotically}

\newcommand{\adap}{adaptive}

\newcommand{\apx}{approximate}

\newcommand{\avg}{average}
\newcommand{\arb}{arbitrary}
\newcommand{\arbly}{arbitrarily}
\newcommand{\alg}{algorithm}

\newcommand{\Alg}{Algorithm}

\newcommand{\assu}{assumption}
\newcommand{\Assu}{Assumption}

\newcommand{\bs}{\backslash}

\newcommand{\cond}{condition}

\newcommand{\constr}{constraint}

\newcommand{\ci}{confidence interval}

\newcommand{\distr}{distribution}

\newcommand{\emp}{empirical}

\newcommand{\elem}{element}

\newcommand{\func}{function}

\newcommand{\gs}{\gtrsim}

\newcommand{\ho}{\mathbb}

\newcommand{\indep}{independent}

\newcommand{\IOW}{In other words}

\newcommand{\imp}{impression}

\newcommand{\ins}{instance}

\newcommand{\ineq}{inequality}

\newcommand{\IFT}{It follows that}

\newcommand{\kehua}{characterize}

\newcommand{\ls}{\lesssim}
\newcommand{\lar}{\leftarrow}

\newcommand{\Ow}{Otherwise}
\newcommand{\ow}{otherwise}

\newcommand{\Omg}{\Omega}

\newcommand{\OTOH}{On the other hand}

\newcommand{\rb}{\right}

\newcommand{\lb}{\left}

\newcommand{\prb}{probability}

\newcommand{\parti}{particular}

\newcommand{\rv}{random variable}

\newcommand{\rand}{random}
\newcommand{\rar}{\rightarrow}

\newcommand{\real}{\mathbb{R}}
\newcommand{\resp}{respectively}

\newcommand{\sat}{satisfy}
\newcommand{\sats}{satisfies}

\newcommand{\sep}{separat}

\newcommand{\sln}{solution}
\newcommand{\sse}{\subseteq}
\newcommand{\sps}{suppose}
\newcommand{\Sps}{Suppose}

\newcommand{\strfwd}{straightforward}

\newcommand{\sym}{symmetry}

\newcommand{\unk}{unknown}

\let\eps\varepsilon

\begin{document}


\RUNAUTHOR{Jia et al.} 

\RUNTITLE{Short-lived High-volume Multi-A(rmed)/B(andits) Testing}

\TITLE{Short-lived High-volume Multi-A(rmed)/B(andits) Testing}

\ARTICLEAUTHORS{%
\AUTHOR{Su Jia}
\AFF{Cornell University, \EMAIL{sj693@cornell.edu}} 
\AUTHOR{Andrew Li, R. Ravi}
\AFF{Carnegie Mellon University, \EMAIL{\{aali1,ravi\}@andrew.cmu.edu}}
\AUTHOR{Nishant Oli, Paul Duff, Ian Anderson}
\AFF{Glance, \EMAIL{\{nishant.oli,paul.duff,ian.anderson\}@glance.com}}
} 

\ABSTRACT{Modern platforms leverage randomized experiments to make informed decisions from a given set of items  (``treatments'').
As a particularly challenging scenario, these items may (i) arrive in high {\it volume}, with thousands of new items being released per hour, and (ii) have short {\it lifetime}, say, due to the item's transient nature or underlying non-stationarity that impels the platform to perceive the same item as distinct copies over time. 
{\icml Thus motivated, we study a Bayesian multiple-play bandit problem that encapsulates the key features of the multivariate testing (or ``multi-A/B testing'') problem with a high volume of short-lived arms.
In each round, a set of $k$ arms arrive, each available for $w$ rounds. 
Without knowing the mean reward for each arm, the learner selects a multiset of $n$ arms and immediately observes their realized rewards.
We aim to minimize the {\em loss} due to not knowing the mean rewards, averaged over instances generated from a given prior distribution.
We show that when $k = O(n^\rho)$ for some constant $\rho>0$, our proposed policy has $\tilde O(n^{-\min \{\rho, \frac 12 (1+\frac 1w)^{-1}\}})$ loss on a sufficiently large class of prior distributions. 
We complement this result by showing that every policy suffers $\Omg (n^{-\min \{\rho, \frac 12\}})$ loss on the same class of distributions.
We further validate the effectiveness of our policy through a large-scale field experiment on {\em Glance}, a content-card-serving platform that faces exactly the above challenge.}
A simple variant of our policy outperforms the platform's current recommender by 4.32\% in total duration and 7.48\% in total number of click-throughs.
\footnote{A preliminary version of this work \citep{pmlr-v202-jia23b} appeared in the {\em Proceedings of the 40th International Conference on Machine Learning} (ICML'23). 
This work substantially expands the proceedings version by including (i) proof sketches for the results, (ii) new theoretical results based on discussions from the ICML presentation, (iii) additional analysis of the field experiment on {\em engaged users}, and (iv) offline simulations using real-world data.}}

\HISTORY{This manuscript was first submitted on October 13, 2023.}

\maketitle
\newpage
\section{Introduction}\label{sec:intro}

{\icml There has been a long history where online platforms leverage the scale of data to make personalized decisions about new items.
By and large, these tasks can be categorized along two dimensions: the lifetime and volume of the items. 
For {\em long-lived} items, the problem is straightforward: Collect a sufficient amount of data in the form of user feedback and then fit an offline predictive model, such as collaborative filtering or deep neural network (DNN).

Orthogonal to lifetime, the problem is similarly well understood when there is a {\em low volume} of items compared to the user population. 
Dedicated exploration methods, such as basic A/B testing (or ``A/B/n'' testing for multiple items), are sufficient to find the most favorable item.
Further insights can be found in \citealt{kohavi2017online} and \citealt{thomke2020experimentation}.

\begin{figure}
\centering
\includegraphics[width=0.9\linewidth]{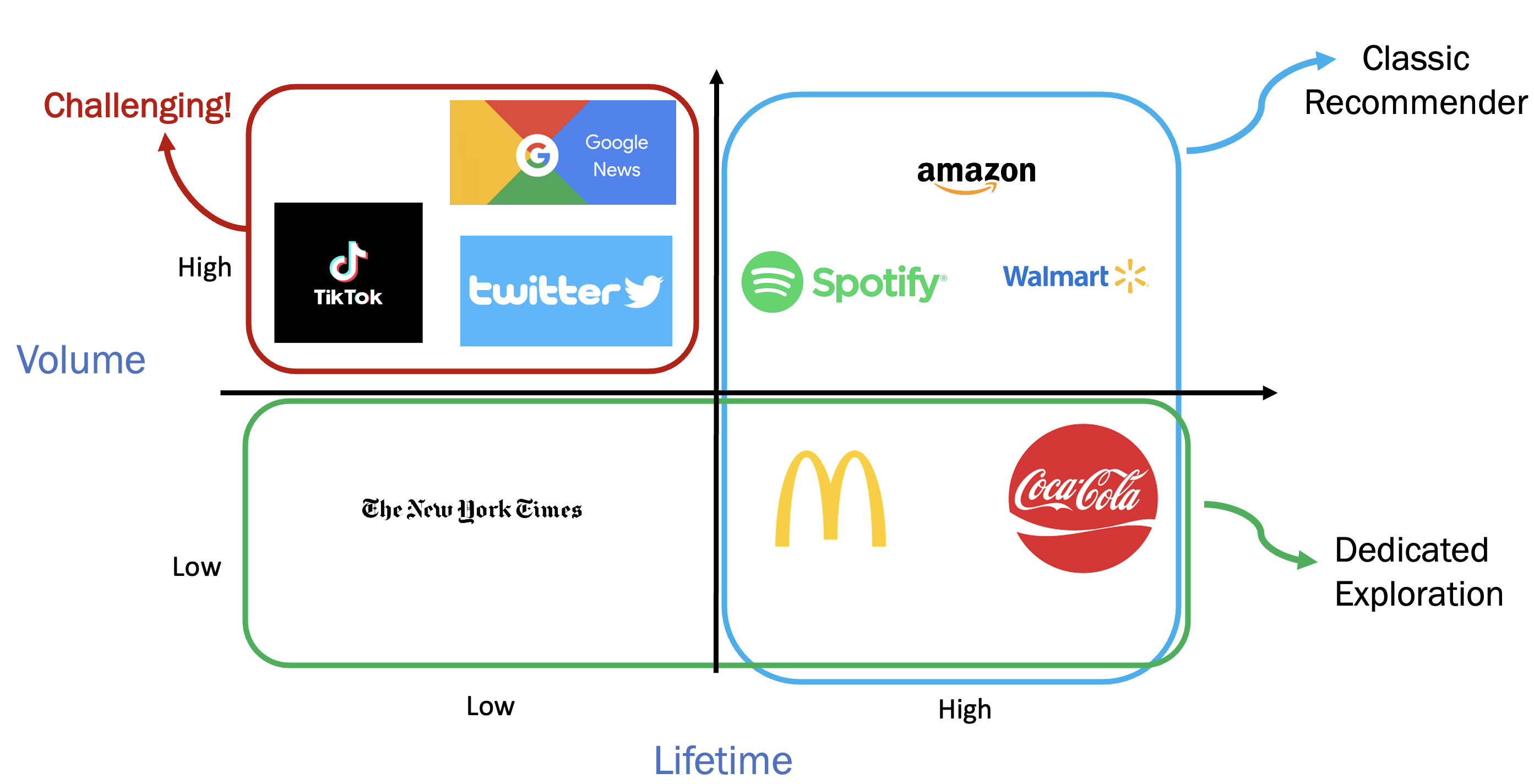}
\centering
\caption{The scenarios where the items (i) are long-lived (boxed in blue) or (ii) arrive in low volume (boxed in green) are relatively easy to handle.
This work centers on the most challenging scenario (boxed in red), that is, wherein the items are short-lived and arrive in high volume.}
\label{fig:quadrants}
\end{figure}

Naturally, then, the most challenging settings are where the contents are {\em short-lived} and arrive in {\em high volume}, as highlighted in the top-left quadrant in Figure \ref{fig:quadrants}.
An example is {\em ephemeral marketing}. 
This marketing strategy takes advantage of the fleeting nature of content on platforms such as social media, where posts and stories are visible for a limited time before they disappear.
Social media platforms like Instagram, Snapchat, and Facebook offer features like ``Stories'', which encompass temporary posts that typically disappear within a predetermined timeframe, usually less than 24 hours \citep{vuleta21}.
In addition, these contents often arrive in large quantities.
A central problem is determining the subset of content to appear at the top of user news feeds to optimize user engagement.

Another example is website optimization. 
In the realm of internet marketing, online platforms engage in {\em multi-variate} testing to evaluate different designs of their user interface, as exemplified by references such as \citealt{mcfarland2012experiment,yang2017framework}. 
As an example, LinkedIn conducts more than 400 concurrent experiments per day to compare different designs of their website. 
These experiments aim to encourage users to build their personal profiles or increase their subscriptions to LinkedIn Premium \citep{xu2015infrastructure}.
The number of items to be tested, specifically different website designs, can scale exponentially in the number of features considered, such as logos, fonts, background colors.
On the other hand, these items may be short-lived.
For instance, when accounting for non-stationarity arising from factors like seasonality or unexpected events, any prediction remains reliable for only a limited duration.
One approach is to partition the time horizon into shorter segments during which the environment can be approximately viewed as stationary. 
Each design is then treated as a distinct copy within these segments.

Intensifying this challenge is the aspect of {\em personalization}. 
A conventional approach to personalized decision making is cluster-then-optimize (see, e.g., \citealt{bernstein2019dynamic,kallus2020dynamic}): First, partition the users into clusters based on similarity information. Then, optimize for each cluster separately.
However, confining ourselves to a specific cluster leads to a notable reduction in the number of user impressions, thereby significantly curtailing the experimental resources at our disposal.

Driven by these considerations, we embark on a comprehensive study of multi-A/B testing in scenarios involving a high volume of short-lived items.
To this end, we introduce the {\em Short-lived High-volume Bandits} (SLHVB) problem which effectively encapsulates the key attributes of the challenge.
Our formulation encompasses the following fundamental elements.} 
\bitem 
\item {\bf High Volume of Short-lived Arms:} In each round, a set of $k=O(n^\rho)$ items (``arms'') arrives where $\rho>0$, and each remains {\em available} for a duration of $w$ rounds. 
The term ``high volume'' suggests the possibility that $k$ can scale {\em polynomially} in $n$. 
(Our framework {\bf does} allow $k$ to grow super-polynomially in $n$. But due to the Bayesian assumption, this scenario is not much different from the case where $k=\Theta(\sqrt n)$; see our discussion in Section \ref{subsec:choice_of_ell}.)
\item {\bf Multiple-play:} 
In each round, we choose a {\em multiset} of $n$ arms, with the possibility of selecting each action multiple times.
This emulates the procedure of, for example, assigning an advertisement to every one of the $n$ user impressions in each time period. 
Each time an arm is selected, an observable reward is generated independently.
The reward in the advertising scenario measures user engagement and is observable in interactions such as click-throughs.
\item {\bf Bayesian Formulation:} 
We assume that the mean reward for each arm is drawn from a known prior distribution. 
In contrast to the pessimistic worst-case version, this Bayesian framework more accurately aligns with the reality that the variability in the mean rewards often stems from the diverse quality of the arms, as opposed to being manipulated by a strategic agent.
We quantify the quality of a policy by the {\em loss} that results from not knowing the mean rewards.
\eitem 

Under the above model, we investigate the best achievable performance of a policy as $n$ tends to infinity, for various regimes of $\rho\in (0,\infty)$.
The asymptotics (i.e., ``big-O'') in this work are with respect to $n\rar \infty$, with $\rho$ fixed.

\subsection{Our Contributions} 
In light of this context, we make the following contributions to the field of {\em Multi-armed Bandits} (MAB) and {\em Online Controlled Experiments} (OCE).
\benum 
\item {\bf Formulation.} Our first contribution is formulating a Bayesian bandit problem that faithfully models a ubiquitous challenge faced by online platforms.
Unlike the worst-case approach, the Bayesian assumption not only better captures the real-world performance of a policy, but more importantly, facilitates a more intricate analysis of elimination-type policies, leading to stronger guarantees than those attainable in the worst-case scenario.
\item {\bf Nearly Optimal Policy.}
We show that an elimination-type policy achieves nearly optimal performance by integrating the following components.
\benum
\item {\bf Connection to Bayesian Batched Bandits.} Our analysis relies on drawing connections to the {\em Batched Bandits} (BB) problem where decisions are made in batches, rather than one at a time. 
We show that any algorithm for the BB problem with Bayesian regret $R(n,k)$ induces a policy for the SLHVB problem with $\tilde O\lb(n^{-\rho} + R(n,k)\rb)$ loss.
(To prevent confusion, we distinctly use the terms ``regret'' and ``loss'' for the BB and SLHVB problems, respectively.)
\item {\bf Bayesian Regret Analysis for the BB Problem.} 
\cite{gao2019batched} proposed an \alg, dubbed {\em Batched Successive Elimination} (BSE), which iteratively updates the confidence intervals for the mean rewards and eliminates arms that are unlikely to be optimal.
We show that for any integer $\ell\ge 1$, the level-$\ell$ BSE \alg\ has $\tilde O((k/n)^{\frac \ell{\ell+2}})$ Bayesian regret if $\rho \ge \frac{\ell-1}{2\ell+1}$.
This bound is stronger than the optimal $\tilde O((k/n)^{1/2}\cdot n^{2^{-\ell}})$ {\em worst-case} regret bound whenever $\ell\ge 2$.
For example, when $\rho=\frac 12$, our Bayesian regret bound becomes $n^{-1/2 + O(1/\ell)}$ whereas the worst-case regret bound is $\tilde \Theta(n^{-1/4})$.
\item {\bf Achieving Vanishing Loss.} Combining the Bayesian regret bound in (b) with the regret-to-loss conversion formula in (a), we obtain a policy for the SLHVB problem with $\tilde O(n^{-\min\{\rho, \frac w{2(w+1)}\}})$ loss by choosing a suitable $\ell$ depending on $\rho$ and $w$.
\item {\bf Near-Optimality.} Our \alg\ is nearly optimal: We show that any policy for the SLHVB problem suffers an $\Omg(n^{-\min\{\rho, \frac 12\}})$ loss. 
In \parti, the upper bound becomes arbitrarily close to this lower bound as $w\rar \infty$.
Furthermore, to highlight the value of the Bayesian assumption, we juxtapose this result with a lower bound on the {\em worst-case} loss which is \asymly\ higher.
\eenum
\item {\bf Experiments.}
We validated the effectiveness of our methodology through extensive online and offline experiments.
\benum 
\item {\bf Field Experiment.} We implemented a variant of our BSE policy in a large-scale field experiment via collaboration with {\em Glance}, a leading lock-screen content platform who faces exactly the aforementioned challenge. 
Their marketing team generates hundreds of {\it content cards} per hour, which are available for at most $48$ hours. 
In a field experiment, our policy outperformed the concurrent DNN-based recommender by a notable margin of 4.32\% and 7.48\% in terms of duration and number of click-throughs per user per day \resp.
\item {\bf Offline Simulations on Real Data.} To further validate the power of our approach, we implemented our policy using real data from the platform.
Our findings reveal that the performance of the BSE policy with three or four layers substantially outperforms the level-one and level-two versions when the number of users is insufficient compared to the number of actions.
\eenum
\eenum

\section{Literature Review}\label{sec:lit_rev}

We provide an overview of the relevant literature in the field of Multi-armed Bandits (in Section \ref{subsec:lit_rev_MAB}), Online Controlled Experiments (in Section \ref{subsec:lit_rev_OCE}) and field experiments on large-scale online platforms (in Section \ref{subsec:field_xp}).

\subsection{Multi-armed Bandits}\label{subsec:lit_rev_MAB}
{\icml Our problem is a variant of the {\it Multi-armed Bandits} (MAB) problem \citep{lai1985asymptotically}.
Three lines of work are most
related to ours: {\it multiple-play bandits}, {\it mortal bandits} and {\it high-volume bandits}. 

\noindent{\bf Multiple-play Bandits.} 
In this variant, several arms are selected in each round.
Many results in single-play bandits can be generalized to the multi-play variant, for example, \cite{komiyama2015optimal} showed that the \ins\ dependent regret bound for Thompson sampling can be generalized to the multi-play setting.
One motivation of the multi-play variant is {\it online ranking} (see, e.g. \citealt{radlinski2008learning, lagree2016multiple,gauthier2022unirank}) where the learner presents an {\it ordered} list of items to each user, viewed sequentially under certain click model.
Furthermore, there is no arrival of new arms, so the learner does not need to take into consideration the ages of the arms.

\noindent{\bf Mortality of Arms.} 
A quintessential motivation for the mortality of arms is online advertising. 
In the classical {\it pay-by-click model}, the ad broker matches each ad from a large corpus to content and is paid by the
advertiser (i.e., who created the ad) only when an ad is clicked. 
As a key feature, an ad becomes unavailable when the advertiser's budget runs out.
\cite{chakrabarti2008mortal} introduced the problem of {\it mortal bandits} and considered two death models. 
In the deterministic model, each arm dies after being selected for a certain number of times, which corresponds to the advertisers' budget in the advertising example. 
In the stochastic lifetime model, an arm dies with a fixed \prb\ every time it is selected.
Relatedly, in {\it rotting bandits} \citep{levine2017rotting}, each arm's mean reward decays in the number of times it has been selected. 
In \parti, if the reward \func\ is an indicator \func, then effectively each arm has a finite deterministic lifetime.
Motivated by demand learning in assortment planning, \cite{farias2011irrevocable} considered the {\it irrevocable bandits} problem that bears both the multi-play and mortality features: Arms are selected in batches and discarded immediately once selected. 
Unlike in our work, however, none of these models considered arrivals, and hence the learner does not need to account for the age of the arms.

\noindent{\bf High Volume of Arms.} Most existing work concerning large volume of arms considered the {\it worst-case} regret of a policy, i.e., the regret on the worst input in a given family (see, e.g., \citealt{berry1997bandit,zhang2021restless}).
As a distinctive feature, we consider a Bayesian model in which the mean rewards follow a known \distr.
As we will soon see, our formulation leads to theoretical results that would be \ow\ impossible.
\cite{wang2008algorithms} also assumed that the reward rates of the arms are \indep ly drawn from a common \distr\ such that the probability of being $\eps$-optimal is $O(\eps^\beta)$ where $\beta\in [0,1]$ is a known constant.
However, unlike in our problem, there are no arrivals and hence the policy does not need to balance the exploration for arms with different ages.

\noindent{\bf Low-Adaptive Bandits Algorithms.} 
This is another variant of MAB closely related to the multi-play bandits (and hence to this work). 
In the {\it batched bandits} problem \citep{perchet2016batched, agarwal2017learning} we aim to achieve low regret using low {\it adaptivity}. 
Unlike in the multi-play setting, here the batch size is part of the decision. 
The learner can partition the time horizon $[T]$ into a given number $w$ {\it batches}, and choose a batch (i.e., a multiset) of arms based on the realizations in the previous batches.
Alternatively, $w$ can be interpreted as a \constr\ on the adaptivity of the policy. 
In the classical setting, the learner has unlimited adaptivity, i.e., $w=T$.}
\cite{jun2016top} examined an even more general version of the problem, which allows an upper limit on the number of times an arm can be pulled in each batch.
The natural question then is: What regret can be achieved with small $w$?
{\icml \cite{gao2019batched} answered this question by showing that for any $k$ arms, we can achieve $\tilde O(\sqrt {kT})$ regret whenever $w=\Omg(\log \log T)$, which is optimal among all policies with {\em unlimited} adaptivity.}

\subsection{Online Controlled Experiments}
\label{subsec:lit_rev_OCE}
At the algorithmic level, our policy iteratively applies the effective new treatments to a larger population of users. 
This is related to {\em controlled rollout} in the literature on randomized controlled trials. 
Inspired by multi-phase clinical trials \citep{pocock2013clinical,friedman2015fundamentals}, many firms employ {\em controlled rollout} or {\em phased release} in A/B tests where they gradually increase traffic to the new treatment to 100\%. 
To balance speed, quality, and risk in a controlled rollout, \cite{xu2018sqr} embedded a statistical algorithm into the process of running every experiment to automatically recommend phase-release decisions.
\cite{xiong2019optimal} focused on the design of a policy that determines the {\em initial treatment time} for each unit, with the objective of obtaining an accurate estimate of the treatment effects. 
Differently from our work, the starting time of each treatment in this work is part of the decision rather than being given as input.
\cite{mao2021quantifying} developed a theoretical framework to quantify the value of iterative experimentation in a two-period, two-treatment model.
As the key distinction, these papers focus on characterizing the bias and variance of estimators, as opposed to cumulative regret or loss, which is the focus of this work.

\subsection{Large-Scale Field Experiment}
\label{subsec:field_xp}
Our work is related to the expanding body of work on field experiments on online platforms.
The reader can refer to the survey by \cite{terwiesch2020om}  for a comprehensive overview.
\cite{schwartz2017customer}  implemented a Thompson-sampling based policy for ad-allocation in a live field experiment in an online display campaign with a large retail bank.
\cite{cui2019learning} explored how consumers learn from inventory availability information on e-commerce platforms by creating exogenous, randomized shocks on inventory information on Amazon.
\cite{zhang2020long} studied short and long term 
price discounts on products from customers' shopping carts in retailing platforms.
In a field experiment, \cite{zeng2022impact} quantified the effect of {\em social nudges} in terms of boosting online content providers to create more content on a video-sharing social network.
\cite{feldman2022customer} conducted a field experiment on Alibaba to evaluate an assortment optimization approach based on the Multinomial Logit (MNL) model and observed a remarkable 28\% revenue advantage compared to the existing ML-based approach of the company.

\section{Formulation}\label{sec:formulation}

{\icml We now formally define the problem.
\Sps\ at the start of each round $t=1,2,\cdots$, a set $A_t$ of $k$ {\it actions} (or {\it arms}) arrives. 
Each action in this set has a {\it lifetime} $w$, i.e., it is available in rounds $t,\dots,t+w$. 
In other words, in round $t$, the learner is only allowed to select actions from the set $A_{t-w}^t$, where we define $A_t^{t'}:= \bigcup_{s=t}^{t'} A_s$ for any $0<t \leq t'$.

In each round, the learner selects a {\it multiset} of $n$ available arms, which means that each arm can be chosen multiple times, as long as the total number of plays is $n$.
If an arm $a$ is selected $m$ times, the learner receives observable rewards $X_1,\dots,X_m$. These rewards are identically independently distributed ({\em i.i.d.}) with a subgaussian {\em reward distribution}, whose mean we denote by $\mu_a$. 
For simplicity, we assume that these distributions have unit variance, although the analysis can be extended to encompass arbitrary subgaussian distributions in a straightforward manner.

We consider a Bayesian formulation in which the mean rewards $(\mu_a)$ are drawn i.i.d. from a {\em known} \distr\ $D$. 
Unlike the minimax framework, our Bayesian formulation more closely reflects real-world scenarios and, notably, empowers us to achieve theoretical guarantees that would otherwise be impossible; see Section \ref{sec:lb}.
In practice, $D$ can be estimated by leveraging historical data. 
For example, in the context of social media recommendations, we can estimate $D$ based on click-through rates of previous content.

We assume that the density function of $D$ is bounded from above and below away from $0$.

\begin{assumption}[Bounded Density Assumption]\label{assu:avg}
The distribution $D$ admits a density function $f$ that has a compact support $\mathcal{C}\sse \real$, and there exist constants $C_1, C_2 >0$ such that $C_1 \leq f(x) \leq C_2$ for all $x\in \mathcal{C}$. 
\end{assumption}

The above \assu\ is quite common in statistical learning; see, e.g., \citealt{ghosal2001convergence,petrone2002consistency} and \citealt{audibert2007fast}.
Without loss of generality ({\em w.l.o.g.}) we assume that $\mathcal{C}=[0,1]$.}

A {\it policy} is a procedure for selecting arms. Formally, it is specified by a sequence of decision rules $\pi=(\pi_t)$ where $\pi_t: A_{t-w}^t \times \prod_{s=1}^{t-1} (A_{s-w}^s \times \real)^n\rar [n]$ maps the ``history'' (i.e., past observations) up to time $t$ to a decision, that is, the number of times to select each available arm in the upcoming round $t$. 
(For completeness, we define $A_t=\emptyset$ for $t<0$.)
By abuse of notation, for each $t$ we denote by $\pi_t(a)$ the (random) number of times that an arm $a$ is selected in round $t$. 
We also use $\pi_t(S):=\sum_{a\in S} \pi_t(a)$ as a shorthand to denote the total count of arms selected from any set $S$ of arms.
Since $n$ actions are selected in each round, a valid policy should satisfy $\sum_{a\in A_{t-w}^t} \pi_t(a) = n$ for any $t$.

{\icml The problem would be simple if the mean rewards $(\mu_a)$ were known. 
In this scenario, we would choose the available arm that has the highest mean in each round. 
Formally, the optimal policy is given by $\pi^*_t(a)= n\cdot {\bf 1}[a= a^*_t]$ where $a_t^*\in \arg\max\{\mu_a: a\in A_{t-w}^t\}$.

When $(\mu_a)$ are unknown, we focus on evaluating a policy's {\em long-run} average loss (or simply {\em loss}) in comparison to the above benchmark.}
Given the intricacy of the formal definition, let us proceed with a step-by-step explanation. 
Fix an instance $\mu=\{\mu_a: a\in A_1^\infty\}$. 
Then, in each round $t$, the expected loss is $\ho{E}_\pi \lb[\sum_{a\in A_{t-w}^t} \pi_t(a) \cdot \lb(\mu^*_t - \mu_a \rb)\rb]$ where $\mu^*_t := \max\{\mu_a: a\in A_{t-w}^t\}$, where $\ho{E}_\pi$ denotes the expectation with respect to ({\em w.r.t.}) policy $\pi$. 
Thus, the average loss up to some time $T>0$ is \[\frac 1{nT} \cdot \ho{E}_\pi \lb[\sum_{t=1}^T \sum_{a\in A_{t-w}^t} \pi_t(a) \cdot \lb(\mu^*_t - \mu_a\rb)\rb].\] 
The long-run loss on the instance $(\mu_a)$ can then be defined by letting $T$ go to infinity. 
This is formally specified as  
\[{\varlimsup_{T\to \infty}} \frac 1{nT} \cdot \ho{E}_\pi \lb[\sum_{t=1}^T \sum_{a\in A_{t-w}^t} \pi_t(a) \cdot \lb(\mu^*_t - \mu_a\rb)\rb].\] 
{\icml Finally, we average the loss over instances $\mu$ drawn from $D$. 
Recall that $\sum_{a\in A_{t-w}^t} \pi_t(a) = n$ for any $t$, we define 
\[\mathrm{Loss}_n(\pi,D) := \ho{E}_{\mu\sim D} \lb[{\varlimsup_{T\to \infty}} \frac 1{nT} \cdot \ho{E}_\pi \lb[\sum_{t=1}^T \sum_{a\in A_{t-w}^t} \pi_t(a) \cdot \lb(\mu^*_t- \mu_a\rb)\rb] \rb].\]
Since the number of new arms in each round $k = O(n^{\rho})$, we are interested in characterizing how rapidly $\mathrm{Loss}_n(\pi)$ vanishes given a fixed $\rho>0$.}

\section{Upper Bounds}\label{sec:ub}

We establish the main upper bound by drawing connections with the {\em Batched Bandits} (BB) problem.
To avoid confusion, we use the term {\it algorithm} for the BB problem and {\it policy} for the SLHVB problem. 

This section is organized as follows. 
We will begin by illustrating how a {\em semi-adaptive} algorithm (to be defined shortly) for the BB problem can be transformed into a policy for the SLHVB problem.
We will also explain how performance guarantees can be translated from one problem to another.
Then, we examine the performance of the {\it Batched Successive Elimination} (BSE) algorithm and present a {\bf Bayesian} regret bound that is asymptotically stronger than the optimal {\bf worst-case} regret bound given in \citealt{gao2019batched}.
Finally, we will employ this result to formulate a policy for the SLHVB problem, achieving a loss of $\tilde O(n^{-\min \{\rho, \frac 12 \cdot (1+\frac 1w)^{-1}\}})$. 

{\icml
\subsection{Batched Bandits}
Introduced by \cite{perchet2016batched}, the Batched Bandits (BB) problem involves a learner making decisions in {\em batches} of rounds, rather than individual decisions at each round. 
Specifically, the learner is given $n$ {\it slots}, $k$ {\it arms} and an {\it adaptivity level} $\ell\ge 1$.
Each arm $a$ is associated with an unknown subgaussian distribution $D_a$ whose mean we denote by $\mu_a$.
Each time an arm $a$ is selected, the learner receives an observable reward drawn from $D_a$.
In each {\it phase} $i=0,1,\dots,\ell$, the learner selects a {\em multiset} $M_i$, called a {\em batch}.
If an arm is selected multiple times in one phase, the realized rewards may be different.
The batches in this context must have sizes that add up to a total of $n$, i.e., $\sum_{i=0}^\ell |M_i| = n$.
The goal is to maximize the expected total reward.

Given an \ins\ $(\mu_a)_{a\in [k]}$, the {\it regret} of an algorithm $\ho{A}$ for the BB problem is defined as 
\[\mathrm{Reg}_n(\ho{A};\mu) := \frac 1n \cdot \ho{E}\lb[\sum_{a\in [k]} (\mu^* - \mu_a) \cdot N_a\rb]\] 
where $\mu^* = \arg\max_{a\in [k]} \{\mu_a\}$ and $N_a$ is the number of times an arm $a$ is selected.
We want to emphasize that the regret mentioned above is scaled by $n$, so we should aim for $o(1)$ regret.

A reasonable performance metric, as used in many existing work for MAB, is the {\it worst-case regret} over all instances. 
Formally, it is given as $\max_{\mu\in [0,1]^k} \mathrm{Reg}_n(\ho{A};\mu).$
However, the loss objective in the SLHVB problem quantifies the {\em average} performance over all instances.
Therefore, to translate a result for the BB problem into a result for the SLHVB problem, we need to consider the {\em average} regret over all instances. 
This is captured by the {\em Bayesian} regret, which we formalize as follows.

\bdefn[Bayesian Regret]
Given a prior distribution $D$ over $[0,1]^k$, we define the {\it Bayesian regret} of a policy w.r.t. $D$ as 
\[\mathrm{BR}_n(\ho{A}, D) := \ho{E}_{\mu\sim D} \lb[\mathrm{Reg}_n(\ho{A};\mu)\rb].\]
\edefn

We will next explain the process of converting an \alg\ for the BB problem into a policy for the SLHVB problem.

\subsection{Semi-adaptive Algorithm}
An important category of algorithms is the class of {\em semi-adaptive} algorithms. 
In these algorithms, the size of each batch $M_i$ of selected arms is predetermined in a non-adaptive manner, while the arm selection may depend upon the rewards in the previous batches.

\bdefn[Semi-adaptive Algorithm]
Given an {\it adaptivity level} $\ell\ge 1$, a {\it semi-adaptive} algorithm is specified by \\
{\rm (i)} a {\em grid} $\eps_0,\eps_1,\dots,\eps_\ell \in (0,1)$ with $\sum_{i=0}^\ell \eps_i = 1$ (to avoid integrality issues, assume that $(\eps_i n)$ are integers), \\
{\rm (ii)} an initial decision rule $\ho{A}_0$ that decides the first batch of arms, and\\
{\rm (iii)} a sequence of decision rules $\ho{A}_j: ([k] \times \real)^{n_{j-1}} \rar [k]^{n_j - n_{j-1}}$ for $j=1,\dots,\ell$ where $n_j := \sum_{i=0}^j \eps_i n$.
\edefn}



It should also be noted that $(\eps_i)$ may depend on $n$. 
For example, the {\em Explore-then-Commit} (ETC) algorithm (see, e.g., Chapter 6 of \citealt{lattimore2020bandit}) can be viewed as a BB algorithm with $\ell=1$ under the grid $\eps_0=n^{-1/3}, \eps_1 = 1-\eps_0$.
More generally, a semi-adaptive algorithm can be viewed as a {\em multi-stage} ETC Algorithm. 
As another example, in the BSE policy (to be introduced shortly), we choose $\eps_0<\dots<\eps_{\ell-1}=o(1)$ and $\eps_\ell = \Omg(1)$.

\begin{algorithm}[h]
\begin{algorithmic}[1]
\State{Input: 
\bitem 
\item $\ho{A}$: a semi-adaptive algorithm for BB
\item $k'$: cardinality of the random subset of arms
\eitem}

\For{$t=1,2,\dots$}
\State{Receive a new set $A_t$ of arms}
\State{Sample a random subset $\tilde A_t$  of $A_t$ of size $k'$}
\Comment{Resampling step}
\State{$A_t \lar \tilde A_t$}
\For{$j=0,\dots, \ell$}
\State{Invoke $\ho{A}$ and receive for each $a\in A_{t-j}$ a number $N_a^t$}
\State\label{step:j}{Select each $a\in A_{t-j}$ for $N_a^t$ times, and feed the observed rewards to $\ho{A}$}
\EndFor
\EndFor
\caption{Induced Policy  $\pi[\ho{A};k']$ for SLHV Bandits}
\label{alg:induced}
\end{algorithmic}
\end{algorithm}

{\icml \subsection{The Induced Policy}
A semi-adaptive algorithm for the BB problem can be transformed into a policy (referred to as the {\em induced} policy) for the SLHVB problem as follows. 
Recall that $A_t$ denotes the set of arms that arrive in round $t$.
For each $j=0,1,\dots,\ell$, the induced policy designates $\eps_j n$ slots to execute the decisions made by the BB algorithm in the $j$-th batch, given $A_{t-j}$ as input.

It can be verified that the induced policy selects {\em exactly} $n$ arms in each round and is therefore valid.
To see this, denote by $N_a^t$ the number of times an arm $a \in A_{t-j}$ is selected in round $t$.
By the definition of $\eps_j$, we have $\sum_{a\in A_{t-j}} N_a^t = \eps_j n$. 
Summing over all phases, we obtain that
\[\sum_{j=0}^\ell\sum_{a\in A_{t-j}} N_a^t = \sum_{j=0}^\ell \eps_j n= n.\]
}

Slightly more general than the above description, we also have an additional {\em resampling} step, where we sample a random subset $\tilde A_t$ of $k'$ arms from the new arrivals and ignore all arms in $A_t \bs \tilde A_t$. 
This step is useful when $k$ is relatively large (more precisely $k=\Omg(\sqrt n)$, as we will soon see from the analysis). 
In this scenario, the available resources may not suffice to adequately explore all arms.
Therefore, rather than obtaining poor estimates for each arm,  we focus our efforts on a smaller subset to obtain more accurate estimates.
Unlike in the worst-case scenario, the resampling approach does not significantly impair the performance of the induced policy: Due to the Bayesian assumption, this subset is likely to contain an arm that is close to being optimal. 
The transformation is formalized in Algorithm \ref{alg:induced}. 

Note that in the induced policy, we only select arms with age at most $\ell$.
This is not practical when $\ell \le w$, as an obviously more effective policy involves exploiting the empirically best arm in $A_{t-w}^{t-\ell-1}$, rather than being restricted to $A_{t-\ell}^t$ alone. 
However, we present the policy in this particular manner because it simplifies the analysis and does not affect the asymptotics. 
In fact, by doing this, the loss increases only by $\tilde O(n^{-\rho})$, which has the same order as one of the lower bounds (Theorem \ref{thm:lb1} of Section \ref{sec:lb}) that we will present later.

{\icml \subsection{The Regret-to-Loss Conversion}
The {\em regret} of any semi-adaptive BB algorithm can be translated into the {\em loss} of the induced policy as follows.

\begin{proposition}[Regret-to-Loss Conversion]\label{prop:translation}
\Sps\ $\ho{A}$ is a semi-adaptive algorithm for the BB problem with $\ell\le w$ batches and has Bayesian regret $R(n,k)$ on any $k$-armed instance.
Let $D$ be a distribution \sat ing Assumption \ref{assu:avg}.
Then for any $k'\le k$, the induced policy $\pi[\ho{A}, k']$ for the SLHVB problem satisfies \[\mathrm{Loss}_n (\pi[\ho{A}, k'], D) = \tilde O\lb(\frac 1{k'} + \frac 1k + R(n,k')\rb).\]
\end{proposition}
}

Intuitively, the term $1/k'$ can be thought of as an estimate of the loss due to confining our choices to the subset of arms that has been resampled.
The second term, $1/k$, accounts for the variability in $\mu_{\max}(A_t)$ between different sets of arrivals. 
More precisely, it arises from the challenge of deciding how many times to choose each {\em newly} arriving arm in $A_t$, without knowing how $\mu_{\max}(A_t)$ compares with $\mu_{\max}(A_{t-w}^{t-1})$.
The final term, $R(n,k')$, corresponds to the Bayesian regret on the resampled instance, which consists of $k'$ arms. 
We formalize the above ideas in Section \ref{apdx:reg_to_loss}.

At this juncture, we have established a reduction of our problem to the BB problem.
Moving forward, in the next subsection, we will describe the specific semi-adaptive BB algorithm we intend to employ and present a novel Bayesian regret bound.

{\icml \subsection{The Batched Successive Elimination Algorithm}
In the preceding subsection we have seen that the ETC algorithm is a special algorithm for the BB problem with $\ell=1$. 
\cite{gao2019batched}
introduced the {\it Batched Successive Elimination} (BSE) Algorithm (formally stated in Algorithm \ref{alg:bse}) which extends the ETC algorithm by performing exploration {\em recursively}. 
The \alg\ explores in the first $(\ell-1)$ batches and then commits to the empirically best arm in the final batch. 
More precisely, in each phase $i=0,\dots,\ell-1$ the algorithm maintains a subset $S_i \sse [k]$ of {\it surviving} arms determined as follows: 
\bitem 
\item Initially $S_0 = [k]$.
\item In each phase $i=0,\dots,\ell-1$, select each arm in $S_i$ equally many times, i.e., $\eps_i n/|S_i|$ times.
\item In the final phase, i.e., phase $\ell$, select an arm from $S_\ell$ arbitrarily. 
\eitem
\cite{gao2019batched} considered the BSE algorithm with the following two grids.

\bdefn[Minimax and Geometric Grids]\label{defn:minimax_grid}
Fix any adaptivity level $\ell\ge 1$.
The {\em minimax} grid $\eps_{\cdot;\ell}^{\rm mm} := (\eps_{i;\ell}^{\rm mm})_{i=0}^\ell$ and {\em geometric grid} $\eps_{\cdot;\ell}^{\rm geo} :=(\eps_{i;\ell}^{\rm geo})_{i=0}^\ell$ are given recursively by 
\[\eps_{i;\ell}^{\rm mm} = a \sqrt{\eps_{i-1;\ell}^{\rm mm}} \quad \text{and} \quad \eps_{i;\ell}^{\rm geo} = b \cdot \eps^{\rm geo}_{i-1;\ell}\] 
\resp\ for $i=1,\dots,\ell$, with \[\eps_{0;\ell}^{\rm mm} = a = n^{\frac 1{2(1-2^{-\ell})}} \quad 
\text{and}\quad \eps^{\rm geo}_{0;\ell} = b = n^{1/\ell}.\]
\edefn    

\cite{gao2019batched} showed that the BSE algorithm under the above two grids achieves nearly optimal minimax regret and \ins-dependent regret, among all semi-adaptive algorithms (or ``algorithms with static grid'', in their terminology).
To formally state this result, we denote by $\cal F_\Delta\sse [0,1]^k$ the family of \ins s where the highest two mean rewards differ by some $\Delta>0$. 

\begin{theorem}[Theorem 1 and Theorem 2 of \citealt{gao2019batched}]
\label{thm:gao19}
For any adaptivity level $\ell\ge 1$, we have \begin{align}\label{eqn:minimax}
\max_{\mu\in [0,1]^k} \mathrm{Reg}_n ({\rm BSE}, \eps_{\cdot;\ell}^{\rm mm})
= \tilde O\lb(\sqrt{k/n} \cdot n^{2^{-\ell}}\rb)
\end{align}
and 
\begin{align}\label{eqn:ins_dep}
\max_{\mu\in\cal F_\Delta}
\mathrm{Reg}_n ({\rm BSE}, \eps_{\cdot;\ell}^{\rm geo}) = \tilde O\lb(\frac k\Delta \cdot n^{\frac 1\ell-1} \rb).
\end{align}
Furthermore, no semi-adaptive algorithm achieves $o(\sqrt{k/n} \cdot n^{2^{-\ell}})$ regret for all instances $\mu\in [0,1]^k$, or $o((k/\Delta) \cdot n^{\frac 1\ell-1})$ regret for all instances in $\cal F_\Delta$.
\end{theorem}

\begin{algorithm}[h]
\begin{algorithmic}[1]
\State Input: 
\bitem 
\item $\ell$: adaptivity level 
\item $\eps_0,\dots,\eps_{\ell-1}\in (0,1)$: exploration intensities
\item $n$: number of arms to be selected in total
\item $A$: a set of $k$ arms (computationally, an ``arm'' is an \indep\ sampling algorithm of a \distr) 
\eitem
\State\label{step:sampling} 
Let $\tilde A$ be a uniformly random subset of $A$ of size  $k'$\Comment{Resampling arms}
\For{$i = 0,1,...,\ell-1$}
\Comment{Phase $i$}
\If{$|S_i| = 1$} \State{
Set $S_\ell = S_i$; Break}
\Comment{Terminate early (we will show this is unlikely)}
\EndIf
\If{$|S_i| \ge 2$}
\State{$n_i \lar \lb\lfloor\frac{\eps_i n}{|S_i|}\rb\rfloor$}\Comment{Number of times to select each arm in $S_i$}
\For{$a\in S_i$}
\State{Select arm $a$ for $n_i$ times and observe rewards $X_{a,1},...,X_{a,n_i}$}
\State{$\overline X_a \lar \frac{1}{n_i} \sum_{j=1}^{n_i} X_{a,j}$}\Comment{Empirical mean}
\EndFor
\State{$\overline X_{\max} \lar \max\lb\{\overline X_a: a\in S_i\rb\}$}\Comment{Empirically maximal reward rate}
\State{$S_{i+1} = \{a\in S_i: |\overline X_a - \overline X_{\max}| \leq 3n_i^{-1/2} \log^{1/2} n \}$}\Comment{Update the surviving arms}
\EndIf
\EndFor
\State{Select any arm in $S_\ell$ for $n- \sum_{i=0}^{\ell-1} \lfloor \eps_i n \rfloor$ times}
\Comment{Last phase}
\caption{Batched Successive Elimination $\mathrm{BSE}(\eps_0,\dots,\eps_{\ell-1};k')$.\label{alg:bse}}
\end{algorithmic}
\end{algorithm}


As a surprising corollary, when $\ell=O(\log \log n)$, we have $n^{2^{-\ell}} = \tilde O(1)$ and hence eqn. \eqref{eqn:minimax} becomes $\tilde O(\sqrt {k/n})$, which matches the $\Omg(\sqrt{k/n})$ lower bound on the minimax regret for $\ell=n$, i.e., the {\em non}-batched bandits problem; see, e.g., Theorem 13.1 of \citealt{lattimore2020bandit}.}

In essence, Theorem \ref{thm:gao19} is established by analyzing the regret incurred by two types of arms. 
If an arm is eliminated before the last batch, then we can bound the total regret incurred by this arm as a function of its suboptimality. 
If an arm manages to withstand all elimination phases, then its mean reward must be close to optimal. 
As we shall see in the next subsection, our key distinction from their analysis is that we leverage the Bayesian \assu\ to {\em explicitly} bound the progress made (i.e., the number of arms remaining) after a certain number of elimination phases.

{\icml Since the minimax bound in eqn. \eqref{eqn:minimax} holds for all instances, we immediately obtain a {\em Bayesian} regret bound of $\tilde O(\sqrt{k/n} \cdot n^{2^{-\ell}})$, formally stated below.

\bcoro[A First Bayesian Regret Bound]\label{coro:BR_from_gao}
Let $D$ be any prior distribution \sat ing Assumption \ref{assu:avg}. 
Then, for any $\ell\ge 1$, we have
\[\mathrm{BR}_n \lb({\rm BSE} (\eps_{\cdot;\ell}^{\rm mm}, D) \rb)= \tilde O\lb(\sqrt{k/n} \cdot n^{2^{-\ell}}\rb).\]
\ecoro
}

Now consider eqn. \eqref{eqn:ins_dep}, the \ins\ dependent bound.
At first sight, it appears that {\em another} Bayesian regret bound should readily follow by taking expectation over $\Delta$. 
Interestingly, this approach does not work. Actually, it leads to an {\em infinite} regret bound!
This can be seen from the following.

\begin{proposition}[Eqn. \eqref{eqn:ins_dep} Leads to Infinite Bayesian Regret]
\label{lem:E1/Delta_is_infty}
\Sps\ $(X_i)_{i=1}^m$ are drawn i.i.d. from a distribution $D$ \sat ing Assumption \ref{assu:avg} where $m\ge 2$. 
Let $(X^{(i)})$ be the order statistics, i.e., $X^{(i)}$ is the $i$-th largest value, and write $\Delta = X^{(1)} - X^{(2)}$. 
Then, \[\ho{E}\lb[\frac 1\Delta\rb] = + \infty.\]  
\end{proposition}

To see this, take $D=U(0,1)$ and assume that $X^{(1)} = 1$. 
Then, we have 
\[\ho{P}\lb[\frac 1\Delta \ge y\rb] = 1- \ho{P}\lb[\Delta > \frac 1y\rb] \ge 1- \lb(1-\frac 1y\rb)^k = \frac ky + o\lb(\frac 1y\rb),\]
as $y\rar \infty$.
The claimed unboundedness then follows, since $\ho{E}[\Delta^{-1}] = \int_0^\infty \ho{P} [\Delta^{-1} \ge y]\ dy$.
This implies that instance-dependent regret bounds (e.g., Theorem 1 of \citealt{gao2019batched}) or sample complexity bounds (e.g., Theorem 1 of \citealt{agarwal2017learning}) cannot be straightforwardly applied to yield meaningful results in our setting.


So far we have a preliminary Bayesian regret bound. 
Next, we show how to improve the Bayesian regret bound using the structure of the prior.

\subsection{The Revised Geometric Grid}
{\icml The next two subsections are dedicated to explaining how the $\tilde O(\sqrt {k/n})$ Bayesian regret bound in Corollary \ref{coro:BR_from_gao} can be improved by leveraging the structure of the prior distribution $D$.
To state this result, we need a different type of geometric grid. 
Unlike in \citealt{gao2019batched}, the common ratio in our {\em revised} geometric grid incorporates {\bf both} $n$ and $k$ in order to balance the regret incurred in different elimination phases.
}

The definition of the revised geometric grid is naturally motivated by our regret analysis. 
It proceeds alternately between  finding an {\bf upper} bound on the number of surviving arms after $j\ge 1$ elimination phases, which in turn leads to a {\bf lower} bound on the number of times that each surviving arm is selected in the $(j+1)$-st phase.

To further illustrate this, consider $\ell=2$ and an arbitrary grid $(\eps_0,\eps_1)$. 
In the first phase, every arm is selected $\eps_0 n/k$ times, so we obtain a confidence interval for each of them of width $\delta_1 \sim  (\eps_0 n/k)^{-1/2}$.
By Assumption \ref{assu:avg}, the density of $D$ is bounded by $O(1)$, so the number $K_1$ of surviving arms is $O(\delta_1 k)$ with high \prb.
In the second phase, each surviving arm is selected $\eps_1 n/K_1$ times, and the new confidence intervals have length $\delta_2 \sim (\eps_1 n/K_1)^{-1/2}$.
Therefore, each arm that survived both elimination phases is suboptimal by only $O(\delta_2)$. 
Consequently, the regret can be bounded as 
\begin{align*}
\eps_0 \cdot 1 + \eps_1 \cdot \delta_1  + (1-\eps_0-\eps_1)\cdot \delta_2 &\ls \eps_0 + \eps_1 \lb(\frac {\eps_0 n}{k}\rb)^{-1/2} + \lb(\frac{\eps_1 n}{K_1}\rb)^{-1/2}\\
&\ls \eps_0 + \eps_1 \lb(\frac {\eps_0 n}{k}\rb)^{-1/2} + \lb(\frac {\eps_1 n}{\delta_1 k}\rb)^{-1/2} \\
&\sim \eps_0 + \eps_1 \lb(\frac {\eps_0 n}{k}\rb)^{-1/2} + \lb(\frac {\eps_1 n}{(\eps_0 n/k)^{-1/2}\cdot k}\rb)^{-1/2}.
\end{align*}

Now, it becomes evident what grid to choose: To minimize the above, we select $\eps_i$'s so that the three terms are on the same \asym\ order, i.e.,
\[\eps_0 \sim \lb(\frac kn\rb)^{1/2} \quad \text{and}\quad \eps_1 \sim \lb(\frac kn\rb)^{1/4}.\]
We emphasize that in the Bayesian setting, it is w.l.o.g. to assume that $k\le n$. 
In fact, as we recall from Proposition \ref{prop:translation}, when $k\ge n$, we may narrow our focus to a randomly sampled subset of $k'$ arms, which incurs only an $O(1/k')$ increase in regret.
Consequently, we will proceed with the assumption that $k<n$ in our subsequent analysis.

We can generalize the above argument to any $\ell\ge 1$, which motivates the following grid.

{\icml \bdefn[Revised Geometric Grid]
\label{def:geometric_grid}
For any adaptivity level $\ell\ge 1$, we define the {\em revised geometric grid} as $\eps^\star_{\cdot;\ell} = (\eps^\star_{i;\ell})_{0\le i \le\ell}$ where 
\[\eps_{i,\ell}^\star := \lb(\frac kn\rb)^{\frac{\ell-i}{\ell+2}}, \quad  \text{for any}\ i=0,1,\dots,\ell-1, \quad \text{and}\quad \eps_{\ell,\ell}^\star := 1-\sum_{i=0}^{\ell-1} \eps_{i,\ell}^\star.\] 
\edefn
With the above definition in mind, in the next subsection we formally state our Bayesian regret bound and discuss its implications.}

\subsection{Breaking the $\tilde O(\sqrt {k/n})$  Bayesian regret}
The meticulous reader may have noticed a limitation in the analysis for $\ell=2$ in the previous section.
We argued that the number of arms that survive the first elimination phase is approximately $(\eps_0 n/ k)^{-1/2} \cdot k$, but this quantity could be less than $1$, which is not logically valid.
In other words, we implicitly assumed that $(\eps_0 n/ k)^{-1/2} \cdot k > 1$. 

{\icml Here is a more concrete example. Suppose $\ell=2$, $\rho=0.04$ and consider a uniform prior $U(0,1)$.
By Definition \ref{def:geometric_grid}, in the revised geometric grid, we have $\eps^\star_{0;2} = \tilde O(\sqrt{k/n})$, so each arm is selected around $\eps_{0;2}^\star \cdot n /k = (n/k)^{1/2} = n^{0.48}$ times immediately upon arrival. 
Thus, for each arm we have a confidence interval of width around $(n^{0.48})^{-1/2} = n^{-0.24}$.
On the other hand, in a ``typical'' instance drawn from $U(0,1)$, the mean rewards are spaced at $n^{-\rho}=n^{-0.04}$ distance apart on average, which is much {\em greater} than $n^{-0.24}$. 
Therefore, an optimal arm is likely to be identified after only one elimination phase, and there is no need for a second phase.

In general, if the number of arrivals is small, i.e., $\rho$ is small, there is no need to have an excessive number of elimination phases.
To avoid this degeneracy, our result requires that $\rho$ exceed the following {\em threshold exponent}. 

\begin{definition}[Threshold Exponent]\label{def:thr_expo}
For each integer $\ell\ge 1$, we define the {\it threshold exponent} as $\theta_\ell := \frac{\ell-1}{2\ell+1}$.
\end{definition}

We are now ready to state our main Bayesian regret bound.
We will show in Appendix \ref{apdx:gen_ell} that if $\rho\ge \theta_\ell$, then the above degeneracy is unlikely and hence the analysis remains valid.
For convenience, we denote by $\rm{BSE}_\ell^\star$ the $\mathrm{BSE}_\ell$ algorithm with the revised geometric grid $\eps_{\cdot;\ell}^\star$. 

\begin{theorem}[Bayesian regret of ${\rm BSE}^\star_\ell$]
\label{prop:ub_gen_ell}
\Sps\ the prior distribution $D$ satisfies Assumption \ref{assu:avg}.
Then, for any adaptivity level $\ell\ge 1$ with $\rho\ge \theta_\ell$, we have 
\[\mathrm{BR}_n \lb(\mathrm{BSE}_\ell^\star, D\rb) = \tilde O\lb(\lb(\frac kn\rb)^{\ell/(\ell+2)}\rb).\]
\end{theorem}
It should be noted that the bound in Theorem \ref{prop:ub_gen_ell} decreases in $\ell$, but $\ell$ can not be \arbly\ large since $\theta_\ell \le \rho$. 
Given any $k=n^\rho<n$, the maximum feasible $\ell$ can be determined as follows:\\
(i) If $\rho< \frac 12$, choose the maximum $\ell$ with $\rho \ge \theta_\ell$, i.e., $\ell = \ell^*(\rho) := \lb\lfloor \frac {1+\rho}{1-2\rho} \rb\rfloor$.\\
(ii) If $\rho \ge \frac 12$, then the \cond\ $\rho \ge \theta_\ell$ holds for any $\ell$, and so we can choose \arbly\ large $\ell$.

We summarize the above as the following corollary.

\begin{corollary}[Bayesian Regret, Explicit Form]
\label{coro:explicit} \Sps\ the prior $D$ satisfies Assumption \ref{assu:avg}. Then, the following holds under the revised geometric grid $(\eps_{\cdot;\ell}^\star)$ given in Definition \ref{def:geometric_grid}: \\
{\rm (i)} If $0<\rho <\frac 12$, then for the adaptivity level $\ell^*(\rho) = \lb\lfloor \frac {1+\rho}{1-2\rho} \rb\rfloor$, we have
\[\mathrm{BR}_n \lb(\mathrm{BSE}_{\ell^*(\rho)}^\star, D\rb) = \tilde O\lb(\lb(\frac kn\rb)^{\frac{\ell^*(\rho)}{\ell^*(\rho)+2}}\rb).\]
{\rm (ii)} If $\rho\ge \frac 12$, then for any adaptivity level $\ell\ge 1$, we have 
\[\mathrm{BR}_n \lb(\mathrm{BSE}_\ell^\star, D\rb) = \tilde O\lb (\lb(\frac kn\rb)^{\frac \ell{\ell+2}}\rb).\] 
In \parti, as $\ell\rar \infty$ we have 
\[\mathrm{BR}_n \lb(\mathrm{BSE}_\ell^\star, D\rb) = \lb(\frac kn\rb)^{1-O\lb(\frac 1\ell\rb)}.\]
\end{corollary}
Note that since $n^{2^{-\ell}} > 1$ for any $\ell$, the $\tilde O(\sqrt{k/n} \cdot n^{2^{-\ell}})$ bound in Theorem \ref{thm:gao19} is \asymly\ {\em strictly} higher than $(k/n)^{1/2}$.
In contrast, whenever $\rho \ge \frac 15$, we have $\ell^*(\rho) \ge 2$, and thus the bound in Corollary \ref{coro:explicit} is $\tilde O(\sqrt{k/n})$, which is strictly lower.
This contrast becomes sharper as $\rho$ increases. 
For example, with $\rho=\frac 13$, we have $\ell^*(\rho) = 4$, and hence 
\[{\rm BR}_n \lb(\mathrm{BSE}^\star_{\ell^*(\rho)}, D\rb) = \tilde O\lb(\lb(\frac kn\rb)^{2/3}\rb),\] 
which is \asymly\ stronger than $(k/n)^{1/2}$. 
(Again, we emphasize that we may assume that $k<n$, because otherwise we may restrict ourselves to a random subset of $k'<n$ arms.)
In the extreme case, for $\eps>0$ and $\rho =\frac 12 - \eps$, we have $\ell^*(\rho) = \Omg(1/\eps)$ and hence 
\[{\rm BR}_n \lb(\mathrm{BSE}^\star_{\ell^*(\rho)}, D\rb) = \lb(\frac kn\rb)^{1-O(\eps)},\] 
which is \asymly\ much stronger than $(k/n)^{1/2}$.

So far we have established a Bayesian regret bound of the BSE Algorithm.
Next, we explain how to choose $\ell\le w$ to obtain the best bound on the loss of the induced policy for the SLHVB problem.

\subsection{Loss of the Induced Policy}\label{subsec:choice_of_ell}
As the main result of this section, we present an upper bound on the loss of the induced policy for the SLHVB problem.
This result follows by combining (i) the Bayesian regret bound of the BSE algorithm in Proposition \ref{prop:ub_gen_ell} and (ii) the regret-to-loss conversion formula in Proposition \ref{prop:translation}. 

To formally state this result, we first recall some notation. 
Given a semi-adaptive algorithm $\ho{A}$ for the BB problem and an integer $k'$, we denote by $\pi[\ho{A}, k']$ the induced policy for the SLHVB problem with resampling size $k'$ given in Algorithm~\ref{alg:induced}. 
Also, let ${\rm BSE}^\star\lb(\ell, k\rb) \equiv {\rm BSE}_\ell^\star$ denote the BSE algorithm with the revised geometric grid $\eps^\star_{\cdot;\ell}$ as specified in Definition \ref{def:geometric_grid}.
We first state a loss bound that holds for any number $\ell\le w$ of batches. 


\begin{proposition}[Loss of the Induced Policy, Generic Version]
\label{prop:loss_induced}
\Sps\ the prior distribution $D$ satisfies Assumption \ref{assu:avg}.
Then, for any $\ell\le w$, $k'\le k$ and $\rho\ge \theta_\ell$, we have 
\begin{align}\label{eqn:loss}
\mathrm{Loss}_n \lb({\rm BSE}^\star\lb(\ell, k'\rb), D\rb) 
= \tilde O\lb(\frac 1{k'} + n^{-\rho}+ n^{(\rho-1)\cdot \frac\ell{\ell+2}} \rb) = \begin{cases}
\tilde O\lb(\frac 1{k'} + n^{-\rho}\rb), &\text{if } \rho < \frac \ell{2(\ell+1)},\\
\tilde O\lb(\frac 1{k'} + n^{(\rho-1)\cdot \frac\ell{\ell+2}}\rb), &\text{\ow.}
\end{cases}
\end{align}
\end{proposition}

Note that the above is a generic loss that holds for any number $\ell\le w$ of phases. 
Next, we discuss how to select $\ell$.
Since the exponent $\ell/(\ell+2)$ increases in $\ell$, at first sight, one is tempted to choose $\ell=w$.
However, the loss bound in Proposition \ref{prop:ub_gen_ell} requires $\rho \ge\theta_\ell = \frac{\ell-1}{2\ell+1}$.
This motivates us to select the maximum $\ell$ that satisfies $\theta_\ell \le \rho$ and $\ell \le w$ at the same time. 
Our {\em Hybrid policy} explicitly specifies this choice for any given $\rho$, as detailed in Algorithm \ref{alg:hybrid}. 

\begin{algorithm}[h]
\begin{algorithmic}[1]
\State Input: $\rho, w$
\If{$\rho < \frac 15$} \label{line:15} set $\ell=1$ and $k'=k$.
\EndIf
\If{$\frac 15 \le \rho< \frac w{2w+2}$} \label{step:med} 
set $k'=k$ and choose $\ell\leq w$ such that $\theta_\ell = \frac{\ell-1}{2\ell+1} \le \rho < \frac \ell {2\ell+2}$.
\EndIf
\If{$\rho\ge \frac w{2w+2}$}\label{step:k'}
set $\ell=w$ and $k' = n^{\frac w{2w+2}}$
\EndIf
\State{Invoke the policy ${\rm BSE}^\star(\ell, k')$.}
\caption{Hybrid Policy $\ho{H}(\rho,w)$}
\label{alg:hybrid}
\end{algorithmic}
\end{algorithm}

\begin{theorem}[Loss of Hybrid Policy]\label{thm:ub}
For any SLHVB \ins\ with volume exponent $\rho>0$, lifetime $w\ge 1$, and prior distribution $D$ that satisfies Assumption \ref{assu:avg}, we have \[\mathrm{Loss}_n \lb(\ho{H}\lb(\rho;w\rb), D\rb) = \tilde O\lb(n^{-\min\lb\{\rho, \frac 12 (1+ \frac 1w)^{-1}\rb\}}\rb).\]
\end{theorem} 

The proof of Theorem \ref{thm:ub} is \strfwd. 
For each of the three regimes outlined in Algorithm \ref{alg:hybrid}, we substitute the choice of $\ell$ and $k'$ into eqn. \eqref{eqn:loss} in Theorem \ref{prop:ub_gen_ell}. 
Then, we simplify the loss bound by suppressing the terms that are \asymly\ lower. 

To better understand how the lifetime $w$ affects the loss bound, let us compare the above bounds for $w=1$ and $2$. 
The regret bounds are \asymly\ $r_1(n):= n^{-\min\{\rho, 1/4\}}$ and $r_2(n):= n^{-\min\{\rho, 1/3\}}$.
In \parti, when $\rho > \frac 14$, the loss is $r_2= \Theta(n^{-1/3})$ for $w=2$, which vanishes faster than $r_1 =\Theta( n^{-1/4})$.
}

\section{Lower Bounds}\label{sec:lb}

{\icml In this section, we show that the upper bound in Theorem \ref{thm:ub} is nearly optimal.
Specifically, we prove that every policy suffers $\Omg(n^{-\min\lb\{\rho, \frac 12\rb\}})$ Bayesian regret, if the prior \distr\ \sats\ Assumption \ref{assu:avg}.
This result follows immediately by combining {\em two} lower bounds, stated in Propositions \ref{thm:lb1} and \ref{thm:lb2} \resp.

Our first lower bound, stated below, is essentially due to not knowing whether one of the newly arriving arms is significantly better than the older available arms.

\begin{proposition}[Lower Bound I]\label{thm:lb1}
\Sps\ the prior \distr\ $D$ satisfies Assumption \ref{assu:avg}. 
Then for any $\rho\geq 0$ and policy $\pi$, it holds that
\[\mathrm{Loss}_n(\pi, D) \geq \frac 1{12w^2}\cdot n^{-\rho}.\]
\end{proposition}
}

We establish this by showing that an $\Omg(n^{-\rho})$ loss is incurred on average in each round, regardless of whether the policy adequately explores the newly-arriving arms. 
Formally, we consider the loss \[L_t = \frac 1n \cdot \sum_{a\in A_{t-w}^t} \pi_t(a)\cdot (\mu^*_t - \mu_a)\] in round $t$ and  show that $\ho{E}[L_t] = \Omg(n^{-\rho})$.

We will use the following key observation: 
Since there are $wk$ arms available, by \Assu\ \ref{assu:avg}, the gap in mean rewards between the best two available arms is \apx ly $\frac 1{wk}$.
We will use this insight to lower bound the loss in two distinct cases.

{\bf Case A:} \Sps\ $\pi_t(A_t) \ge \frac n2$. Consider the event that $A_t$ does not contain the optimal available arm, i.e., $a^*_t \in A_{t-w}^{t-1}$.
This event occurs with \prb\ $1-\frac 1w$. 
In fact, by \sym, $a^*_t$ is contained in each of the $w$ age groups with equal likelihood.
When this event occurs, we have 
\[\mu(a^*_t) - \mu_{\max}(A_t) = \mu_{\max}(A_{t-w}^{t-1}) - \mu_{\max}(A_t) \sim \lb(1-\frac 1{wk}\rb) - \lb(1-\frac 1k\rb) \gs \lb(1-\frac 1w\rb) \frac 1k.\]
Since $\pi_t (A_t) \ge \frac n2$, we have 
\[L_t\gs \frac 1n \cdot \frac n2 \lb(1-\frac 1w\rb) \cdot\frac 1k =\frac 1{wk}.\]
 
{\bf Case B:} \Sps\ $\pi_t(A_t) < \frac n2$, or equivalently, $\pi_t(A_{t-w}^{t-1})\ge \frac n2$. 
Consider the event that $a^*_t\in A_t$, which occurs with \prb\ $\frac 1w$. 
By Assumption \ref{assu:avg}, conditioning on this event, we have \[\mu(a^*_t) - \mu_{\max}\lb(A^{t-1}_{t-w}\rb) = \mu_{\max}(A_t) - \mu_{\max}\lb(A^{t-1}_{t-w}\rb) \sim \lb(1-\frac 1{wk}\rb) - \lb(1-\frac 2{wk}\rb) =\frac 1{wk}.\]
Since $\pi_t (A_t) \ge \frac n2$, we have 
\[L_t \gs \frac 1n\cdot \frac n2 \cdot \frac 1w \cdot \frac 1{w k} = \Omg(w^{-2}\cdot n^{-\rho}).\]

{\icml 
However, this $\Omg(n^{-\rho})$ bound becomes very weak when $\rho$ is large.
To bridge this gap, we next complement this result with a lower bound that is tailored to the domain where $\rho\ge \frac 12$. 

\begin{proposition}[Lower Bound II]\label{thm:lb2} 
If $\rho\geq \frac 12$, then for any policy $\pi$ and prior \distr\ $D$ satisfying Assumption \ref{assu:avg}, we have \[\mathrm{Loss}_n (\pi,D) \geq \frac 1{96 w^2}\cdot n^{-1/2}.\]
\end{proposition}
}

The idea works as follows. 
For any $\eps>0$, an arm is called {\em $\eps$-good} if it has mean reward $1-O(\eps)$. 
\Sps\ at round $t$, all $n^{-1/2}$-good arms expire.
Then, to prevent a loss of $n^{-1/2}$ per round in the next $w$ rounds, a policy needs to identify an $n^{-1/2}$-good arm.
To achieve this, by Assumption \ref{assu:avg}, the policy must explore $\Omg(\sqrt n)$ {\em distinct} arms in expectation in the upcoming $w$ rounds. 
Note that every time we select a ``fresh'' arm (i.e., an arm that has never been selected), an $\Omg(1)$ loss is incurred (in expectation) due to Assumption \ref{assu:avg}. 
Therefore, in the next $w$ rounds, the policy suffers a loss of $\frac 1n \cdot \Omg(\sqrt n)\cdot \Omg(1)=\Omg(n^{-1/2})$.

{\icml Note that when $\rho = \frac 12$, the two lower bounds above have the same asymptotic order. We immediately obtain the following result by combining them.

\begin{theorem}[Combined Lower Bound]\label{thm:main_lb}
For any $\rho>0$, policy $\pi$ and prior \distr\ $D$ satisfying Assumption \ref{assu:avg}, we have \[\mathrm{Loss}_n(\pi; D) \ge \frac 1{96 w^2}\cdot n^{-\min\lb\{\rho, \frac 12\rb\}}.\]
\end{theorem} 

The above lower bound no longer holds under the {\em worst-case} loss, which highlights the advantages of our Bayesian formulation.
Formally, consider the worst-case loss 
\[\mathrm{Loss}^{\rm wc}_n(\pi) := \varlimsup_{T\rar \infty} \max_{\mu\in [0,1]^{kT}} \ho{E}_\pi \lb[\frac 1{nT} 
\sum_{t=1}^T \sum_{a\in A_{t-w}^t} \pi_t(a) \cdot \lb(\mu^*_t- \mu_a\rb)\rb].\]
Here, the only distinction from the Bayesian regret lies in replacing ``$\ho{E}_{\mu\sim D}$''
with ``$\max_{\mu\in [0,1]^{kT}}$''.
We show that no policy achieves $o(1)$ worst-case loss.}

\begin{proposition}[Worst-case Loss]\label{prop:wc_lb} For any policy $\pi$ and lifetime $w>0$, we have 
\[\mathrm{Loss}^{\rm wc}_n (\pi) \ge \frac 1{2w.}\]
\end{proposition}

To see this, consider $k=1$ (note that a lower bound for any $k$ also holds for larger $k$). 
Consider the following two instances (A) and (B). 
In both instances, the arm that arrives in the first round has mean rewards $\frac 12$.
In the second round, the arriving arm has a mean reward $1$ in (A), and mean rewards $0$ in (B).

For these instances, any policy would suffer a high loss since it is unable to learn the true instance based on observations from the first round.
More precisely, consider the event $\cal E$ that $\pi_2(A_2) \ge n/2$.
Note that this event relies on the observations in the first round and, therefore, it carries the same probabilities under both instances.
It is then \strfwd\ to verify that on the event $\cal E$, the policy has $\Omg(1/k)$ loss under (B); and on the event $\overline{\cal E}$, the policy has $\Omg(1/k)$ loss under (A).

Until this point, we have introduced a policy and demonstrated its theoretical near-optimal performance.
In the remainder of the paper, we proceed to validate the practical effectiveness of our policy through a field experiment and an offline simulation using real data.

\section{Field Experiment Design}\label{sec:field_xp_summary}

{\icml We validated the effectiveness of our policy in a field experiment, through collaboration with {\em Glance}, a leading lockscreen content platform that faces the above challenge.
Specifically, the firm's marketing team curates around 200 {\it content cards} (or simply, {\it cards}) per hour.
Of these cards, around $70\%$ become obsolete within $48$ hours.
Each card consists of a link to external content (e.g., video, or news article), along with a short text description.
The primary objective of the company is to provide users with card recommendations that optimize overall user engagement. 
Engagement is measured by both the total {\it duration} of user interactions and the number of {\em click-throughs} (CTs) to the content.

This problem can be cast as an SLHVB problem. 
Here, the cards correspond to the arms, the conversions correspond to the rewards, and the number $n$ corresponds to the number of impressions per round, which we choose to be an hour.
Two primary metrics are especially significant for each card: (i) the {\it click-through rate} (CTR) and (ii) the {\it \avg\ duration} per impression.
These metrics are initially unknown when a card is introduced. 

To combine these metrics, we introduce the concept of {\em conversions} in consultation with the business and data science teams.
A conversion occurs if either (i) a CT occurs or (ii) the duration exceeds a threshold of $\theta=5$ seconds. 
For simplicity, we assume that the rewards across different impressions are i.i.d. \rv s.

The platform sends cards to users on an hourly basis.
For each individual user, the platform updates the cards they have previously seen with fresh content cards if their device is connected to the internet.
Users can swipe through the cards stored in the platform's App.
When a user is interested in the content, they can click on the provided link, be redirected to an external source for further engagement, and then be redirected back to the App when finished.
To decide which cards to send, the firm deployed a recommendation system based on a state-of-the-art Deep Neural Network (DNN); see
\citealt{olipersonalizing}.
This DNN predicts, for each user-card pair, the expected conversion probability using the user's interaction history and the card's text and image features.

Although the current recommender works reasonably well, there is considerable potential for improvement.
Most notably, it solely employs user feedback to update the users' behavioral signature for future predictions, and does not harness this feedback to make recommendations to users with similar preferences.
In \parti, it does not use feedback to {\em directly} adjust the conversion rate predictions.
This may have caused a substantial loss in user engagement.
It is thus vital for the platform to find a recommendation policy that (i) can learn the true conversion rates of new cards quickly based on user interaction data, and (ii) is computationally simple to deploy. 

Our policy is well-suited for this task.
We implemented a Bayesian variant of the level-$1$ BSE-induced policy, employing a Beta-Bernoulli reward model.
Specifically, we use the DNN prediction to fit a Beta prior distribution for the mean reward of each arm.
Then, at the end of each round (set to one hour),
the posterior reward distribution for each arm is updated independently. 
This update is performed efficiently, benefiting from the fact that the Beta distributions constitute a conjugate family for the Bernoulli distribution.

To conduct online experiments, the company had organized its users into $100$ {\em buckets} for various online experiments. 
For our study, we integrated a randomized variant of the level-$1$ BSE-induced policy into their live system during the first 14 days of July 2021.
We selected three buckets to form the treatment group, consisting of 514,111 users and 17,984,977 impressions.  
This accounted for \apx ly 1\% of the total traffic during the 14-day duration when the experiment was conducted.
To facilitate comparisons, we also examined interaction data for the first 14 days of May in the same year.
We defer the implementation details to Appendix \ref{apdx:details_field_xp}.


{\icml Through an offline simulation, we determined that the empirically optimal parameter is approximately $\eps_0 = 0.20$, which we applied in the field experiment. 
This choice aligns well with our theoretical analysis. 
In fact, as we recall from Proposition \ref{prop:ub_gen_ell}, the optimal parameter is $\eps^\star_{0;1} \sim (k/n)^{1/3}$. 
In our scenario, based on past data, we observed an average of around $11$ million impressions every $14$ days. 
Therefore, the number $n$ of impressions per hour is approximately $3.2 \times 10^4$. 
Furthermore, considering an average release rate of 150 cards per hour, the theoretically optimal parameter is estimated to be $\eps_{0,1} \approx 0.14$, which closely matches the empirically optimal choice.

Finally, we emphasize that in our implementation, our BSE-based policy is {\em not} personalized. 
Despite the disadvantage, our policy still surpasses their {\em personalized} DNN recommender in all metrics of user engagement, both at the per-impression and per-user levels, as illustrated in the next section.
This underscores the effectiveness of our approach and sets the stage for further improvements in the future.}

\section{Field Experiment Results}\label{sec:field_xp_analysis}
We now provide a statistical analysis of the results of the field experiment, including (i) a basic hypothesis test, (ii) a bootstrapping hypothesis test, and (iii) a {\em difference-in-differences} (DID) regression analysis. 
Our policy is more effective than the firm's concurrent DNN-based recommender by 4.32\% in total duration and by 7.48\% in the total number of CTs per user per day.

\begin{table}[]
\caption{Four types of data for our analysis. 
}
\centering
\begin{tabular}{lll}
& \multicolumn{1}{c}{Click-through} & \multicolumn{1}{c}{Duration} \\ \hline
Per user per day   & \multicolumn{1}{c}{integral} & \multicolumn{1}{c}{numeric}  \\ \hline
Per impression & \multicolumn{1}{c}{binary} & \multicolumn{1}{c}{numeric} \\ \hline
\end{tabular}
\label{tab:data_types}
\end{table}

\begin{figure}
\centering
\begin{minipage}{.5\textwidth}
\centering \includegraphics[width=8cm]{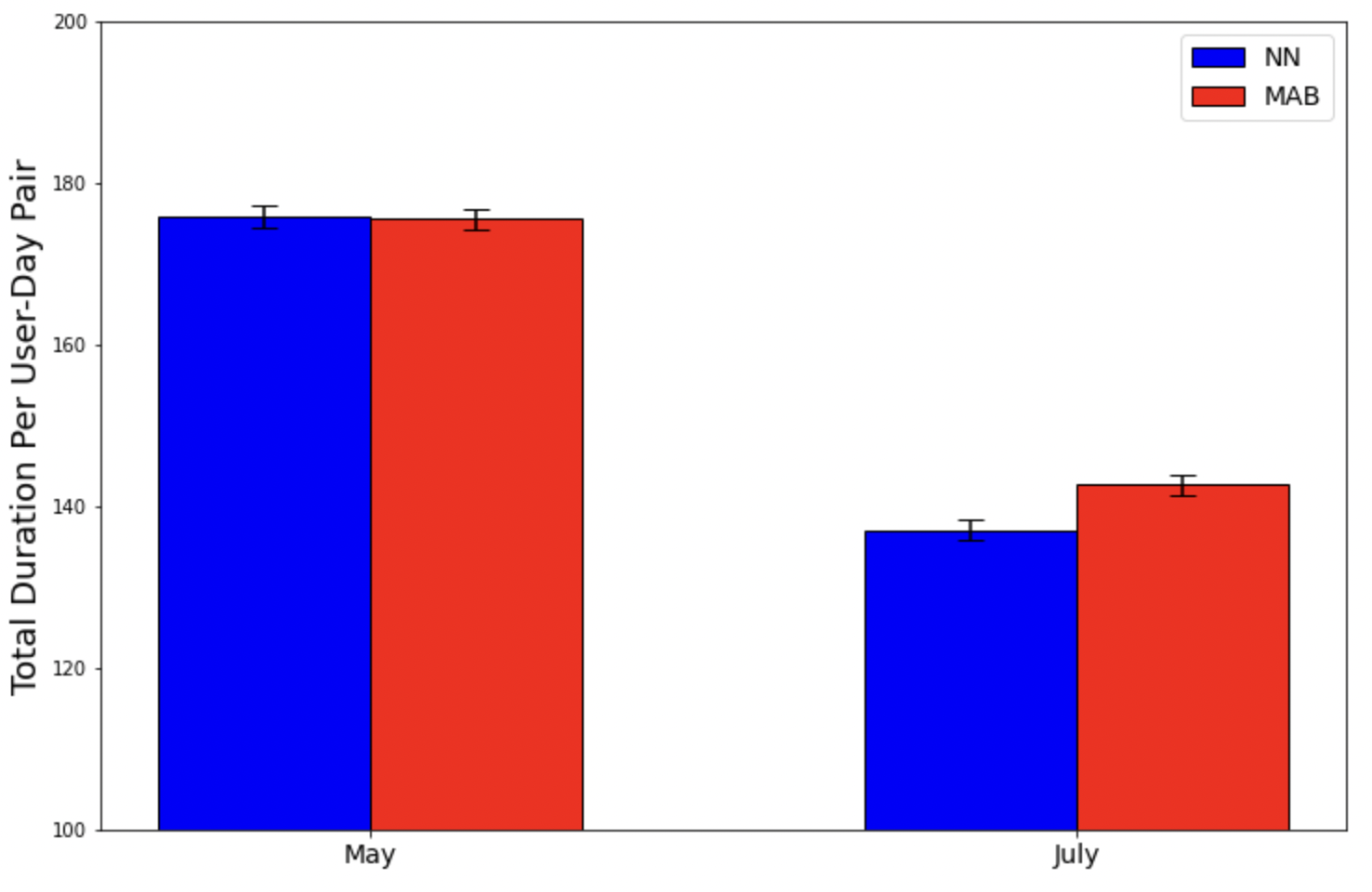}
\captionof{figure}{Duration per user-day pair.}
\label{fig:test1}
\end{minipage}%
\begin{minipage}{.5\textwidth}
  \centering
\includegraphics[width=8cm]{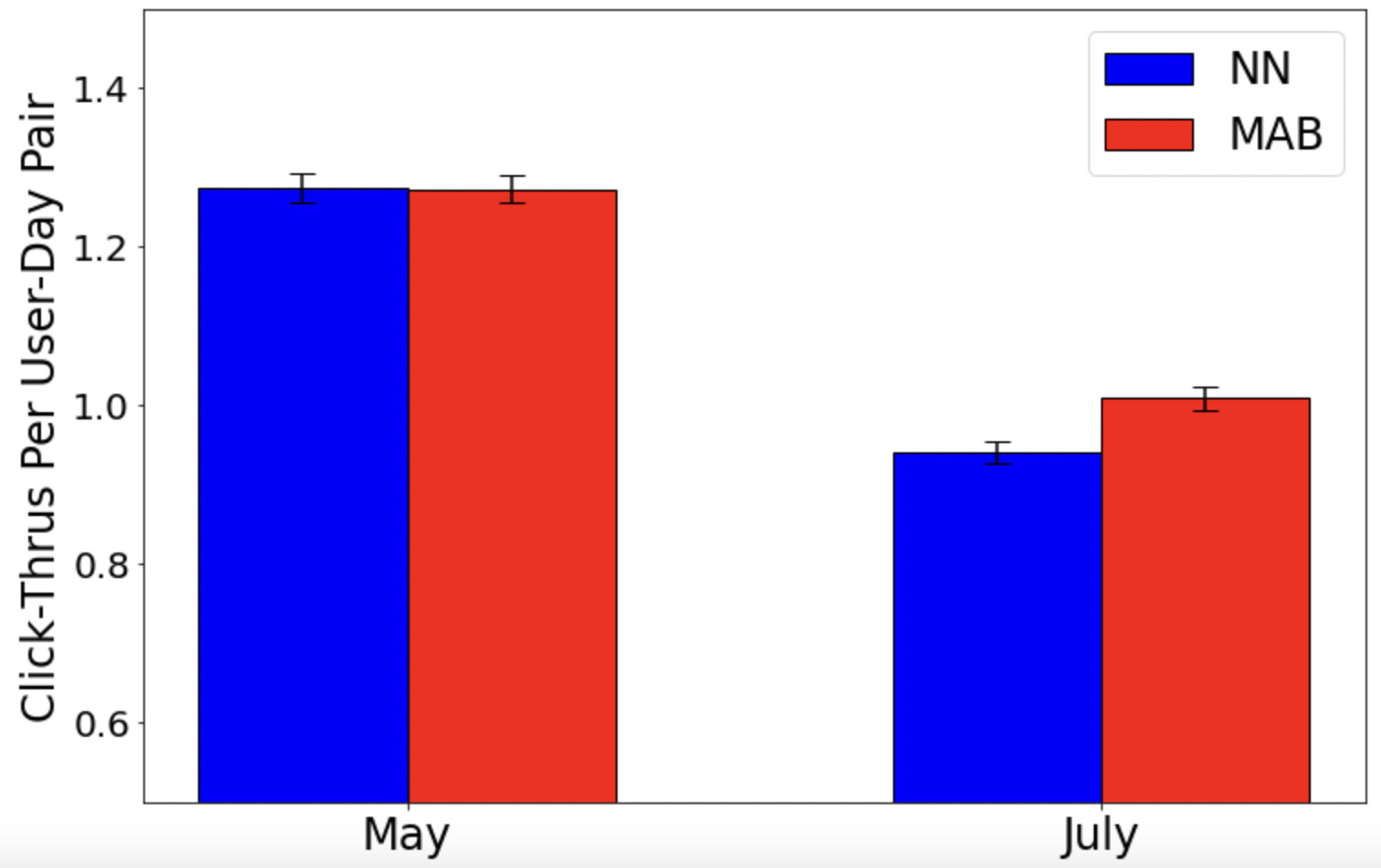}
  \captionof{figure}{Click-through per impression.}
  \label{fig:test2}
\end{minipage}
\end{figure}

\begin{table}[]
\caption{Overall Statistics}
\centering
\begin{tabular}{lllllll}
& & & \multicolumn{2}{c}{May} & \multicolumn{2}{c}{July} \\ \cline{4-7}
& & & \multicolumn{1}{c}{NN} & \multicolumn{1}{c}{MAB} & \multicolumn{1}{c}{NN} & \multicolumn{1}{c}{MAB} \\ \hline
\multirow{5}{*}{Per user per day} & \multirow{3}{*}{Duration} & Mean & 175.910 & 175.548 & 137.059 & 142.618 \\ \cline{3-7} 
& & SE Mean & 0.699 & 0.659 & 0.6081 & 0.597 \\ \cline{3-7}
& & Median  & 44.250 & 44.279 & 32.973 & 34.430 \\ \cline{2-7} 
& \multirow{2}{*}{\#CT} & Mean & 1.275 & 1.273 & 0.941 & 1.010 \\ \cline{3-7} 
 & & SE Mean & 9.251e-03 & 8.814e-03 & 7.276e-03 & 7.549e-03 \\ \hline
\multirow{5}{*}{Per impression} & \multirow{3}{*}{Duration} & Mean    &   3.9697 & 4.0195 & 4.1183 & 4.2391\\ \cline{3-7} 
 & & SE Mean & 4.529e-03  & 4.402e-03 & 5.738e-03 & 5.599e-03 \\ \cline{3-7} 
 & & Median  & 0.693 & 0.697 & 0.702 & 0.703 \\ \cline{2-7} 
 & \multirow{2}{*}{CTR} & Mean    & 2.887e-02 & 2.915e-02 & 2.827e-02 & 3.001e-02 \\ \cline{3-7} 
 & & SE Mean & 4.698e-05 & 4.568e-05 &5.804e-05 & 5.671e-05 \\ \hline
\end{tabular}
\label{tab:overall_stat}
\end{table}

\subsection{Overview}
Before delving into an in-depth analysis of the results, let us define the necessary concepts and explain our approach to handling outliers.
{\icml 
\subsubsection{Business Metrics}
The firm's metrics focus on the {\em per-user-per-day} and the {\em per-impression} level. 
Their primary performance metric of interest is user engagement, measured using duration and click-throughs (CT). 
As a result, we have a total of four metrics, as illustrated in Table \ref{tab:data_types}.

For the per-impression analysis, we assume that interactions are independent among impressions, which is described in Table \ref{tab:overall_stat}. 
In the month of May, prior to implementing our BSE-induced policy, user engagement (measured by duration and CTR) is roughly consistent between the two groups. 
However, after applying our policy within the treatment group (``MAB group'') in July, there is a noticeable and statistically significant increase in user engagement within the MAB group compared to the control group (``NN'' group). 
Further in-depth statistical analysis will be presented in the next sections.


However, the company's primary objective is the engagement per user rather than per impression. 
This motivates us to analyze the results at the per-user-per-day level.
At first sight, it might seem reasonable to consider a user's engagement over all 14 days of the experiment. 
However, this metric can be unreliable because the frequency with which people access the app is influenced by various external factors, such as holidays and weekends, which introduce additional variability.
To address this, we will focus only on the days when a user has at least one impression. 
Formally, for each user $u$ and day $d$ where the user has at least one impression, we define a tuple $(u, d, D_{ud})$, where $D_{ud}$ represents the total duration of the user $u$ on day $d$. 
Consequently, the number of tuples associated with each user can range from $1$ to $14$.

\subsubsection{Handling Outliers} 
Outliers can arise in two ways. 
First, users may unintentionally swipe through two cards in quick succession without engaging genuinely with the first one. 
Consequently, we filter out any impressions with a duration of less than $0.2$ seconds.
Furthermore, a user can leave their device unattended for an extended period, resulting in an unusually long duration of a single card. 
To account for this, we exclude any impressions that last more than $300$ seconds, since most cards' contents can be fully consumed within this time frame.

\subsubsection{Overview of results}
Given the definition of user engagement and our approach to handling outliers, we summarize the experimental results in Table \ref{tab:overall_stat}. 
We also visualize the engagement per user per day in Figure \ref{fig:CT_per_user_per_day} and Figure \ref{fig:duration_per_user_per_day}.
The unit of duration in all tables and figures is in seconds.

We observe that in May, the user engagement levels between the two groups are approximately at the same level. 
However, in July, the MAB group showed a significantly higher average user engagement. 
Additionally, this improvement is further supported by the increase in the median duration, implying that this difference is not primarily driven by a heavy tail in the data distribution.

\begin{figure}
\centering
\begin{minipage}{.5\textwidth}
\centering \includegraphics[width=\linewidth]{fig/_CT_per_user_day.png}
\caption{Number of CTs per user per day} 
  \label{fig:CT_per_user_per_day}
\end{minipage}%
\begin{minipage}{.5\textwidth}
  \centering
 \includegraphics[width=\linewidth]{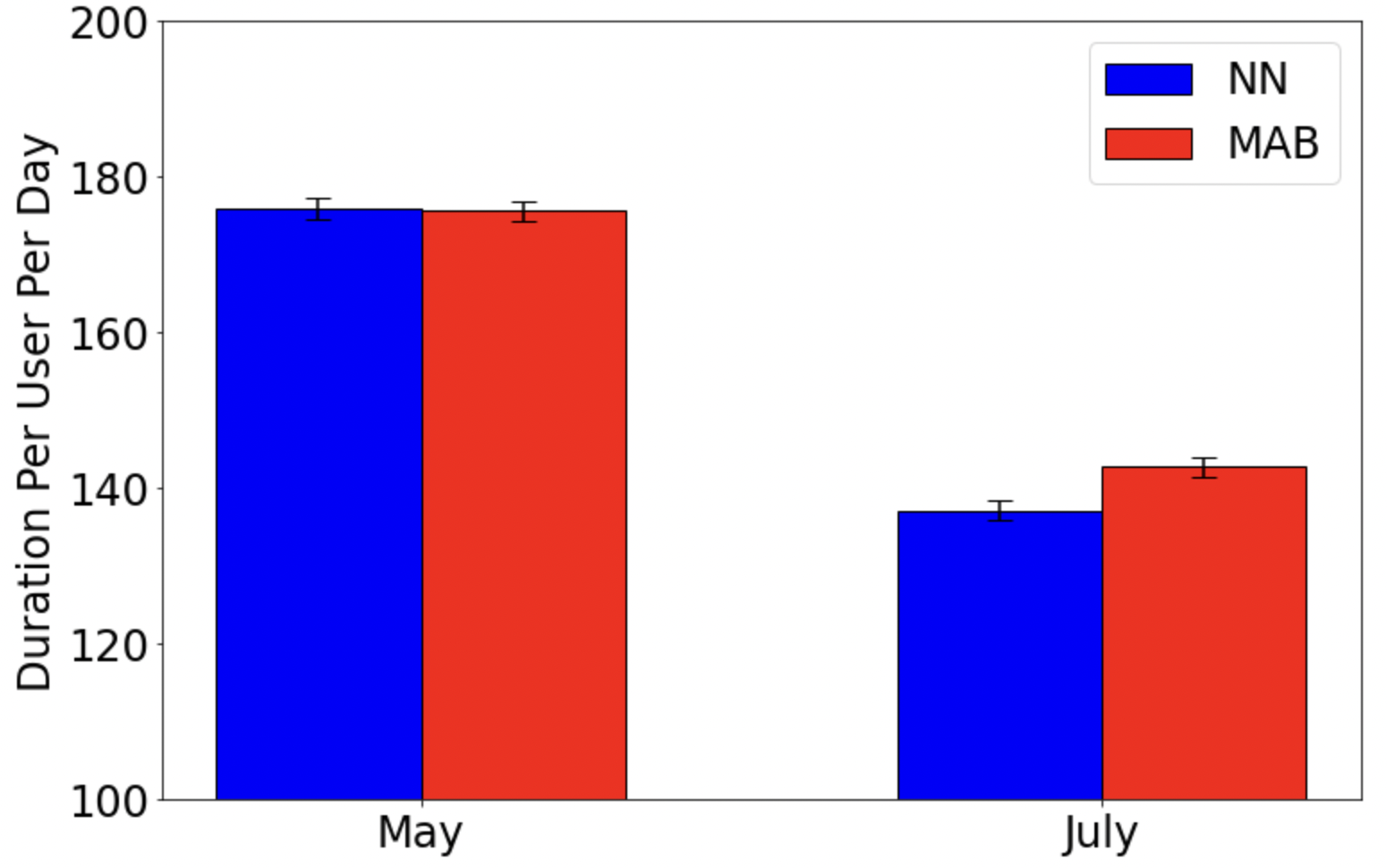}
 \caption{Duration per user per day}
\label{fig:duration_per_user_per_day}
\end{minipage}
\end{figure}

It is important to note that user engagement per user per day decreased from May to July. 
This decline is likely attributed to the fact that the Covid-19 pandemic had reached its peak in May 2021 in the nation where most of the users were located. 
During the lockdown imposed at that time, users may have had more free time to use the app, resulting in an increase in total engagement.
To account for the underlying change in the environment over time, we will perform a {\em difference-in-differences} (DID) analysis in Section \ref{subsec:DID}.

\subsection{Significance Tests}\label{sec:DID}
We will now test whether our observations from Figure \ref{fig:CT_per_user_per_day} and Figure \ref{fig:duration_per_user_per_day} hold statistical significance.
Our primary interest lies in measuring the difference in the differences of user engagement before and after the implementation of our policy.

\bdefn[Difference in differences]
For each month $m\in$ \{May, July\}, we denote by $X^m$ and $Y^m$ the user engagement in the control and treatment group \resp. 
Similarly, we denote by $\overline{X}^m, \overline{Y}^m$ the sample means of $X^m, Y^m$.
The {\em difference in differences} is defined as
\[\Delta = (Y^{\mathrm{July}}-X^{\mathrm{July}}) - (Y^{\mathrm{May}}-X^{\mathrm{May}}).\]
\edefn

To test whether there is a statistically significant improvement, we examine the hypotheses
\[H_0: \ho{E}[\Delta] \le 0 \quad \text{vs.}\quad H_1: \ho{E}[\Delta] >0,\] which correspond to the treatment having a negative effect (null hypothesis) and a positive effect (alternative hypothesis).

\subsubsection{Basic Hypothesis Testing} To begin, we perform a basic one-sided hypothesis test in which the impressions are assumed to be independent.
The $Z$ score is given by
\begin{align}
Z = \frac {\left(\overline{Y}^{\mathrm{July}} - \overline{X}^{\mathrm{July}}\right) - \left(\overline{Y}^{\mathrm{May}} -\overline{X}^{\mathrm{May}}\right)}{\hat S},
\end{align}
where $\hat S$ is the estimated standard deviation, given by 
\begin{align*}
\hat S = \mathrm{SE} \left(\left(\overline{Y}^{\mathrm{July}}-\overline{X}^{\mathrm{July}}\right) - \left(\overline{Y}^{\mathrm{May}}-\overline{X}^{\mathrm{May}}\right)\right) = \sqrt{\mathrm{Var} \left(\left(\overline{Y}^{\mathrm{July}}-\overline{X}^{\mathrm{July}}\right) - \left(\overline{Y}^{\mathrm{May}}-\overline{X}^{\mathrm{May}}\right)\right)}.
\end{align*}
For $Z\in \{ X^{\mathrm{May}}, X^{\mathrm{July}}, Y^\mathrm{May}, Y^\mathrm{July}\}$, we denote by $N_Z$ the sample size of $Z$, and by $S^2_Z$ the sample variance.
Assuming that the samples are independent, we can approximate the above as
\[\hat S \approx \sqrt{\frac {S_{X^{\mathrm{May}}}^2}{N_{X^\mathrm{May}}}  + \frac {S_{X^{\mathrm{July}}}^2 }{N_{X^\mathrm{July}}} + \frac {S_{Y^{\mathrm{May}}}^2}{N_{Y^\mathrm{May}}} +\frac {S_{Y^{\mathrm{July}}}^2}{N_{Y^\mathrm{July}}}}.\]
As indicated in the column ``Basic'' in Table \ref{tab:ht}, the $p$-values for all four metrics are extremely small. 
Therefore, we reject $H_0$ and deduce that the treatment effect is statistically significant.

\begin{table}[]
\centering
\caption{Significance Testing}
\begin{tabular}{llllll}
 & & \multicolumn{2}{c}{Basic} & \multicolumn{2}{c}{Bootstrap} \\ \cline{3-6} 
 & & \multicolumn{1}{c}{$Z$-score} & \multicolumn{1}{c}{$p$-value} & \multicolumn{1}{c}{$Z$-score} & \multicolumn{1}{c}{$p$-value} \\ \hline
\multirow{2}{*}{Per-User-Per-Day}   & Duration & 4.610 &2.018e-06 & 4.6197 & 1.921e-06 \\ \cline{2-6}
 & CT & 4.259 & 1.027e-05 & 4.2556 & 1.042e-05 \\ \hline
\multirow{2}{*}{Per Impression} & Duration & 6.963 & 1.665e-12 & 6.972 & 1.556e-12 \\ \cline{2-6} 
 & CT &12.999 & 6.127e-39 & 12.933 & 1.469e-38\\ \hline
\end{tabular}
\label{tab:ht}
\end{table}

\subsubsection{Hypothesis Testing with the Bootstrap}
The above analysis assumes that the observations are independent, which is a simplification of the real world. However, in reality, these observations may be dependent because (1) each user can appear in both months, (2) each user contributes multiple data points within each month, and (3) the same set of cards is presented to both the treatment and control groups. 

To mitigate this dependence, we implement a bootstrapping procedure.
From each of these four subpopulations, we randomly draw one million ($1\times 10^6$) samples with replacement and redefine each $\bar Z$ as the bootstrap sample mean where $Z\in \{X^{\mathrm{May}}, X^{\mathrm{July}}, Y^{\mathrm{May}},Y^{\mathrm{July}}\}$ (see the ``Bootstrap'' column in Table \ref{tab:ht}).
Consistent with the basic hypothesis test, we still observe very low $p$-values, which further validates our conclusion. 

\subsection{Difference-in-differences Regression}\label{subsec:DID}
To incorporate external factors, we re-evaluate the experimental results using a {\em difference-in-differences} (DID) regression approach. We will begin by illustrating the concept of DID regression using the example of duration per impression in Figure \ref{fig:DID}.

\subsubsection{DID Regression Basics} To apply the DID regression, we vectorize each tuple $(u,d,Y_{ud})$ into a four-dimensional vector $(t_{ud},i_{ud},t_{ud}\cdot i_{ud},Y_{ud})$, where 
\[t_{ud}={\bf 1}[\text{day } d \text{ is in July}] \quad \text{and} \quad i_{ud} ={\bf 1}[\text{user } u \text{ is in MAB group}]\]
are the time and intervention indicators, and $Y_{ud}\in \{C_{ud}, D_{ud}\}$ is the metric under consideration (i.e., CTs or duration of user $u$ on day $d$).
The fundamental premise in a DID analysis (see, e.g., Section 6.2 of \citealt{greene2003econometric})
is that the outcomes can be explained by a linear model, given by
\begin{align}\label{eqn:nov12}
Y_{ud} = \beta_0 + \beta_1 t_{ud} + \beta_2 i_{ud} + \beta_3 t_{ud}\cdot i_{ud} + \eps_{ud}
\end{align}
where $\eps_{ud}\sim N(0,\sigma^2)$ with \unk\ variance $\sigma^2$.
In this model, $\beta_1$ captures the overall trend over time, $\beta_2$ measures the effect of being assigned to the treatment group, and $\beta_3$ represents the interaction effect between time and treatment.

More precisely, if there is no treatment effect, the differences between the two groups should remain unchanged between May and July. 
Consequently, the means of the samples of the four subpopulations, represented as red dots in Figure \ref{fig:DID}, should form a perfect parallelogram.
Now suppose that there is indeed a positive treatment effect. In that case, the top right corner of this quadrilateral will be elevated, as indicated by the green dot in Figure \ref{fig:DID}. 
This indicates that the treatment group experienced a noticeable improvement compared to the control group.
This lift is measured precisely by the variable $\beta_3$. 
Observe that the upper-right corner of the parallelogram is $\beta_1 + \beta_2 + \beta_0$.
In contrast, if both $i_{ud}$ and $t_{ud}$ are set to $1$, then $\ho{E}[Y_{ud}] = \beta_0 + \beta_1 + \beta_2 + \beta_3$, which is higher by $\beta_3$ than the red point on the upper right. 

\begin{figure}
\centering
\includegraphics[width=0.6\linewidth]{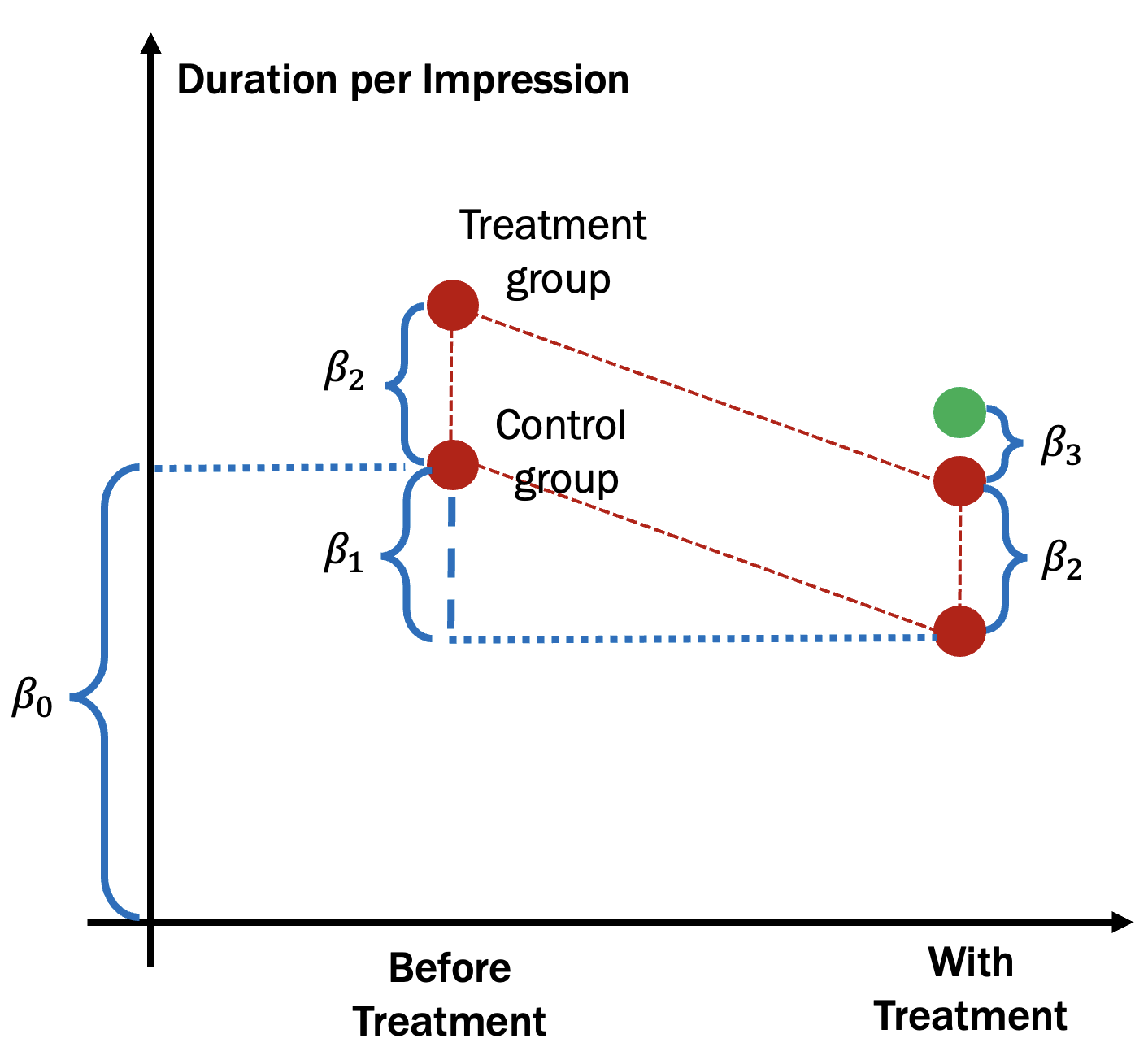}
\centering
\caption{Illustration of DID regression for  duration per impression. The two upper (lower) red dots in the parallelogram the represent the observation from the treatment (control) group. 
We note that in this figure, the overall trend is decreasing, so $\beta_1<0$.}
 \label{fig:DID}
\end{figure}

\subsubsection{Analysis of User Engagement} We have calculated confidence intervals and $p$-values for each coefficient $\beta_i$ in Table \ref{tab:DID}. 
For both duration and CTs, the coefficients $\beta_3$ are positive and have exceptionally low $p$-values. 
This indicates that it is indeed significant whether a user is assigned to the MAB group.
Furthermore, it is worth noting that the coefficients $\beta_2$ have high $p$-values. 
This suggests that the division of users into control and treatment groups appears to be reasonably random in terms of per-user-per-day user engagement.


We also extend our analysis to {\em per-impression} user engagement, as shown in the second half of Table \ref{tab:DID}.
Similarly to the previous analysis, we transform each impression $j$ into a three-dimensional binary vector $(t_j, i_j, Y_j)$, where $Y_j$ represents either the duration or CT indicator for impression $j$.
Note that the duration is a numerical value, allowing us to analyze it using linear regression as described in eqn. \eqref{eqn:nov12}. 
However, for per-impression CTRs, which involve binary outcomes, we use logistic regression. 
We assume that the CTs follows a logistic model, i.e., $Y_j \sim {\rm Ber}\lb(p_j\rb)$ where \[\ln \lb(\frac {p_j}{1-p_j}\rb) = \beta_0 + \beta_1 t_j + \beta_2 i_j + \beta_3 t_j i_j.\]



In contrast to the per-user-per-day regression, in this scenario, all coefficients exhibit very low $p$-values for both duration and number of CTs.
Specifically, within the per-impression analysis, even the coefficient $\beta_2$ for intervention exhibits a low $p$-value, suggesting that the initial user partitioning may not be entirely random with regard to per-impression engagement.
However, this disparity is interpretable: Our experiment was conducted on user groups that the company had been using for several months prior to our field experiment, and these user groups had also been subject to previous experiments. 

We conclude the section by quantifying the improvement. 
One the per-user-per-day level, 
the total duration improved by \[\frac{\beta_3}{\beta_0+\beta_1}= \frac{5.9208}{175.9103-38.8514} \approx 4.32\%.\] 
Similarly, the number of CTs improved by approximately $0.0704/(1.2750 - 0.3341) \approx 7.48\%.$ 
On the per-impression level, the duration improved by approximately $0.0711/(3.9697 - 0.1486) \approx 1.860\%.$
Finally, note that the $\beta$'s for CTR are based on {\it logistic} regression.
The improvement in the odds ratio is $e^{\beta_3}-1\approx 4.854\%.$}

\begin{table}[]
\centering
\caption{Difference-in-differences regression.}
\label{tab:DID}
\begin{threeparttable}
\begin{tabular}{lllcccccc}
 & & & Coef. & Std. Dev. & $t$ & $p$-value & 0.025Q & 0.975Q \\ \hline
\multirow{8}{*}{Per-user-per-day}   & \multirow{4}{*}{Duration} & $\beta_0$ & 175.9103 & 0.640    & 274.941 & 0.000   & 174.656 & 177.164 \\ \cline{3-9} 
 & & $\beta_1$ & -38.8514 & 0.942    & -41.263 & 0.000   & -40.697 & -37.006 \\ \cline{3-9} 
 & & $\beta_2$ & -0.3622  & 0.887    & -0.409  & 0.683   & -2.100  & 1.375           \\
 \cline{3-9} 
 & & $\beta_3$ & {\bf 5.9208}   & 1.303 & 4.544   & {\bf 2.759e-06}   & 3.367   & 8.475  \\ \cline{2-9} 
 & \multirow{4}{*}{\#CT}       & $\beta_0$ & 1.2750  & 0.008    & 153.851 & 0.000   & 1.259  & 1.291 \\ \cline{3-9} 
 & & $\beta_1$ & -0.3341 & 0.012 & -27.394 & 1.616e-165 & -0.358 & -0.310 \\ \cline{3-9} 
 & & $\beta_2$  & -0.0016 & 0.011 & -0.141  & 0.888 & -0.024 & 0.021  \\ \cline{3-9} 
 & & $\beta_3$ & {\bf 0.0704}  & 0.017 & 4.171 & {\bf 1.516e-05} & 0.037  & 0.103 \\ \hline
\multirow{8}{*}{Per-impression} & \multirow{4}{*}{Duration} & $\beta_0$ & 3.9697& 0.005& 863.796 & 0.000& 3.961& 3.979\\ \cline{3-9}
& & $\beta_1$ & 0.1486 & 0.007 & 20.234 & 2.753e-89 & 0.134& 0.163\\ \cline{3-9} 
& & $\beta_2$ & 0.0497 &0.006 & 7.781& 3.597e-15 & 0.037& 0.062 \\ \cline{3-9} 
& & $\beta_3$ & {\bf 0.0711} & 0.010 & 6.998 & {\bf 1.298e-12} & 0.051& 0.091\\ \cline{2-9} 
& \multirow{4}{*}{CTR} & $\beta_0$ & -3.5198 & 0.002 & -2092.794 & 0.000 & -3.523 & -3.517\\ \cline{3-9} 
& & $\beta_1$ & -0.0161 & 0.003 & -5.947 & 1.365e-09 & -0.021 & -0.011 \\ \cline{3-9} 
& & $\beta_2$ &  0.0133 & 0.002 & 5.712 & 5.582e-09 & 0.009 & 0.018\\ \cline{3-9} 
& & $\beta_3$ & {\bf 0.0474} & 0.004 & 12.819 & {\bf 6.417e-38} & 0.040 & 0.055 \\ \hline
\end{tabular}
    \begin{tablenotes}
      \small
      \item Note: All regression are linear regression except for per impression CTR, where we applied logistic regression due to binary labels.
    \end{tablenotes}
  \end{threeparttable}
\end{table}

\section{Conclusion and Future Directions}\label{sec:conlusion}
We introduce the {\em Short-lived High-volume Bandits} (SLHVB) problem.
We present a nearly optimal policy by drawing connections to the batched bandits problem.
Furthermore, we substantiated the practical effectiveness of our policy through a large-scale field experiment conducted in collaboration with a prominent content-card serving company. 
Our policy exhibited substantial performance improvement compared to the firm's concurrent recommender system.
This work paves the way for exploring a variety of directions, which are detailed below.
\benum
\item {\bf Personalization.} An important avenue for further development is to include user features. 
Integrating personalization introduces additional complexity to the challenge, as learning would need to be limited to data points that are close to the specific user of interest.
\item {\bf Bayesian analysis of elimination-type policies.} 
We showed that vanishing loss is achievable if the prior \distr\ 
$D$ \sats\ Assumption \ref{assu:avg}. 
We also showed that if $D$ is ``imbalanced'', then no policy achieves vanishing loss (see Proposition \ref{prop:wc_lb}). 
It would be valuable to quantify how the structure of the prior distribution affects the performance of elimination-type bandit policies.
\item {\bf From learning to optimization.} A related problem is dynamic optimization in the face of a high volume of short-lived arms, where the decision maker must maintain a feasible \sln\ using only the available arms. 
It is not clear whether an elimination-type algorithm would perform effectively in this scenario.
\eenum 

\ACKNOWLEDGMENT{R. Ravi is supported in part by the U.S. Office of Naval
Research under award number N00014-21-1-2243 and the Air Force Office of
Scientific Research under award number FA9550-23-1-0031. Andrew Li is in part supported by NSF CAREER Grant 2238489.
The authors thank Sai Dinesh Dacharaju, Farhat Habib and Alan L Montgomery for helpful discussions.}

\bibliographystyle{informs2014}
\bibliography{main.bib}

\ECSwitch


\ECHead{E-Companion}

{\icml \section{Proof of Proposition \ref{prop:translation}: Regret-to-loss Conversion }\label{apdx:reg_to_loss}
Given a policy $\pi=(\pi_t)$ for the SLHVB problem, we define the loss in round $t$ as
\[L_t = \frac 1n \cdot \ho{E}\lb[\sum_{a\in A_{t-w}^t} \pi_t(a)\cdot (\mu^*_t - \mu_a)\rb].\]
We first decompose $L_t$ into the following {\it internal} and {\it external} loss.

\bdefn[External and Internal Loss]
Let $\ho{A}$ be a semi-\adap\ \alg\ for the BB problem with adaptivity level $\ell$ and $\pi = (\pi_t)$ be the induced policy for the SLHVB problem.
For any round $t$, integer $j\le w$, define  $\Delta_{t,j} := \mu_{\max}(A_{t-w}^t)- \mu_{\max}(A_t).$
We define the {\it external} and {\it internal loss} as 
\begin{align}\label{eqn:032223}
L^{\mathrm{ext}}_t := \ho{E}\lb[ \max_{1\le j\le w} \Delta_{t,j}\rb] 
\quad \text{and} \quad 
L_t^{\rm{int}} := \ho{E} \lb[\sum_{j=0}^\ell \sum_{a\in A_{t-j}} \frac{\pi_t(a)}n \cdot \lb(\mu_{\rm max}(A_{t-j}) - \mu(a)\rb)\rb].
\end{align}
\edefn

Here, the term $\Delta_{t,j}$ 
is {\it external} in the sense that it does {\bf not} depend on the policy. 
Rather, it characterizes the heterogeneity in the maximum mean rewards of the sets $(A_t)$.
On the other hand, the term $L_t^{\rm int}$ is {\em internal}: For each $t$, it measures the gap between the mean rewards of the arms selected from $A_t$ and the maximum mean reward of $A_t$.

\begin{lemma}[Loss Decomposition]
\label{lem:decomp}
In any round $t$, the loss \sats\ $L_t \le L^{\mathrm{int}}_t + L^{\mathrm{ext}}_t.$ 
\end{lemma}
\proof{Proof.}
By the definition of internal and external loss, we have
\begin{align*}
L_t &= \frac 1n \cdot  \ho{E}\lb[\sum_{a\in A_{t-w}^t} \pi_t(a)\cdot (\mu^*_t - \mu_a)\rb] \\
&= \ho{E} \lb[\sum_{j=0}^\ell \sum_{a\in A_{t-j}} \frac{\pi_t(a)}n \cdot \lb(\mu_{\rm max}(A_{t-w}^t) - \mu_a \rb)\rb] &(\text{since } \pi_t(a) = 0 \text{ if } a\in A_{t-w}^t \bs A_{t-\ell}^t)\\
&=  \ho{E} \lb[\sum_{j=0}^\ell \sum_{a\in A_{t-j}} \frac{\pi_t(a)}n \cdot \lb(\Delta_{t,j} + (\mu_{\max}(A_{t-j}) - \mu_a)\rb)\rb] & \text{(by definition of }  \Delta_{i,j})\\
&\le\ho{E}\lb [\max_{1\le j\le w} \Delta_{t,j}\rb] + \ho{E} \lb[\sum_{j=0}^\ell \sum_{a\in A_{t-j}} \frac{\pi_t(a)}n \cdot \lb(\mu_{\max}(A_{t-j}) - \mu_a \rb)\rb]. &\text{(since }\sum_{j=0}^\ell \sum_{a\in A_{t-j}} \frac{\pi_t(a)}{n} = 1)
\end{align*}
\hfill\Halmos



The above suggests that in order to bound the overall loss, it suffices to consider the internal and external losses separately.
We start by showing that the external regret is $\tilde O(1/k)$.
Recall from \Assu\ \ref{assu:avg} that the density of the prior \distr\ is bounded by $C_2> C_1>0$ from above and below.

\bprop[External Loss] \label{prop:ER}
For any round $t$, for any sufficiently large $n$,\footnote{We say that a property ${\cal P}_n$  (e.g., an \ineq) holds for ``any sufficiently large $n$'' if there exists a constant $n_0>0$ such that ${\cal P}_n$ holds whenever $n\ge n_0$.} the external loss can be bounded as \[L^{\mathrm{ext}}_t \leq  \frac{3\rho\log n}{C_1 k}.\] 
\eprop
\proof{Proof.}
Consider any $i\in [w]$ and $Y_i:=\mu_{\max} (A_{t-i})$.
For any $\eps < 1$ and arm $a\in A$, by \Assu\ \ref{assu:avg}, we have $\ho{P}[\mu_a > 1-\eps] \ge C_1 \eps$, or equivalently, 
\[\ho{P}[\mu_a \le 1-\eps] \le 1 - C_1 \eps.\]
In \parti, for $\eps = \frac {2\rho \log n} {C_1 k}$,  it holds that
\[\ho{P}\lb[Y_i < 1 - \frac {2\rho \log n} {C_1 k}\rb] \le \lb( 1 - C_1 \cdot \frac {2\rho \log n} {C_1 k}\rb)^k =  \lb(1 - \frac {2\rho \log n}k\rb)^{\frac { k}{2\rho \log n}\cdot {2\rho \log n}} \le e^{-2\rho \cdot \log n} = n^{-2\rho }.\]
Thus, for the event $B:= \bigcup_{i\in [w]} \lb\{Y_i < 1-\frac {2\rho \log n} {C_1 k}\rb\}$, by the union bound we have  
\[\ho{P}\lb[B\rb]\le wn^{-2\rho}.\]
\IFT\ 
\[\ho{E}\lb[\min_{i\in [w]} Y_i\rb] = \ho{E}\lb[\min_{i\in [w]} Y_i \ \big|\ \bar B\rb] \cdot \ho{P}\lb[\bar B\rb] \ge \lb(1 - \frac {2\rho\log n}{C_1 k}\rb) \cdot \lb(1- w n^{-2\rho}\rb)\geq 1- \frac {3\rho\log n}{C_1 k},\]
where the last \ineq\ follows since $wn^{-2\rho} \le \frac {2\rho\log n}{C_1 k}$ for any sufficiently large $n$.
Therefore, \[\ho{E}\lb[\max_{i\in  [w]}\Delta_{t,i}\rb] \le 1-\ho{E}\lb[\min_{i\in [w]} Y_i \rb] \le \frac {3\rho\log n}{C_1 k}.\eqno\Halmos\]

Now we are ready to prove the regret-to-loss conversion formula.

\noindent{\bf Proof of Proposition  \ref{prop:translation}.} 
Recall that the BSE-induced policy samples a subset $A_t'$ of $k'$ arms from each set $A_t$ of arrivals. 
For any $T>0$, by re-arranging eqn. \eqref{eqn:032223}, we have 
\begin{align}\label{eqn:040323}
\sum_{t=1}^T L_t^{\rm{int}} 
&= \sum_{t=1}^T \ho{E} \lb[\sum_{j=0}^\ell \sum_{a\in A_{t-j}} \frac{\pi_t(a)}n \cdot \lb(\mu_{\rm max}\lb(A_{t-j}\rb) - \mu_a \rb)\rb]\notag\\
&= \sum_{t=1}^T \ho{E} \lb[\sum_{j=0}^\ell \sum_{a\in A_{t-j}} \frac{\pi_t(a)}n \cdot \lb[ \lb(\mu_{\rm max}\lb(A_{t-j}\rb) - \mu_{\rm max}\lb(A'_{t-j}\rb)\rb) + \lb(\mu_{\rm max}\lb(A'_{t-j}\rb) - \mu_a\rb)\rb]\rb]\notag\\
&\le \sum_{t=1}^T \ho{E} \lb[\max_{0\le j\le\ell} \big|\mu_{\rm max}\lb( A'_{t-j}\rb) - \mu_{\rm max}\lb(A_{t-j}\rb)\big| + \sum_{j=0}^\ell \sum_{a\in A_{t-j}} \frac {\pi_t(a)}n \lb(\mu_{\rm max}\lb(A'_{t-j}\rb) - \mu_a\rb) \rb]
\end{align}
where the \ineq\ follows since for any $t$, the total number of arms  the policy selects \sats\ $\sum_{j=0}^\ell \sum_{a\in A_{t-j}} \pi_t(a) = n.$
Note that \[\ho{E} \lb[\max_{0\le j\le\ell} \big|\mu_{\rm max}\lb( A'_{t-j}\rb) - \mu_{\rm max}\lb(A_{t-j}\rb)\big| \rb] \le \frac 1{k'},\]
so we have 
\begin{align*}
\eqref{eqn:040323} &\le \frac T{k'} + \sum_{t=\ell}^{T-\ell} \ho{E}\lb[\sum_{a\in A_t} \frac 1n \sum_{j=0}^\ell \pi_{t+j}(a) \cdot \lb(\mu_{\rm max}(A_t) - \mu_a\rb)\rb] + 2\ell\\
&\le \frac T{k'} + (T-2\ell) \cdot R(n,k') + 2\ell,
\end{align*}
where the $2\ell$ term in the first \ineq\ is because we are summing from $\ell$ to $T-\ell$.
Finally, by Proposition \ref{prop:ER}, we conclude that
\begin{align*}
\mathrm{Loss}_n(\pi) &\le \varlimsup_{T\rar \infty} \frac 1T  \sum_{t=1}^T \lb(L_t^{\rm{int}} + L^{\rm ext}_t \rb) \\
&\le \frac 1{k'} + \frac{3\log k}{C_1 \rho k} + \varlimsup_{T\rar \infty} \frac{(T-2\ell) \cdot R(n, k') + 2\ell}T \\
&= \frac 1{k'} + \frac{3\log k}{C_1 \rho k} + R(n,k').
\end{align*}
where the last identity follows since $R(n,k')$ does not depend on $T$, and $\ell=O(1)$ as $T\rar \infty$.\hfill\Halmos


\section{Proof of Proposition \ref{prop:ub_gen_ell}: Bayesian Regret of BSE}\label{apdx:ub_proof}
We will prove the $\tilde O((k/n)^{\ell/(\ell+2)})$ bound on the Bayesian regret for the BSE \alg\ with the revised geometric grid specified in Definition \ref{def:geometric_grid}.
To help the reader better digest the ideas, we first examine the case where $\ell=1$ as a warm-up in Section \ref{sec:ell=1}. 
Then in Section \ref{sec:ell=2}, we explain how to improve the bound by adding one more level of adaptivity (i.e., $\ell=2$) by leveraging the structure of the prior distribution.
Finally, in Section \ref{apdx:ub_gen}, we generalize the idea beyond $\ell=1,2$ by presenting an analysis for an \arb\ $\ell\ge 1$.

\subsection{Preliminaries}\label{sec:prelim}

We first introduce some tools for our analysis.
For any arm $a$, consider i.i.d. Bernoulli \rv s $(Z_i^a)_{i\in [n]}$ with mean $\mu_a$ which represents the reward of arm $a$ when it is selected for the $i$-th time.
We first state a standard concentration bound, whose proof can be found in, e.g., Section 2 of \cite{vershynin2018high}.

\begin{lemma}[Concentration Bounds]
\label{lem:hoeffding}
Let $Z_1,...,Z_m$ be \indep\ \rv s supported on $[0,1]$, and $\bar Z = \frac{1}{m}\sum_{i=1}^m Z_i$, then for any $\delta>0$, it holds that
\begin{align*}
\ho{P}(|\bar Z - \ho{E}(\bar Z)| > \delta) &\leq \exp\lb(-\frac {\delta^2 m}2 \rb) &\text{\rm (Hoeffding's \ineq)},\\
\ho{P}\lb(|\bar Z - \ho{E}(\bar Z)| > \delta\cdot \ho{E}(\bar Z)\rb) &\leq \exp\lb(-\frac {\ho{E}\left(\bar Z\right)} 2 \delta^2 m\rb)
&\text{\rm (Chernoff's \ineq).}
\end{align*}
\end{lemma}

The following event says that the \emp\ mean rewards of all arms available at time $t$ \sat\ Hoeffding's \ineq.

\bdefn[Clean Event]
For any integers $m\ge 1$ and arm $a$, define the event 
\[\mathcal{C}_a^m = \lb\{\Big|\sum_{j=1}^m Z_i^a - m\mu_a\Big| \le \sqrt{ 2(\rho+3)m}\cdot \log n \rb\}.\] 
We define the {\it clean event} as 
$\mathcal{C} = \bigcap_{m\in [n], a\in A} \mathcal{C}_a^m$.
\edefn

We will use Lemma \ref{lem:hoeffding} to show that $\mathcal{C}$ occurs with high \prb. 
Denote by $\overline{\mathcal{C}}$ the complement of the event.

\begin{lemma}[Clean Event is Likely]\label{lem:clean_event}
It holds that $\ho{P} \lb[\overline{\mathcal{C}}\rb] \le n^{-2}.$
\end{lemma}
\proof{Proof.}
Fix an \arb\ $m\in [n]$ and arm $a\in A$.
By Hoeffding's \ineq\ (Lemma \ref{lem:hoeffding}),
\begin{align*}
\ho{P}\lb[\overline{\mathcal{C}_a^m}\rb] &= \ho{P}\lb[ \Big|\sum_{j=1}^m Z_i^a - m\mu_a\Big| > \sqrt {2m(\rho+3) \log n} \rb] \le \exp\lb(-\frac 1{2m} \cdot 2m(\rho+3) \log n\rb) \le    n^{-3}.
\end{align*}
By the union bound over all $k$ arms and $m\in [n]$, we have 
\begin{align*}
\ho{P}\lb[\bigcup_{m\in [n], a\in A}\overline{\mathcal{C}_a^m} \rb]
&\le \sum_{a\in A, m\in [n]} \ho{P}\lb[\overline{\mathcal{C}_a^m}\rb] \\
&=\sum_{a\in A, m\in [n]} \ho{P}\lb[\Big|m\mu_a - \sum_{j=1}^m Z_i^a\Big| > \sqrt {2m(\rho + 3) \log n} \rb] \\
&\le nwk n^{-\rho-3} \le n^{-2},
\end{align*} 
where the last \ineq\ follows since $k=n^\rho$. \hfill\Halmos

To bound the number of arms that survive a certain number of elimination phases, we will repeatedly apply the following \ineq.

\begin{lemma}[$3\Delta$-Inequality]
\label{lem:3delta}
Let $X=\{x_j\}_{j\in [k]}, X'=\{x'_j\}_{j\in [k]}$ be two sets of real numbers and $\Delta := \max_{j\in [k]} |x_j-x'_j| \le \Delta$.
\Sps\ for some $i\in [k]$ we have $x'_i \ge \max X' - \Delta$.
Then, \[x_i \ge \max X - 3\Delta.\]
\end{lemma}
\proof{Proof.}
Since $|x_j - x'_j|\le \Delta$ for all $j\in [k]$, we have $|\max X - \max X'|\le \Delta$. Thus, $x'_i \ge \max X' - \Delta \ge \max X - 2\Delta$. Therefore, $x_i \ge x_i'-\Delta \ge \max X - 2\Delta - \Delta = \max X -3\Delta$.\hfill \Halmos

\subsection{Warm-up: $\ell=1$}
\label{sec:ell=1}
Now let us consider $\ell=1$.
First, we prove that the surviving arms are close to optimal.

\begin{lemma}[Suboptimality of $S_1$]\label{lem:suboptimlaity_S1}
Let $S_1\sse A$ denote the set of surviving arms after the first phase in the BSE \alg\ with any grid $\eps_0$.
Define $\delta_1 = 3\lb(\frac{\eps_0 n}k\rb)^{-1/2} \log^{1/2} n$. 
Then, on the clean event $\mathcal{C}$, it holds that \[\mu_{\max}(A) - \mu_{\min} (S_1) \le \delta_1.\]
\end{lemma}
\proof{Proof.}
By Hoeffding's \ineq\ (Lemma  \ref{lem:hoeffding}), we have
$|\hat \mu_a - \mu_a|\leq \frac 13 \delta_1$ for all arms $a\in A$.
Consider the arms \[\hat a = \arg\max_{a\in A} \hat \mu_{a} \quad \text{and} \quad a^* = \arg\max_{a\in A} \mu_a.\]
By definition of $S_1$ (see \Alg\ \ref{alg:bse}), an arm $a$ survives {\em only} if $|\hat \mu_a - \hat \mu_{\hat a}| \leq \frac 13 \delta_1$. 
Thus, by the $3\Delta$-\ineq\ (Lemma \ref{lem:3delta}) we conclude that $\mu_a \geq \mu_{\max} (A) - \delta_1$ for any $a\in S_1$.\hfill\Halmos

The above holds for any grid, and we will use it again in the next subsection, where we analyze the scenario where $\ell=2$.
Now consider the level-$1$ BSE \alg\ with the revised geometric grid
\[\eps_{0,1}^\star = \lb(\frac kn\rb)^{1/3}\log^{1/3} n\]
given in Definition \ref{def:geometric_grid}.
We obtain the following by combining Lemma \ref{lem:suboptimlaity_S1} with the regret decomposition formula given in Lemma
\ref{lem:decomp}.

\begin{proposition}[Bayesian Regret of ${\rm BSE}^\star_1$]\label{prop:ell=1}
Let $D$ be any prior \distr.
Then the BSE \alg\ with grid $\eps_{0,1}^\star$ has Bayesian regret 
\[\mathrm{BR}_n ({\rm BSE}^\star_1, D) \le 5\lb(\frac kn \rb)^{1/3}\log^{1/3} n.\]
\end{proposition}
\proof{Proof.}
By Lemma \ref{lem:suboptimlaity_S1}, on the clean event $\cal C$ we have \[\mu_{\max}(A) - \mu_{\min}(S_1) \le \delta_1.\] 
It follows that
\begin{align*}
\mathrm{BR}_n ({\rm BSE}^\star_1, D) &\le \eps_{0,1}^\star + (1-\eps_{0,1}^\star) \cdot \ho{E} \lb[ \mu_{\max}(A) - \mu_{\min} (S_1) \rb] \\
&\le \eps_{0,1}^\star + \delta_1 \cdot \ho{P}\lb[\mathcal{C}\rb] + 
\ho{P}\lb[\overline{\mathcal{C}}\rb]\\ 
&\le \eps_{0,1}^\star + \delta_1 + n^{-1},
\end{align*}
where the last \ineq\ follows from Lemma \ref{lem:clean_event}.
Expanding $\delta_1$ using Lemma \ref{lem:suboptimlaity_S1}, we conclude that
\begin{align*}
\mathrm{BR}_n ({\rm BSE}^\star_1, D) \leq 4\lb(\frac kn\rb)^{\frac 13}\log^{\frac 13} n +n^{-1} \le 5\lb(\frac kn\rb)^{\frac 13}\log^{\frac 13} n,
\end{align*}
where the last \ineq\ follows since $\frac 1n \le \frac kn \le \lb(\frac kn\rb)^{1/3}$.\hfill\Halmos}

We want to highlight that the above analysis remains valid for {\it any} prior \distr\ $D$, irrespective of whether it satisfies Assumption \ref{assu:avg}. 
However, this is no longer true in our analysis for $\ell=2$. 
In this scenario, it is essential for $D$ to satisfy Assumption \ref{assu:avg}, so that we can bound the number of survivors after the initial elimination phase. 
We should also emphasize that the outcome mentioned above does not place any restrictions on the parameter $\rho$. 
This observation aligns with the definition of the threshold exponent (as defined in Definition \ref{def:thr_expo}) where $\theta_1=0.$

{\icml
\subsection{Bounding the Number of Survivors: Analysis for $\ell=2$}\label{sec:ell=2}
Next we show that with $\ell=2$, we can achieve better performance than the Bayesian regret of
$\tilde O((k/n)^{1/3})$ in the $\ell=1$ case. We start with an informal recap of the BSE \alg\ for $\ell=2$:
\benum
\item First, explore each arm equally many times;
\item 
Compute a subset $S_1$ of surviving arms whose \ci s are {\em not} dominated by any other arm;\footnote{we say an arm $a$ dominates another arm $a'$ if the lower  confidence bound  of $a$ is greater than the upper confidence bound of $a'$}
\item Select another batch of arms and compute a further subset $S_2 \subseteq S_1$ in a similar manner;
\item 
Finally, in the exploitation phase,  choose an \arb\ arm from $S_2$.
\eenum
The key step in the analysis is bounding the number of survivors after the first phase. With an {\bf upper} bound on $|S_1|$, we can then {\bf lower}-bound the number of times that each arm in $S_1$ is selected in the second phase. 
This leads to a guarantee on the width of the \ci. 

To implement this idea, we introduce the following event to simplify the analysis, which says that the number of $\delta$-good arms does not deviate much from its expectation.


\bdefn[Uniform Event]
Consider any constant $\delta \in (0,1)$, and define $N_\delta = \big|\{a\in A: \mu_a \geq \mu_{\max}(A)-\delta\}\big|$. We define the {\it uniform event} as  
\[U_\delta = \lb\{\frac 12 C_1 \delta k \le N_\delta \leq \frac 32 C_2 \delta k \rb\}.\]
\edefn

We show that the uniform event is likely to occur when $k$ is large.
This is a direct consequence of Assumption \ref{assu:avg}, and its proof is similar to that of Proposition \ref{prop:ER}.

\begin{lemma}[Uniform Event is Likely]\label{lem:good_event} 
\Sps\ $\delta >0$ and $\delta k \ge \frac 8{C_1} \log n$, then 
\[\ho{P}\lb[\overline{U_\delta}\rb] \leq 2n^{-1}.\] 
\end{lemma} 
\proof{Proof.} Index the arms from $1$ to $k$, and denote by $f$ the density of the prior $D$.
Subsequently, we fix an \arb\ $i\in [k]$. 
Denote by $Z_i = {\bf 1}[\mu_i \ge \mu_{max}-\delta]$ and consider the highest reward rate in $[k]\bs \{i\}$, formally defined as
\[\mu_{-i}^{\max} := \max \lb\{\mu_1,\dots,\mu_{i-1},\mu_{i+1},\dots, \mu_k\rb\}.\]
Since $\mu_i$'s are \indep, we deduce that $\mu_{-i}^{\max}$ and $\mu_i$ are \indep, and hence 
\begin{align*} 
\ho{P}\lb[Z_i =1\ |\ \mu_{\max} = 1-\gamma\rb] = \ho{P}\lb[\mu_i \in [1- \gamma - \delta,\  1-\gamma]\ |\ \mu^{\max}_{-i}-\delta\rb] = \int_{1-\gamma-\delta}^{1-\gamma} f(z)\ dz.
\end{align*}
To bound the above, recall that by \Assu\ \ref{assu:avg}, $C_1 \le f(z) \le C_2$, so 
\[C_1 \delta\  \le\ \ho{P}\lb[\ Z_i = 1\ |\ \mu_{\max} = 1-\gamma\ \rb]\ \le \ C_2 \delta.\]
Note that conditional on the event $\{\mu_{\max} = 1-\gamma\}$, the \rv s $Z_i$ are still i.i.d., so by Chernoff's \ineq\ (see Lemma \ref{lem:hoeffding}) we obtain 
\begin{align*}
\ho{P}\lb[\bar G\ \Big|\ \mu_{\max} = 1-\gamma \rb]
&\le \ho{P} \lb[\sum_{i\in [k]} Z_i > \frac 32 \cdot C_2 \delta k \ \Big|\ \mu_{\max} = 1-\gamma \rb] + \ho{P}\lb[\sum_{i\in [k]} Z_i < \frac 12 \cdot C_1 \delta k \ \Big|\ \mu_{\max} = 1-\gamma\rb]  \\
&\le 2 \exp\lb(-\frac 12 \cdot \lb(\frac 12\rb)^2 \cdot C_1 \delta k\rb) \\
&\le 2 \exp\lb(-\frac {C_1}8 \cdot \frac 8{C_1} \log n\rb) \\
& = 2k^{-\frac {C_1}8 \cdot \frac 8{C_1 \rho}} = 2n^{-1}.
\end{align*}
Therefore,
\begin{align*}
\ho{P}[\bar G] = \ho{E}_\gamma \lb[\ho{P}[\bar G\ |\ \mu_{\max}=1-\gamma] \rb] \le 2n^{-1}, 
\end{align*}
where $\ho{E}_\gamma$ is over the randomness of $\mu_{\max}$. \hfill\Halmos

Next, we bound the suboptimality of the arms in $S_2$. To this end, observe that on the event $U_{\delta_1}$ and $\mathcal{C}$, we have $|S_1|\sim \delta_1 k$.
Thus, in the second phase, each surviving arm is selected $\gtrsim \frac {\eps_1 n}{\delta_1 k}$ times and therefore the confidence intervals have widths $\ls \lb(\frac {\eps_1 n}{\delta_1 k}\rb)^{-1/2}$. We now formally define these quantities.

\bdefn[Widths of Confidence Intervals]\label{def:delta}
For any $\eps_0,\eps_1 \in (0,1)$, define 
\[\delta_1 = \delta_1 (\eps_0) =  \frac 8 {C_1} \eps_0^{-\frac 12} \lb(\frac kn\rb)^{\frac 12} \log n \quad \text{and} \quad 
\delta_2 = \delta_2 (\eps_0, \eps_1) = 
6^{\frac 32} \sqrt{\lb(\rho+3\rb) \frac{C_2}{C_1}} \cdot \eps_0^{-\frac 14}  \eps_1^{-\frac 12} \lb(\frac kn\rb)^{\frac 34} \log^{\frac 54} n.\]
\edefn

Under the above notation, we can bound the suboptimality of the arms in $S_2$ as follows.

\begin{lemma}[Suboptimality of $S_2$]
\label{lem:w=2}
Consider the \alg\ ${\rm BSE}_2(\eps_0,\eps_1)$ such that $\delta_1 = \delta_1(\eps_0,\eps_1)$ \sats\ $\delta_1 k> \frac 8{C_1} \log n$. 
If the clean event $\mathcal{C}$ and the uniform event $U_{\delta_1}$ both occur, then
\[\mu_{\max}(A) - \mu_{\min}(S_2)\le \delta_2.\]
\end{lemma}
\proof{Proof.}
We first upper bound the cardinality of $S_1$.
By Lemma \ref{lem:suboptimlaity_S1}, since $\mathcal{C}$ occurs, we have
$\mu_{\max}\lb(A\rb) - \mu_{\min} \lb(S_1\rb) \le \delta_1.$
\IOW, for an arm $a\in A$ to survive the first phase, its mean reward must be $\delta_1$-optimal.
Thus by Lemma \ref{lem:good_event}, on the uniform event $U_{\delta_1}$ we have 
\begin{align}\label{eqn:040723}
|S_1| \le \frac 32 C_2 \delta_1 k.
\end{align} 

Note that in phase two, each arm in $S_1$ is selected $m_1 := \frac{\eps_1 n}{|S_1|}$ times.
By the definition of the clean event $\mathcal{C}$, the \emp\ mean reward of every arm deviates from the mean by at most $\sqrt{2(\rho+3)}\cdot m_1^{-1/2} \log^{1/2} n$.
Thus by the $3\Delta$-\ineq\ (Lemma \ref{lem:3delta}), we have 
\[\mu_{\max}\lb(A\rb) - \mu_{\min}\lb(S_2\rb) \le  3\sqrt{2(\rho+3)}\cdot m_1^{-1/2} \log^{1/2} n.\] 
To further bound the above, we use \eqref{eqn:040723} to lower bound $m_1 = \frac{\eps_1 n}{|S_1|}$.
Specifically, we have
\begin{align}\label{040723b}
\mu_{\max}\lb(A\rb) - \mu_{\min}\lb(S_2\rb) 
&\le 3\sqrt{2(\rho+3)} \cdot  \lb(\frac{\eps_1 n}{\frac 32 C_2 \delta_1 k}\rb)^{-\frac 12} \log^{\frac 12} n \notag\\
&\le 3^{\frac 32}\sqrt{(\rho+3) C_2}  \cdot \eps_1^{-\frac 12} \delta_1^{\frac 12} \lb(\frac kn\rb)^{\frac 12} \log^{\frac 14} k\cdot \log^{\frac 12} n \notag\\
&= 6^{\frac 32}\sqrt{\lb(\rho+3\rb) \frac{C_2}{C_1}} \cdot \eps_1^{-\frac 12} \eps_0^{-\frac 14} \lb(\frac kn\rb)^{\frac 34} \log^{\frac 14} k\cdot \log n \notag\\
&= \delta_2,
\end{align}
where the second last equality follows since
\[ \delta_1 = \frac 8{C_1}\eps_0^{-\frac 12} \lb(\frac kn\rb)^{\frac 12} \log n\] 
and the last equality follows from the definition of $\delta_2$; see our Definition \ref{def:geometric_grid}.
\hfill\Halmos

We emphasize that the \assu\ $\delta_1 k > \frac{8\log n}{C_1}$ holds if $\rho \ge \theta_2$, where we recall that $\theta_\ell$ is the level-$\ell$ threshold exponent, formally defined in Definition \ref{def:thr_expo}.
We can now bound the Bayesian regret of the BSE \alg\ under any grid. 

\bprop[Bayesian Regret Under Arbitrary Grid, $\ell=2$]\label{prop:ell=2}
\Sps\ the prior \distr\ $D$ \sats\ Assumption \ref{assu:avg}.
\Sps\ $0<\eps_0<\eps_1 <1$ and  $\delta_1 k > \frac{8\log n}{C_1}$. 
Then,
\[\mathrm{BR}_n \lb({\rm BSE}_{\ell=2} \lb(\eps_0, \eps_1\rb), D\rb) \le \eps_0 + \eps_1 \delta_1 + \delta_2 +  1.\]
\eprop

\proof{Proof.}
By definition of regret, we have 
\begin{align*}
\mathrm{BR}_n \lb({\rm BSE}_2 \lb(\eps_0, \eps_1\rb), D\rb) 
= \eps_0 + \eps_1 \cdot \ho{E} \lb[\mu_{\max}(A) - \mu_{\min}(S_1)\rb] + \lb(1-\eps_0 - \eps_1 \rb) \cdot \ho{E} \lb[ \mu_{\max}\lb(A\rb) - \mu_{\min}\lb(S_2\rb)  \rb].
\end{align*}
We bound each term \sep ely.
By Lemma \ref{lem:clean_event} and Lemma \ref{lem:suboptimlaity_S1}, the second term can be bounded as
\begin{align*}
&\quad \ho{E}\lb[\mu_{\max}(A) - \mu_{\min}(S_1)\rb] \notag\\
&= \ho{E}\lb[\mu_{\max}(A) - \mu_{\min}(S_1)\mid \mathcal{C}\rb] \cdot\ho{P}[\mathcal{C}] + \ho{E}\lb[\mu_{\max}(A) - \mu_{\min}(S_1)\mid \overline{\mathcal{C}}\rb] \cdot \ho{P}[\overline {\mathcal{C}}]\notag\\
&\leq \delta_1\cdot 1 + 2n^{-1}.
\end{align*}
By Lemma \ref{lem:w=2}, if the events $U_{\delta_1}$ and $\mathcal{C}$ both occur, then $\mu_{\max}(A) - \mu_{\min}(S_2) \le \delta_2.$
Further, by Lemma \ref{lem:good_event}, we have $\ho{P}\lb[U_{\delta_1}\rb] \le 2n^{-1}$ whenever $\delta_1 k \ge   \frac {8\log n}{C_1}$.
Combining the above facts, we obtain
\[\mathrm{BR}_n \lb({\rm BSE}_2 \lb(\eps_0, \eps_1\rb)\rb) \le \eps_0 + \eps_1 \cdot \lb(\delta_1 + 2n^{-1}\rb) + \delta_2 + n^{-1}. \eqno \Halmos\]

Next, we apply Proposition \ref{prop:ell=2} to the revised geometric grid, i.e., choose $\eps_i = \eps_i^\star$.
To reduce clutter, we write $\delta_j^\star := \delta_{j}(\eps_0^\star, \eps_1^\star)$.
The proof of the following is \strfwd\ -- we only need to verify that $\delta_1^\star k \ge \frac {8\log n}{C_1}$. 

\bcoro[Bayesian Regret of ${\rm BSE}^\star_2$]
\Sps\ the prior $D$ \sats\ Assumption \ref{assu:avg}. 
Then for any $\rho \ge \theta_2$, we have
\[\mathrm{BR}_n \lb({\rm BSE}^\star_{\ell=2}, D \rb)= \tilde O\lb(\lb(\frac kn\rb)^{1/2}\rb).\]
\ecoro
\proof{Proof.}
Recall from Definition \ref{def:geometric_grid} that $\eps_0^\star = \lb(\frac kn\rb)^{\frac 12}$, and from Definition \ref{def:delta} that for any $\eps_0$ we defined
\[\delta_1 = \delta_1 (\eps_0) =  \frac 8 {C_1} \eps_0^{-\frac 12} \lb(\frac kn\rb)^{\frac 12} \log n.\]
Expanding the expressions for  $\eps_0^\star$ and $\delta_1^\star=\delta_1(\eps_0^\star)$, we have 
\begin{align*}
\delta^\star_1 k \ge \frac 8{C_1} \eps_0^{-\frac 12} \lb(\frac kn\rb)^{\frac 12} \log n \cdot k = \frac 8{C_1} \cdot k^{\frac 54} \cdot n^{-\frac 14} \cdot \log n \ge \frac 8{C_1} \log n,
\end{align*}
where the last \ineq\ follows since $\rho\ge \theta_2 = \frac 15$.
By Proposition \ref{prop:ell=2}, we conclude that
\[\mathrm{BR}_n \lb({\rm BSE}_{\ell=2}^\star, D\rb) \le \eps_0^\star + \eps_1^\star \delta^\star_1 + \delta^\star_2 +1
= O\lb(\lb(\frac kn\rb)^{\frac 12} \log^{\frac 54} n\rb). \eqno\Halmos\]

\subsection{Analysis for General $\ell\ge 1$}
\label{apdx:ub_gen}
We now extend the analysis to the general $\ell$ case.
A level-$\ell$ BSE policy iteratively explores the arms and computes their \ci s. 
We will bound that the widths of these \ci s as follows. 

\bdefn[Width of Confidence Interval]\label{def:delta_j}
Given any $\eps_0,\dots, \eps_{\ell-1}\in \real$, for each $i=1,2, \dots, \ell$ we define 
\[\delta_i = \delta_i(\eps_0,\dots,\eps_{i-1}) = 3\cdot 15^{i-1}\cdot C_2^{\frac {i-1}2} \cdot \eps_0^{-2^{-i}} \cdot \eps_1^{-2^{-(i-1)}} \cdot...\cdot \eps_{i-1}^{-\frac 12} \cdot \left(\frac kn\right)^{1-2^{-i}} \cdot  \log^{1+\frac {i-1}4} n.\]
\edefn

Next, we show that the loss on the $j$-th layer is bounded as follows.
of Lemma \ref{lem:w=2}, which exclusively addressed the $\ell=2$ case.

\begin{lemma}[Suboptimality of $S_j$, General $j\ge 1$]\label{lem:layer_j_loss}
Denote by $\mu^* = \mu_{\max} (A)$.
\Sps\ the events $U_{\delta_1},\dots, G_{\delta_j}$ and $\mathcal{C}$ occur. Then, \begin{align}\label{eqn:induction_goal}
|S_j|\le 6C_2 \delta_j k\cdot \log^{1/2} k \quad\text{and}\quad \mu^* - \mu_{\min}(S_j)\leq \delta_j. \end{align}
\end{lemma}
\proof{Proof.} We show this inductively on $j$. 
To show the base case $j=1$, note that by Lemma \ref{lem:suboptimlaity_S1}, we have $\mu^* - \mu_{\min}(S_1)\leq \delta_1$, and $|S_1|\le 6C_2 \delta_1 k \log^{1/2}k$. 
Therefore, the claim holds for $j=1.$

Now consider $j\ge 2$. As induction hypothesis, we assume that the claim holds for $1,\dots,j-1$. 
Then, $|S_{j-1}| \le 6C_2 \delta_{j-1} k \log^{1/2} k$.
Denote by $m_{j-1}$ the number of times an arm in $S_{j-1}$ is selected. By the induction hypothesis, it holds that
\begin{align}\label{eqn:012423c}
m_{j-1} = \frac{\eps_j n}{|S_{j-1}|} \geq \frac{\eps_j n}{6C_2 \delta_{j-1} k \cdot \log^{1/2} k}\end{align} times. Since the clean event $\mathcal{C}_t$ occurs, the \emp\ mean of each arm from $S_{j-1}$ deviates from the mean by $3m_{j-1}^{1/2}\log^{1/2}n$.
Thus, if an arm $a$ survives, then
\begin{align}\label{eqn:012423b}
\mu^* - \mu_{\min}\lb(S_{j-1}\rb) \le 9 m_{j-1}^{- \frac 12}\log^{\frac 12} m_{j-1}.
\end{align}
Since the good event $U_{\delta_j}$ occurs, we have 
\begin{align}\label{eqn:092623}
|S_j| \le 6C_2 \cdot \lb(3m_{j-1}^{-\frac 12}\log^{\frac 12} m_{j-1} \rb) k \log^{1/2} k.\end{align}
To simplify the above, note from eqn. \eqref{eqn:012423c} that 
\[3m_{j-1}^{-\frac 12}\log^{\frac 12} m_{j-1}\le 3\cdot \lb(\frac{\eps_j n}{6C_2 \delta_{j-1} k \cdot \log^{1/2} k}\rb)^{-\frac 12} \log^{\frac 12} \lb(\frac{\eps_j n}{6C_2 \delta_{j-1} k \cdot \log^{1/2} k}\rb) \le \delta_j.\] 
Combining the above with eqn. \eqref{eqn:092623}, we deduce that $|S_j| \le 6C_2 \delta_j k \log^{1/2} k,$ and hence the first part of our goal (i.e., eqn. \eqref{eqn:induction_goal}) follows.

Next, we prove the second part of our goal. We start with simplifying eqn. \eqref{eqn:012423b}. 
By expanding $m_{j-1}$, we have 
\begin{align}\label{eqn:012423}
\mu^* - \mu_{\min}\lb(S_{j-1}\rb) &\le 3m_{j-1}^{-\frac 12}\log^{\frac 12} m_{j-1} \notag\\
& \le 3\lb(\frac{\eps_j n}{6C_2 \delta_{j-1} k \cdot \log^{1/2} k}\rb)^{-\frac 12} \log^{\frac 12} n\notag\\
& \le 9\sqrt{C_2} \cdot \eps_j^{-\frac 12} \delta_{j-1}^{\frac 12} \cdot \lb(\frac kn\rb)^{\frac 12}\cdot \log^{\frac 14} k\cdot  \log^{\frac 12} n,\end{align}
where the last \ineq\ relies on $\delta_{j-1} k >1$.
By Definition \ref{def:delta_j}, we have 
\[\delta_{j-1}= 3\cdot 5^{j-2} \cdot \eps_0^{-2^{-(j-1)}} \cdot  \eps_1^{-2^{-(j-2)}} \cdots \eps_{j-1}^{-2^{-1}} \cdot  \lb(\frac kn\rb)^{1-2^{-(j-1)}} \log^{1+\frac{j-2}4} n.\]
Substituting this into \eqref{eqn:012423}, we conclude that 
\[\mu^* - \mu_{\min}\lb(S_{j-1}\rb)  
\le 3\cdot 5^{j-1} C_2^{\frac{j-1}2} \eps_0^{-2^{-j}} \eps_1^{-2^{-(j-1)}} \cdots \eps_j^{-2^{-1}} \lb(\frac kn\rb)^{1-2^{-j}} \log^{1+\frac{j-1}4} n
\le \delta_j. \eqno\Halmos\]

We now have the following regret bound for BSE \alg\ with adaptivity $\ell$.

\bprop[Bayesian Regret of BSE, Arbitrary Grid] \label{prop:gen_ell}
Consider an arbitrary adaptivity level $\ell$ and grid sizes $0< \eps_0< \dots< \eps_{\ell-1}<1$. 
\Sps\ the prior \distr\ $D$ \sats\ Assumption \ref{assu:avg} and that $\delta_i k > 1$ for $1\le i\le \ell-1$. Then,
\begin{align*}
\mathrm{BR}_n \lb({\rm BSE}_\ell \lb(\eps_0,\cdots, \eps_{\ell-1}\rb), D\rb) = \tilde O\left(\eps_0 + \eps_1 \delta_1 + \eps_2 \delta_2 +\cdots+\eps_{\ell-1} \delta_{\ell-1} + \delta_\ell + n^{-\rho} \right).
\end{align*}
\eprop
\proof{Proof.}
By the definition of internal regret,  
\begin{align}\label{eqn:122722d}
\mathrm{BR}_n \lb({\rm BSE}_\ell \lb(\eps_0,\cdots, \eps_{\ell-1}\rb), D\rb)
\le \sum_{j=0}^{\ell-1} \eps_j \cdot \ho{E} \lb[\mu^* - \mu_{\min}(S_j)\rb] + \ho{E} \lb[ \mu^* - \mu_{\min}\lb(S_\ell\rb)\rb].
\end{align}
If the events $\mathcal{C}$ and $U_{\delta_j}$ occur for all $j=0,\dots,\ell-1$, then by Lemma \ref{lem:layer_j_loss}, we have 
\[\mu^* - \mu_{\min}(S_j) \le \delta_j \quad \text{for } j=0,\dots,\ell-1.\] Furthermore, for all $j=0,\dots,\ell-1$, by Lemma \ref{lem:good_event}, for each $j$ we have
$\ho{P}\lb[U_{\delta_j}\rb] \le k^{-2}$ provided $\delta_j k \geq 1$. Thus,
\begin{align*} 
\ho{E}\lb[\mu^* - \mu_{\min}(S_j)\rb]  \le \delta_j \cdot \ho{P}\lb[\mathcal{C} \cap\bigcap_{1\le j\le \ell-1} U_{\delta_j}\rb] + 
\sum_{1\le j \le \ell-1} \ho{P}\lb[\overline{U_{\delta_j}}\rb] + \ho{P}\lb[\overline{\mathcal{C}}\rb]
\leq \delta_j \cdot 1 + n^{-2+\rho}\cdot \ell.
\end{align*}
Similarly,
\[\ho{E}\lb[\mu^* - \mu_{\min}(S_\ell)\rb]   \le \delta_{\ell} + k^{-2} w + n^{-2+\rho}w.\]
When $\rho \le 1$, the above is bounded by $\delta_\ell + n^{-2\rho}w +o\lb(\frac 1n\rb)$.
Combining, we conclude that
\begin{align*}
\mathrm{BR}_n \lb({\rm BSE}_\ell \lb(\eps_0,\cdots, \eps_{\ell-1}\rb),D\rb) 
& \leq \eps_0 + \sum_{j=1}^{\ell} \eps_j \lb(\delta_j + n^{-3}\rb)  + n^{-2\rho}w  + o\lb(\frac 1n\rb)\\
&\le \eps_0 + \sum_{j=1}^{\ell} \eps_j \delta_j + n^{-\rho} + o\lb(\frac 1n\rb). &\Halmos 
\end{align*}

\section{Proof of Theorem \ref{thm:ub}: the Main Upper Bound}\label{apdx:gen_ell}
To derive the upper bound for the SLHVB problem, we need to compare the \asym\ orders of the two terms in \eqref{eqn:loss}.
Consider the following three cases for $\rho$. 

\noindent{\bf Case 1.} \Sps\ $\rho \le \frac 15$. 
Consider $\ell=1$. 
By Proposition \ref{prop:loss_induced}, we have
\[{\rm Loss}_n \lb({\rm BSE}^\star (1, k), D\rb) = \tilde O\lb(n^{-\rho} + n^{(\rho-1)\cdot \frac 13}\rb).\]
When $\rho\le \frac 15$, it holds that $n^{-\rho}\ge n^{(\rho-1)\cdot \frac 13}$, so the above becomes $\tilde O\lb(n^{-\rho}\rb).$

\noindent{\bf Case 2.} \Sps\ $\frac 15<\rho\le \frac w{2w+2}$.
In this case, the integer $\ell$ in Step \ref{step:med} of \Alg\ \ref{alg:hybrid} exists
due to the following \elem ary fact: For any integer $w\geq 2$, it holds that
\begin{align}\label{eqn:011923}
\lb[\frac 15,\ \frac w{2w+2}\rb] \subseteq \bigcup_{2\leq \ell\le w} \lb[\frac{\ell-1}{2\ell+1},\ \frac \ell{2\ell+2}\rb].
\end{align}
Now that $\theta_\ell = \frac{\ell-1}{2\ell+1}\le \rho$, by Proposition\ref{prop:loss_induced} we have
\[\mathrm{Loss}_n \lb({\rm BSE}^\star\lb(\ell, k\rb), D\rb) = \tilde O\lb(n^{-\rho} + n^{(\rho-1)\cdot \frac \ell{\ell+2}}\rb).\]
Further, since $\rho < \frac \ell{2\ell+2}$, the first term dominates the second \asymly, i.e., $n^{-\rho} > n^{(\rho-1)\cdot \frac\ell{\ell+2}}$, and therefore the above becomes $\tilde O\lb(n^{-\rho}\rb)$.

\noindent{\bf Case 3.} 
\Sps\ $\rho\ge \frac w{2w+2}=\theta_w$.
Note that the threshold exponent $\theta_\ell$ increases in $\ell$, and so for any $\ell \le w$, we have $\theta_\ell \le \theta_w \le \rho$.
It then follows from Proposition \ref{prop:loss_induced} that
\begin{align}
\label{eqn:020923}
{\rm Loss}_n \lb({\rm BSE}^\star\lb(w,k\rb), D\rb) = \tilde O\lb(n^{-\rho} +n^{(\rho-1)\cdot \frac w{w+2}}\rb).
\end{align}
Although this is already sublinear in $n$, we observe that the two terms in \eqref{eqn:020923}  are, in general, not equal, which suggests a potential improvement. 
Consider the resampling size $k'=n^{\rho'}$ that makes these two terms equal, that is, $\rho' = \frac w{2w+2}$. Note that this choice is feasible since $\rho' \le \rho$.
Then, due to Proposition \ref{prop:translation} we obtain 
\begin{align*}
{\rm Loss}_n \lb({\rm BSE}^\star (w, k'), D\rb) &= \tilde O\lb(n^{-\rho'} + \rho^{-1} n^{-\rho} + n^{(\rho'-1)\cdot \frac w{w+2}}\rb)\\
&\le \Tilde{O} \lb(n^{-\frac w{2w+2}}  + \rho^{-1} \cdot n^{-\frac w{2w+2}} + n^{-\frac w{2w+2}}\rb) \\
&= \Tilde{O} \lb(\rho^{-1}\cdot n^{-\frac w{2w+2}} \rb).
\end{align*}
where the \ineq\ follows since $\rho' \le \rho.$ \hfill\Halmos

\section{Proof of Theorem \ref{thm:main_lb}: the First Lower Bound}
We present a formal proof of the main lower bound.
Consider the event \[W_{t-1}=\lb\{1-\frac 2{(w-1)k} \leq \mu_{\max}(A_{t-w}^{t-1}) \leq 1-\frac 1{(w-1)k}\rb\}.\]
We first show that this event occurs with large \prb.
\begin{lemma}\label{lem:wb}
If $k\ge 10$, then $\ho{P}(W_{t-1}) \geq \frac 18$.
\end{lemma}

\proof{Proof.} Denote $\mu_{\max} = \mu_{max}(A_{t-w}^{t-1})$. 
Consider the events $H = \lb\{\mu_{\max} \geq 1- \frac 1{(w-1)k}\rb\}$ and $H' = \lb\{\mu_{\max} < 1- \frac 2{(w-1)k}\rb\}.$
Then for any $k\ge 10$, we have \[\ho{P}\lb[\overline{H}\rb] = \ho{P}\lb[\mu_{\max} < 1-\frac 1{(w-1)k}\rb] = \lb(1-\frac 1{(w-1)k}\rb)^{(w-1)k} \geq \frac 34 \cdot e^{-1},\]
i.e., $\ho{P}[H]\le 1-\frac 3{4e}$. 
Moreover, since $\ho{P}[H'] = \lb( 1-\frac 2{(w-1)k}\rb)^{\frac{(w-1)k}2 \cdot 2} \leq e^{-2}$, we conclude that
\[\ho{P}[W_{t-1}] \ge 1- \ho{P}[H'] - \ho{P}[H] \geq 1-\lb(1-\frac 3{4e}\rb) - e^{-2} > \frac 18.\eqno\Halmos\]

Thus, by giving up a constant factor in the loss, we may restrict our attention to the event $W_{t-1}$.
For each time period $t$, define \prb\ measure $\ho{P}_t(\cdot)=\ho{P}[\cdot| W_{t-1}]$ and
$\ho{E}_t(\cdot) = \ho{E}[\cdot| W_{t-1}]$.
Consider the event that at time $t$, the policy selects arms from $A_t$ at least $n/2$ times, i.e., $\mathcal{E}_t = \left\{\sum_{a\in A_t} \pi_t(a) \geq \frac n2\right\}.$
Further, define
\[\mathcal{B}^-_t = \left\{\mu_{\max}(A_t) \leq \mu_{\max}(A_{t-w}^{t-1}) -\frac 1k\right\} \text{   and   }
\mathcal{B}^+_t = \left\{\mu_{\max}(A_t) \geq \mu_{\max}(A_{t-w}^{t-1}) + \frac 1{wk}\right\}.\]

Observe that
\begin{align}\label{eqn:feb4}
\ho{E}_t(L_t) &= \ho{E}_t(L_t| \mathcal{E}_t \cap \mathcal{B}^-_t) \cdot \ho{P}_t(\mathcal{E}_t \cap \mathcal{B}^-_t) + \ho{E}_t(L_t| \mathcal{E}_t \cap \mathcal{B}^+_t) \cdot \ho{P}_t(\mathcal{E}_t \cap \mathcal{B}^+_t)\notag\\
&\ +\ho{E}_t(L_t| \overline{\mathcal{E}}_t \cap \mathcal{B}^-_t) \cdot \ho{P}_t(\overline{\mathcal{E}}_t \cap \mathcal{B}^-_t) + \ho{E}_t(L_t| \overline{\mathcal{E}}_t \cap \mathcal{B}^+_t) \cdot \ho{P}_t(\overline{\mathcal{E}}_t \cap \mathcal{B}^+_t)\notag\\
&\geq \ho{E}_t[L_t | \mathcal{E}_t \cap \mathcal{B}^-_t]\cdot \ho{P}_t(\mathcal{E}_t \cap \mathcal{B}^-_t) + \ho{E}_t(L_t| \overline{\mathcal{E}}_t \cap \mathcal{B}^+_t)\cdot \ho{P}_t(\overline{\mathcal{E}}_t \cap \mathcal{B}^+_t).
\end{align}

We next lower bound the above two terms. 
Note that the event in the first term, i.e.,  $\mathcal{E}_t \cap \mathcal{B}^-_t$, says that the policy selects arms from $A_t$ for at least $n/2$ times, but the best arm in $A_t$ is suboptimal (w.r.t. $\mu_{\max} (A_{t-w}^t)$) by $1/k$.
Intuitively, when this event occurs, the policy suffers at least $\frac n2 \cdot \frac 1k = \frac n{2k}$ loss in this round.
\OTOH, the event in the other term, i.e., $\overline{\mathcal{E}}_t \cap \mathcal{B}^+_t$, says that the policy selects $A_t$ less than $n/2$ times, but the best arm in $\mu(A_{t-w}^{t-1})$ is suboptimal by $n/(wk)$.
Using a similar argument, we can show that the loss in this round is at least $\frac 1{2wk}$ on this event.
Next, we formalize this idea.

\begin{lemma}[Loss for Uncertainty in the New Batch]\label{claim:feb5b}
For any $t\ge 1$, we have \[\ho{E}_t \lb[L_t | \mathcal{E}_t \cap \mathcal{B}^-_t\rb] \geq \frac n{2k} \quad\text{and}\quad \ho{E}_t \lb[L_t| \overline{\mathcal{E}}_t \cap \mathcal{B}^+_t \rb] \geq \frac n{2wk}.\]
\end{lemma}
\proof{Proof.}
Write $\mu^*_t = \mu_{\max} (A_t)$.
Note that when $\mathcal{B}^-_t$ occurs, we have $\mu_t^* - \mu_{\max}(A_t) \geq 1/k$.
Thus, if $\pi$ selects arms from $A_t$ for $\Omg(n)$ times, an $\Omg\lb(n/k\rb)$ loss is incurred.
Formally, 
\begin{align}\label{eqn:060423}
\ho{E}_t \lb[L_t|\ \mathcal{E}_t\cap \mathcal{B}^-_t\rb]
&= \ho{E}_t \lb[\sum_{a\in A_{t-w}^t} \pi_t(a)\cdot (\mu^*_t - \mu_a)\ \Big|\ \mathcal{E}_t \cap \mathcal{B}^-_t \rb] \notag\\
&\geq \ho{E}_t \lb[\sum_{a\in A_t} \pi_t(a)\cdot (\mu^*_t - \mu_a)\ \Big|\ \mathcal{E}_t \cap \mathcal{B}^-_t\rb] \notag\\
&\geq  \ho{E}_t \lb[\sum_{a\in A_t}\pi_t(a)\cdot (\mu^*_t - \mu_{\max}(A_t))\ \Big|\ \mathcal{E}_t \cap \mathcal{B}^-_t\rb].
\end{align}
Note that conditional on $\mathcal{E}_t, \mathcal{B}^-_t$ and $W_{t-1}$ (recall that $\ho{E}_t[\cdot] = \ho{E}[\cdot | W_{t-1}]$), we have $\sum_{a\in A_t} \pi_t(a) \geq \frac n2$ and $\mu^*_t - \mu_{\max}(A_t)\geq 1/k$.
Thus, \[\eqref{eqn:060423}\geq \frac n2 \cdot \frac 1k=\frac n{2k}.\]

Now consider the second term in \eqref{eqn:feb4}. 
When $\mathcal{B}^+_t$ occurs, we have $\mu_{\max}(A_t)-\mu_{\max}(A_{t-w}^{t-1}) \geq \frac 1{wk}$, so if the 
$\pi$ selects arms in $A_{t-w}^{t-1}$ for $\Omg(n)$ times, an $\Omg(n/k)$ loss is incurred.
Formally, 
\begin{align}\label{eqn:060423b}
\ho{E}_t \lb[L_t|\ \overline{\mathcal{E}_t}\cap \mathcal{B}^+_t\rb]
&= \ho{E}_t \lb[\sum_{a\in A_{t-w}^t}\pi_t(a)\cdot (\mu^*_t - \mu_a)\ \Big |\ \overline{\mathcal{E}_t}\cap \mathcal{B}^+_t\rb] \notag\\
&\ge \ho{E}_t \lb[\sum_{a\in A_{t-w}^{t-1}} \pi_t(a)\cdot \lb(\mu_{\max}\lb(A_t\rb) - \mu_{\max}(A_{t-w}^{t-1})\rb)\ \Big|\ \overline{\mathcal{E}_t}\cap \mathcal{B}^+_t\rb].
\end{align}

Note that conditional on $\overline{\mathcal{E}_t}, \mathcal{B}^-_t$ and $W_{t-1}$, we have $\sum_{a\in A_{t-w}^{t-1}} \pi_t(a)\geq \frac n2$ and $A_t - \mu_{\max}(A_{t-w}^{t-1})\geq \frac1{wk}$. 
Therefore, \[\eqref{eqn:060423b}\ge \frac n2 \cdot \frac 1{wk} = \frac n{2wk}.\eqno \Halmos\]

Now we are ready to show the $\Omg(1/k)$ lower bound.

\noindent{\bf Proof of Proposition \ref{thm:lb1}.} 
By \eqref{eqn:feb4} and Lemma \ref{claim:feb5b}, we have 
\begin{align}\label{eqn:060523}
\ho{E}_t(L_t) &\ge \ho{E}_t[L_t | \mathcal{E}_t \cap \mathcal{B}^-_t]\cdot \ho{P}_t(\mathcal{E}_t \cap \mathcal{B}^-_t) + \ho{E}_t(L_t| \overline{\mathcal{E}}_t \cap \mathcal{B}^+_t)\cdot \ho{P}_t(\overline{\mathcal{E}}_t \cap \mathcal{B}^+_t)\notag\\
&\ge \frac n{2k} \cdot  \ho{P}_t(\mathcal{E}_t \cap \mathcal{B}^-_t) + \frac n{2wk} \cdot \ho{P}_t (\overline{\mathcal{E}_t}\cap \mathcal{B}^+_t).
\end{align}
Note that the events $\mathcal{E}_t$ and $\mathcal{B}_t^+$ ($\overline{\mathcal{E}_t}$ and $\mathcal{B}_t^-$ resp.) are \indep\
conditional on $G_{t-1}$, so 
\[\eqref{eqn:060523} \ge \frac n{2wk} \cdot \lb(\frac 12 \cdot \ho{P}_t (\mathcal{E}_t) + \frac 1w \cdot \ho{P}_t \lb(\overline{\mathcal{E}_t} \rb)\rb) \geq \frac n{2w^2 k}.\]
Therefore, 
\begin{align*}
\ho{E} [L_t] &\geq \ho{E}[L_t \cdot {\bf 1}(G_{t-1})]  = \ho{E}[L_t | G_{t-1}] \cdot \ho{P}[G_{t-1}]\geq \frac n{2w^2 k} \cdot \frac 18
= \frac n{16w^2 k}.
\end{align*}
Summing over $t$ and taking the limit, we conclude that
\[\mathrm{Loss}_n (\pi) = \varlimsup_{T\rar \infty} \frac 1{nT}\sum_{t=1}^T \ho{E}[L_t] \geq \frac 1{16w^2 k}.\eqno\Halmos
\]

\section{Proof of Proposition \ref{thm:lb2}: The Second Lower Bound}
\label{apdx:n^-1/2_lb}

Consider the event $G_t = \lb\{\mu_{\max}(A_{t-w}^t) \geq 1-n^{-1/2} \rb\}$ where $t \geq w$.
Recall by Assumption \ref{assu:avg} that the density $f$ of $D$ is bounded as $0< C_1\le f(x) \le C_2$. 

\begin{lemma}[$\mu_{\max}$ is Close to $1$]\label{lem:close_to_1}
If $C_2 k>\sqrt n$, then $\ho{P}[G_t]\geq \frac 12.$
\end{lemma}
\proof{Proof.} Write $\delta = \frac 1 {\sqrt n}$. Note that $|A_{t-w}^t|=wk$, so 
\[\ho{P}\lb[\overline{G_t}\rb] \le (1 - C_2 \delta)^{kw} = (1-C_2\delta)^{\frac 1{C_2 \delta} \cdot C_2 kw\delta}\le e^{-C_2 kw\delta}.\]
Since $C_2 k>\sqrt n$, we have
$C_2 k\delta > n^{\frac 12}\cdot n^{-\frac 12} \ge 1$, so
\[\ho{P}[\overline{G_t}] \leq e^{-C_2 kw\delta} \leq e^{-w} \leq \frac 12.\eqno\Halmos\]


An arm is said to be {\it unexplored} at time $t$ if it has never been selected by the policy before.
Our proof considers the number of unexplored arms selected in each round, as formalized below.

\bdefn[Unexplored Arms]
We say an arm $a$ is {\it unexplored} at round $t$ if $\sum_{\tau=1}^{t-1} \pi_\tau (a) = 0$.
For each round $t$, define $M_t$ as the subset of unexplored arms selected in this round; formally, $M_t = \{a\in A_{t-w}^t:\pi_t (a) >0 \text{ and } \sum_{s=1}^{t-1} \pi_s (a) =0\}.$
Furthermore, as for any $t,t'$, we write $M_t^{t'} = \bigcup_{\tau=t}^{t'} M_\tau$. 
\edefn

From a myopic perspective, selecting too many unexplored arms in a round leads to high loss since, by Assumption \ref{assu:avg}, each unexplored arm is suboptimal by $\Omg(1)$  on \avg.
We formalize this in the following lemma. 
Recall that $L_t$ is the loss in round $t$.
\begin{lemma}[Cost of Selecting Many Unexplored Arms]\label{lem:apr13}
\Sps\ $C_2 k>\sqrt n$. For any $t\geq 1$, consider the event $V_t = \lb\{|M_t| \geq \frac {\sqrt n}{4w}\rb\}$. 
Then, \[\ho{E}\lb[L_t|V_t\rb] \geq \frac {\sqrt n}{24w}.\]
\end{lemma}
\proof{Proof.} Write $\mu_t^* = \mu_{\max}(A_{t-w}^t)$ and $\ho{E}_t[\cdot] = \ho{E}[\cdot | G_t]$ for any $t$.
We start with lower bounding the conditional loss. 
Observe that 
\begin{align}\label{eqn:apr14}
\ho{E}_t \lb[L_t | V_t\rb] &= \ho{E}_t \lb[\sum_{a\in A} \pi_t (a)\cdot (\mu^*_t - \mu_a)\Big|\ V_t\rb] \notag\\
&\ge \sum_{a\in A} \ho{E}_t \lb[\pi_t (a)\cdot {\bf 1}(a\in M_t) \cdot (\mu^*_t - \mu_a) \Big|\ V_t\rb] \notag\\
&\ge \sum_{a\in A} \ho{E}_t \lb[{\bf 1}(a\in M_t)\cdot \lb(1- \frac 1{\sqrt n} - \mu_a\rb)\Big|\ V_t\rb],
\end{align}
where the last \ineq\ follows by definition of $\ho{E}_t$ and that $a\in M_t$ implies $\pi_t (a)\ge 1$.
To further simplify, note that the mean reward of any unexplored arm $a$ is independent of $M_t$, so 
\begin{align}\label{eqn:121022}
\eqref{eqn:apr14} &= \sum_{a\in A} \ho{E}_t \lb[{\bf 1}(a\in M_t)|\  V_t \rb] \cdot \ho{E}_t \lb[1- \frac 1{\sqrt n} - \mu_a |\ V_t \rb].
\end{align}
Note that $\ho{E}_t [\mu_a | V_t] = \ho{E}[\mu_a] =\frac 12,$ 
we have \[\ho{E}_t \lb[1- \frac 1{\sqrt n} - \mu_a \Big|\ V_t \rb] = 1- \frac 1{\sqrt n} -\frac 12 \ge \frac 13\] 
for any $n\geq 1$. Therefore, 
\[\eqref{eqn:121022} 
\geq \ho{E}_t \lb[\sum_{a\in A} {\bf 1}(a\in M_t) \Big |\ V_\tau\rb] \cdot 
\frac 13 = \ho{E}_t \lb[|M_t| \Big | V_\tau\rb] \cdot \frac 13
\geq \frac {\sqrt n}{12w}.\]
By Lemma \ref{lem:close_to_1}, we have $\ho{P}[G_t]\geq \frac 12$, and thus  
\[\ho{E} [L_t|V_t] \geq \ho{E}_t [L_t |V_t]\cdot \ho{P}[G_t] \geq \frac 12 \cdot \frac {\sqrt n}{12w} = \frac{\sqrt n}{24w}.\eqno\Halmos\]

We have just seen that selecting too many unexplored arms leads to high loss. 
On the other hand, selecting too few unexplored arms also leads to a high loss.  
To show this, consider the {\it cold-start} event where no $\delta$-good arms that $\pi$ ever selected are still available at the start of round $t$.

\bdefn[Cold-start Event]
For each $t\geq 1$, the {\it cold-start event} is defined as 
\[B_t = \lb\{\mu_{\max}(M_{t-w}^t) \leq 1-n^{-1/2}\rb\}.\]
\edefn

As the name suggests, when this event occurs, the policy has little information about the available arms, and therefore behaves as if the time horizon has {\em restarted}.
This leads to high loss. 
In fact, a policy has to identify a $\delta$-good arm to have a low loss from time $t-w$ to $t+w$. 
This forces the policy to explore $\Omg(1/\delta) =  \Omg(n^{1/2})$ unexplored arms, leading to a loss of $\Omg(n^{1/2})$.
This result is implied by the lower bound for {\it infinite-armed bandits} \citep{wang2008algorithms}.

\begin{lemma}[Cold-start Incurs High Loss]\label{lem:feb6}
For any $t$, it holds that
\[\ho{E}\lb[\sum_{\tau=t}^{t+w} L_\tau \Big| B_t\rb] \geq \frac{\sqrt n}{6w}.\]
\end{lemma}

To apply the above, we next \kehua\ when $B_t$ would occur. 
Consider the number $N_t$ of new arms selected in $[t-w,t]$. 
If $\ho{E}N_t\ge 1/\delta$, then on \avg\ the policy suffers $\Omg(1/\delta)=\Omg(\sqrt n)$ loss. 
If $\ho{E}N_t < 1/\delta$, then with $\Omg(1)$ \prb, {\bf none} of those $N_t$ arms are $\delta$-good, which leads to the cold-start event $B_t$. 
We formalize this idea below.

\begin{lemma}[Underexploration Leads to Cold-start]
\label{lem:121222}
If $\ho{E}\lb[\sum_{s=t-w}^{t} L_s \rb] \leq \frac{\sqrt n}{96 w^2}$, then $\ho{P}[B_t]\geq \frac 12$.
\end{lemma}
\proof{Proof.}
Consider the event $E_s = \lb\{\mu_{\max}(S_s)\geq 1-\frac 1{\sqrt n}\rb\}$ that
for any $s=1,2\dots$, then our goal reduces to showing that $\ho{P}[E_s]\leq \frac 1{2w}$ for every $s=t-w,\dots, t$.
In fact, if none of the events $E_{t-w},\dots,E_t$ occurs, i.e. no unexplored arm selected in the past $w$ rounds is $n^{-1/2}$-good, then $B_t$ occurs. 
It then follows from the  union bound that
\[\ho{P}\lb[B_t\rb] \geq \ho{P}\lb[\bigcap_{\tau=t-w}^t \overline{E_\tau} \rb] = 1-\ho{P}\lb[\bigcup_{\tau=t-w}^t E_\tau\rb] 
\geq 1- w\cdot \frac 1{2w} = \frac 12,\]
and the proof will be complete.

To show $\ho{P}[E_s]\leq \frac 1{2w}$ for $s\in \{t-w,...,t\}$, observe that for any fixed $s$,  
\begin{align}\label{eqn:apr27}
\ho{P}[E_s] = \ho{P}[E_s|V_s]\cdot \ho{P}[V_s]+ \ho{P}[E_s|\overline{ V_s}] \cdot \ho{P} [\overline{ V_s}] \leq \ho{P}[V_s] + \ho{P}[E_s|\overline{ V_s}],
\end{align}
where we recall that
$V_t = \lb\{|M_t| \geq \frac {\sqrt n}{4w}\rb\}$.
we will bound each of the two terms in \eqref{eqn:apr27}
by $\frac 1{4w}$ \sep ely. 
To see
$\ho{P}[V_s]\leq \frac 1{4w}$, note that 
\[\frac{\sqrt n}{96 w^2} \geq \sum_{t'=t-w}^t \ho{E}[L_{t'}] \geq \ho{E}[L_s] \geq  \ho{E}[L_s|V_s]\cdot \ho{P}[V_s].\] 
Note that  by Lemma~\ref{lem:apr13}, we have
$\ho{E}[L_s|V_s] \geq \frac{\sqrt{n}}{24w}$, so $P[V_s]\leq \frac1{4w}$.

To bound the second term in \eqref{eqn:apr27}, note that by definition of $E_s$, for any $C,\delta>0$ we have 
\begin{align*}
\ho{P}\lb[\overline{E}_s \Big| S_s\leq C \rb] \geq (1-\delta)^C \geq (1-\delta)^{\frac 1\delta\cdot \delta C} \geq e^{-\delta C}\geq 1-\delta C,
\end{align*}
where the last \ineq\ follows since $e^{-x} \geq 1-x$ for any $x\in \real$. 
In \parti, for $C = \frac {\sqrt n}{4w}$ and $\delta = \frac 1{\sqrt n}$, we have 
\[
\ho{P}\lb[\overline{E_s} \Big| \overline{V_s}\rb] = \ho{P}\lb[\overline{E_s} \Big|S_s \leq \frac{\sqrt n}{4w}\rb]
\geq 1- \frac 1{\sqrt n} \cdot \frac {\sqrt n}{4w} = 1-\frac 1{4w},\]
i.e.,
\[\ho{P}\lb[E_s \Big| \overline{V_s} \rb] 
=1-\ho{P}\lb[\overline{E}_s \Big| \overline{V_s}\rb] \leq \frac 1{4w}.\]
\hfill
\hfill\Halmos

We now combine the above to establish the $\sqrt n$ lower bound.

\noindent{\bf Proof of Proposition \ref{thm:lb2}.}
Fix any round $t$. 
Decompose  $L_{t-w}^{t+w}$ into the loss before and after round $t$ as 
\[L_{t-w}^{t+w}= \sum_{\tau=t-w}^{t+w} \ho{E}L_\tau = \sum_{\tau=t-w}^{t-1} \ho{E}L_\tau + \sum_{\tau=t}^{t+w} \ho{E}L_\tau.\]
If the first term, i.e., the loss before time $t$, is greater than $\frac {\sqrt n}{96 w^2}$, then the claim holds trivially.
\Ow, by Lemma~\ref{lem:121222}, the cold-start event $B_t$ occurs w.p. $\ho{P}[B_t]\geq \frac 12$. 
Thus, the loss after time $t$ can be bounded using Lemma~\ref{lem:feb6} as
\[\sum_{\tau=t}^{t+w} \ho{E}L_\tau \geq \ho{E}\lb[\sum_{\tau=t}^{t+w} L_\tau \Big| B_t\rb] \cdot \ho{P}[B_t]\geq \frac{\sqrt n}{6w} \cdot \frac 12 = \frac {\sqrt n}{12w}> \frac {\sqrt n}{96 w^2}.\]
\hfill
\Halmos

\section{Field Experiment: Implementation Details}\label{apdx:details_field_xp}
In this section, we provide more details of the field experiment.
\subsection{Randomized BSE: A Thompson Sampling Variant}

We implemented a variant of BSE-induced policy with $\ell=1$, which was modified due to the following practical concerns.
The first issue is the lack of knowledge of $n$, which corresponds to the number of impressions in a time period (an hour).
This can be easily fixed using randomization: For each impression, with \prb\ $\eps_0$ we assign a newly arriving card (for {\it exploration}) and with \prb\ $1-\eps_0$ we assign an old card (for {\it exploitation}).

\begin{algorithm}
\caption{SetPrior($g$): Fit a Beta Prior Using DNN Output}
\label{alg:set_prior}
\begin{algorithmic}[1]
\State Input: a card $a$ and $m$ users
\State Output: $\hat \alpha,\hat \beta$ 
\State{Randomly sample $m$ users $u_1,\dots, u_m$}
\For{$i=1,\dots, m$}
\State $\mu(u_i,a)\lar$ DNN-predicted reward on $(u_i,a)$
\EndFor
\State Compute the sample mean $\bar \mu$ and variance $\bar v$:
\[\bar \mu \lar \frac1m \sum_{i=1}^m \mu(u_i,a),\quad \bar v \lar \frac 1{m-1}\sum_{i=1}^m \lb (\mu(u_i,a)-\bar \mu\rb)^2\]
\State Return 
\[\hat \alpha\lar \bar \mu \left(\frac{\bar \mu (1-\bar \mu)}{\bar v} -1\right) \quad \text{ and } \quad \hat \beta \lar \frac{1-\bar \mu}{\bar \mu} \hat \alpha\]
\end{algorithmic}
\end{algorithm}

The second issue is the {\it prior information} for the rewards rates of the newly arriving cards.
Although the DNN predictions are sometimes inaccurate, they provide useful information. 
To utilize such information, we fit a Beta prior \distr\ for each card's mean reward using the {\it method of moments} (see, e.g.,  Section 9.2 of \citealt{wasserman2006all}) based on the DNN predictions (see \Alg\ \ref{alg:set_prior}).

Specifically, recall that the DNN returns a predicted conversion rate for each {\em pair} of user and card.
For a fixed card, denote by $\bar \mu,\bar v$ the mean and variance of the predicted conversion rates on $m=500$ randomly sampled users.
The fitted Beta prior $\rm{B}(\hat\alpha, \hat \beta)$ is then given by  \[\hat\alpha = \bar \mu \left(\frac{\bar \mu (1-\bar \mu)}{\bar v} -1\right) \text{\quad and \quad} \hat \beta = \frac{1-\bar \mu}{\bar \mu}\hat \alpha.\]

The final issue is the recommended contents' {\it diversity}. 
Recall that the basic version of the BSE policy selects the \emp ly best arm to ``exploit''.
However, this is not reasonable in a practical setting where such an extreme promotion of a single card is undesirable. 
We thus consider the Thompson Sampling version of the BSE policy under a Beta-Bernoulli reward model.
Specifically, for each slot we draw a score for each card from its posterior. 
Then we assign to this slot the card with the highest score.
Note that the posterior can be efficiently updated using the Bayesian rule, since the Beta distribution is a conjugate prior for the Bernoulli \distr.
We provide a detailed pseudo-code in \Alg\ \ref{alg:ts}.

\begin{algorithm}[h]
\caption{\bf{Randomized Batched Successive Elimination}}
\label{alg:ts}
\begin{algorithmic}[1]
\State{Input:}
\State{\quad $\eps\in [0,1]$}\Comment{Probability for forced exploration}
\State{\quad $\theta\ge 0$} \Comment{Threshold for being well-explored}
\For{each hour $t=1,2,...$} 
\State Receive a set $A_{\rm new}$ of new cards 
\State Update the set $A$ of available cards 
\For{each card $a\in A$}
\If{$a\in A_{\rm new}$}
\State $(\alpha_a,\beta_a)\lar \mathrm{SetPrior}(a)$.
\Comment{Compute a prior reward \distr}
\Else
\State $n_a\lar$ number of interactions of $a$ in the last hour
\State $h_a\lar$ number of conversions of $a$ in the last hour 
\State $\alpha_a \lar \alpha_a + h_a,\ \beta_a\lar \beta_a + n_a- h_a$
\Comment{Update the posterior \distr}
\EndIf
\EndFor
\State{$A_{\rm well}\lar \{a:\alpha_a + \beta_a >\theta\}$} \Comment{Well-explored cards}
\For{each user $u$}
\Comment{Thompson Sampling step}
\State Receive the number $r_u$ of cards requested by $u$
\State{$A_u\lar$ cards that have never been assigned to $u$}\Comment{Firm's practical constraint}
\For{each card $a\in A_u$} \Comment{Sample from the posterior}
\State Draw $X_{u,a}\sim \mathrm{Beta}(\alpha_a, \beta_a)$ 
\EndFor
\State Sort $A_u \bigcap A_{\rm well}$ by $X_{u,a}$ in non-increasing order  as $a_1,a_2,\dots$\Comment{Maintain two queues}
\State Sort $A_u \bs A_{\rm well}$ by $X_{u,a}$ in non-increasing order as $a_1',a_2',\dots$
\State{$i,i' \lar 1$}
\For{$j=1,\dots,r_u$}
\Comment{Select and rank the cards}
\State{$Z_j \lar \mathrm{Ber}(\eps)$}
\Comment{Decide whether to explore or exploit}
\If{$Z_j=1$} 
\Comment{Exploitation}
\State{$S_j \lar a_i; \quad i\lar i+1$}
\Else{
\Comment{Exploration}
\State{$S_j\lar a_{i'}'; \quad i'\lar i'+1$}
}
\EndIf 
\State Send cards $\{S_j: j=1,...,r_u\}$
\EndFor
\EndFor
\EndFor
\end{algorithmic}
\end{algorithm}
}

\section{Field Experiment Results: Engaged Users}\label{apdx:field_xp_engaged}
In this subsection, we analyze the engagement of special subsets of {\em engaged} users.
A user is said to be {\it engaged} in month $m\in \{$May, July\} if she has at least $k$ click-throughs in month $m$. 

We first consider the retention rate from May to July. 
The retention rate (for the NN and MAB group \resp) is defined as the proportion of engaged users in May that are still engaged in July.
Although there is no obvious choice for the threshold $k$, Figure~\ref{fig:attrition} shows that for each $k=1,2,...,10$, the attrition rates of the MAB group are consistently higher than those of the NN group. 
This suggests that the choice of $k$ is not essential.
Thus, in the remaining analysis we will fix this threshold as $k=4$ and repeat the analysis in Section~\ref{sec:field_xp_analysis}.

By comparing Table~\ref{tab:overall_stat} and Table \ref{tab:overall_stat_engaged}, we observed that for engaged users, both the per-user-per-day duration and click-throughs are notably higher than the average over all users, indicating that our definition for ``engaged'' user indeed captures the enthusiasm of users.
As another noteworthy observation, for each of the four DID regressions in Table~\ref{tab:DID} and \ref{tab:DID_engaged}, the coefficient $\beta_3$ for the magnitude of the composite variable is higher for the engaged users than for all users.
This suggests that the treatment effect is even more significant for the engaged users. 

\begin{table}[]
\centering
\caption{Overall Statistics For Engaged Users}
\begin{tabular}{lllllll}
 & & & \multicolumn{2}{c}{May} & \multicolumn{2}{c}{July} \\ \cline{4-7}
 & & & NN         & MAB        & NN         & MAB         \\ \hline
\multirow{5}{*}{Per User-Day}   & \multirow{3}{*}{Duration} & Mean    & 404.707 & 403.228 & 303.002 & 314.272 \\ \cline{3-7} 
& & SE Mean &2.142 & 2.010 & 2.670 & 2.496\\ \cline{3-7} 
& & Median  & 196.970 & 199.845 & 133.348 & 146.230 \\ \cline{2-7} 
& \multirow{2}{*}{\#CT} & Mean & 4.235 & 4.236 & 2.657 & 2.866\\ \cline{3-7} 
& & SE Mean & 3.241e-02 & 3.080e-02 & 3.642e-0s2 & 3.661e-02\\ \hline
\multirow{5}{*}{Per Impression} & \multirow{3}{*}{Duration} & Mean & 3.790 & 3.866 &3.659 & 3.850 \\ \cline{3-7} 
 & & SE Mean & 5.922e-03 & 5.802e-03 & 1.022e-02 & 1.076e-02 \\ \cline{3-7} 
 & & Median  & 0.619 & 0.622 & 0.594 & 0.598 \\ \cline{2-7} 
& \multirow{2}{*}{CTR} & Mean    & 3.954e-02 & 4.043e-02 & 3.198e-02 & 3.498e-02 \\ \cline{3-7} 
 & & SE Mean & 6.802e-05 & 6.671e-05 & 1.074e-05 & 1.062e-05 \\ \hline
\end{tabular}
\label{tab:overall_stat_engaged}
\end{table}

\begin{table}[]
\centering
\caption{Significance Testing For Engaged Users}
\begin{tabular}{llllll}
& & \multicolumn{2}{c}{Basic} & \multicolumn{2}{c}{Bootstrap} \\ \cline{3-6} 
& & \multicolumn{1}{c}{$Z$-score} & \multicolumn{1}{c}{$p$-value} & \multicolumn{1}{c}{$Z$-score} & \multicolumn{1}{c}{$p$-value} \\ \hline
\multirow{2}{*}{Per User-Day}   & Duration & 2.719 & 3.273e-03 & 2.717 & 3.289e-03 \\ \cline{2-6} 
 & \#CT &3.056  & 1.121e-02 & 3.064 & 1.089e-02 \\ \hline
\multirow{2}{*}{Per Impression} & Duration & 6.996 &  1.321e-12 & 7.060  & 8.301e-13 \\ \cline{2-6}
& CTR & 11.626 &1.513e-31 & 11.478 & 8.482e-31 \\ \hline
\end{tabular}
\end{table}

\begin{figure}
\centering
\begin{minipage}{.45\textwidth}
  \centering
  \includegraphics[width=\linewidth]{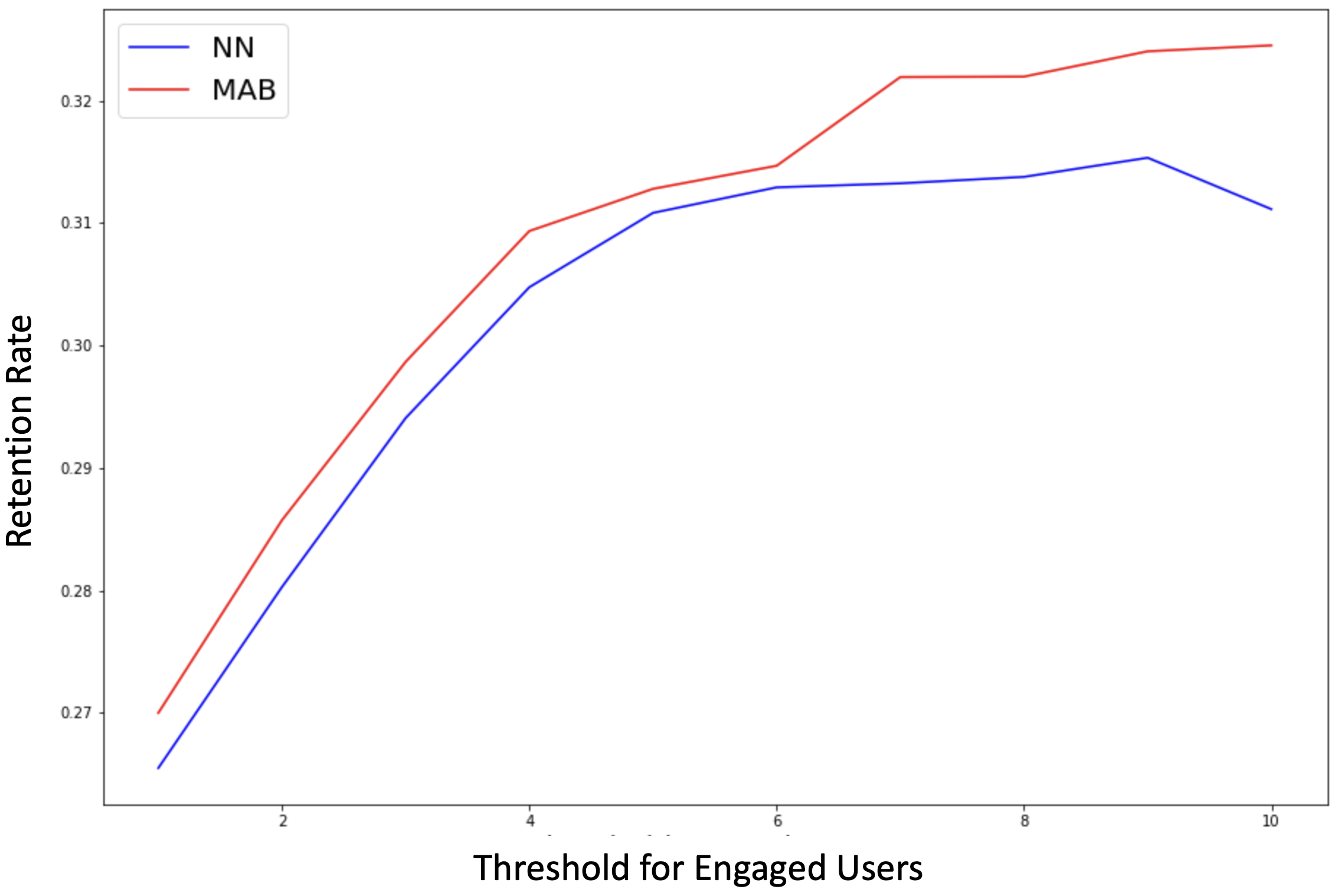}
  \caption{Attrition rates of the MAB and NN group.}
  \label{fig:attrition}
\end{minipage}%
\begin{minipage}{.45\textwidth}
  \centering
\includegraphics[width=\linewidth]{fig/duration_user_day.png}
  \caption{Duration per user-day pair.}
  \label{fig:test2}
\end{minipage}
\end{figure}

\begin{table}[]
\caption{Difference-In-Differences Regression for Engaged Users}
\label{tab:DID_engaged}
\begin{threeparttable}
\begin{tabular}{lllcccccc}
                 & &           & Coef. & Std. Dev. & $t$ & $p$-value & 0.025Q & 0.975Q \\ \hline
\multirow{8}{*}{Per User-Day}   & \multirow{4}{*}{Duration} & $\beta_0$ & 404.7074 & 2.008 & 201.569 & 0.000 & 400.772 & 408.643\\ \cline{3-9} 
 & & $\beta_1$ &  -101.7057 & 3.692 & -27.548 & 2.338e-167 & -108.942 & -94.470\\ \cline{3-9} 
 & & $\beta_2$ &  -1.4791 & 2.782 & -0.532 & 0.595 & -6.932 & 3.974\\ \cline{3-9} 
 & & $\beta_3$ & 12.7489 & 5.083 & 2.508 & 0.012 & 2.786 & 22.711 \\ \cline{2-9} 
 & \multirow{4}{*}{CT}       & $\beta_0$ & 4.2353 & 0.030 & 140.602 & 0.000 & 4.176 & 4.294 \\ \cline{3-9} 
 &                           & $\beta_1$ & -1.5787 & 0.055 & -28.502 & 5.532e-179 & -1.687 & -1.470\\ \cline{3-9} 
 &                           & $\beta_2$ & 0.0006 & 0.042 & 0.015 & 0.988 & -0.081 & 0.082\\ \cline{3-9} 
 &                           & $\beta_3$ & 0.2086 & 0.076 & 2.736 & 0.006 & 0.059 & 0.358\\ \hline
\multirow{8}{*}{Per Impression} & \multirow{4}{*}{Duration} & $\beta_0$ &  3.7789 & 0.006 & 634.095 & 0.000 & 3.767 & 3.791 \\ \cline{3-9} 
 & & $\beta_1$ &  -0.131 & 0.012 & -10.854 & 9.544e-28 & -0.154 & -0.107  \\ \cline{3-9} 
& & $\beta_2$ &  0.0723 & 0.008 & 8.708 & 1.546e-18 & 0.056 & 0.089  \\ \cline{3-9} 
& & $\beta_3$ & 0.1162 & 0.017 & 6.998 & 1.298e-12 & 0.084 & 0.149 \\ \cline{2-9} 
& \multirow{4}{*}{CT} & $\beta_0$ & -3.190 & 0.002 & -1771.359 & 0.000 & -3.193 & -3.186\\ \cline{3-9} 
& & $\beta_1$ &  -0.220 & 0.004 & -55.949 & 0.000 & -0.228 & -0.212  \\ \cline{3-9} 
 & & $\beta_2$ &   0.0237 & 0.002 & 9.500 &  1.049e-21 & 0.019 & 0.029  \\ \cline{3-9} 
 & & $\beta_3$ & 0.0691 & 0.005 & 12.942 & 1.303e-38 & 0.059 & 0.080 \\ \hline
\end{tabular}
    \begin{tablenotes}
      \small
      \item Note: As in the previous section, all regression are linear regression except for per impression CT, where we applied logistic regression due to binary labels.
    \end{tablenotes}
  \end{threeparttable}
\end{table}

\section{Offline Simulations}\label{sec:offline_simulations}

In this section, we conduct a more in-depth exploration of the practical performance of the BSE-induced policy through offline simulations, using synthetic data and also real user-interaction data from the platform.
Our findings align with our intuition and theoretical analysis, demonstrating that the performance of the policy is improved when (i) there are more layers involved, as discussed in Section \ref{subsec:impact_of_ell}, and
(ii) there is prior information (predictions provided by the DNN) available, as outlined in Section \ref{subsec:warm_start}. 

\subsection{Impact of the Number of Levels}
\label{subsec:impact_of_ell}
Theoretically, the performance of the BSE-induced policy improves with the number $\ell$ of levels. 
In Theorem~\ref{thm:ub}, we have established that the level-$\ell$ BSE policy has a loss of $\tilde O(\frac 1k + \left(\frac{k}{n}\right)^{\frac{\ell}{\ell+2}})$ under the revised geometric grid. 
Apparently, for fixed $n$ and $k$, this bound decreases in $\ell$.

\begin{figure}
\centering
\begin{minipage}{.5\textwidth}
  \centering
  \includegraphics[width=\linewidth]{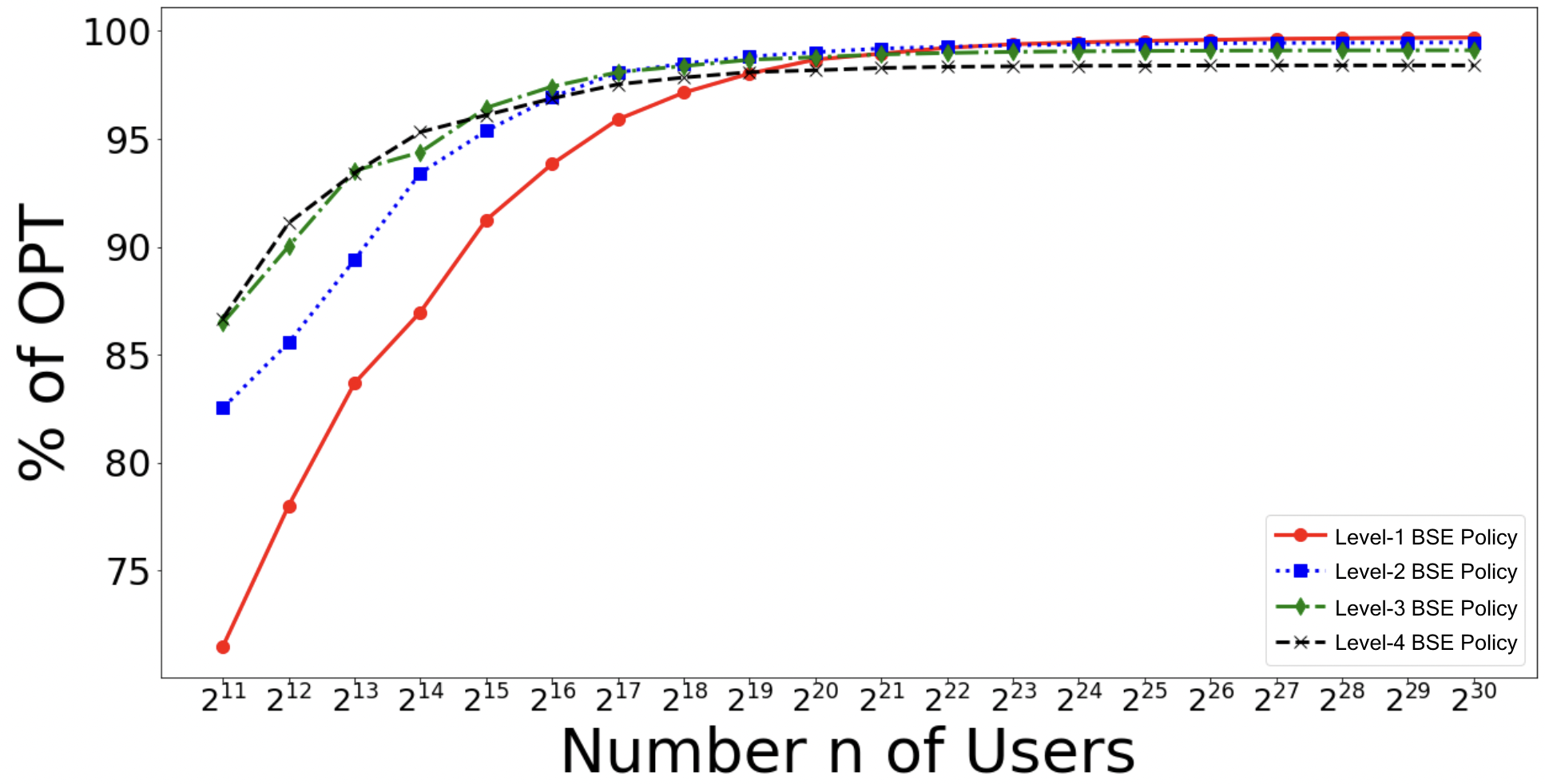}
  \captionof{figure}{Synthetic Simulation: $k=100$}
  \label{fig:purely_syn1}
\end{minipage}%
\begin{minipage}{.5\textwidth}
  \centering
\includegraphics[width=\linewidth]{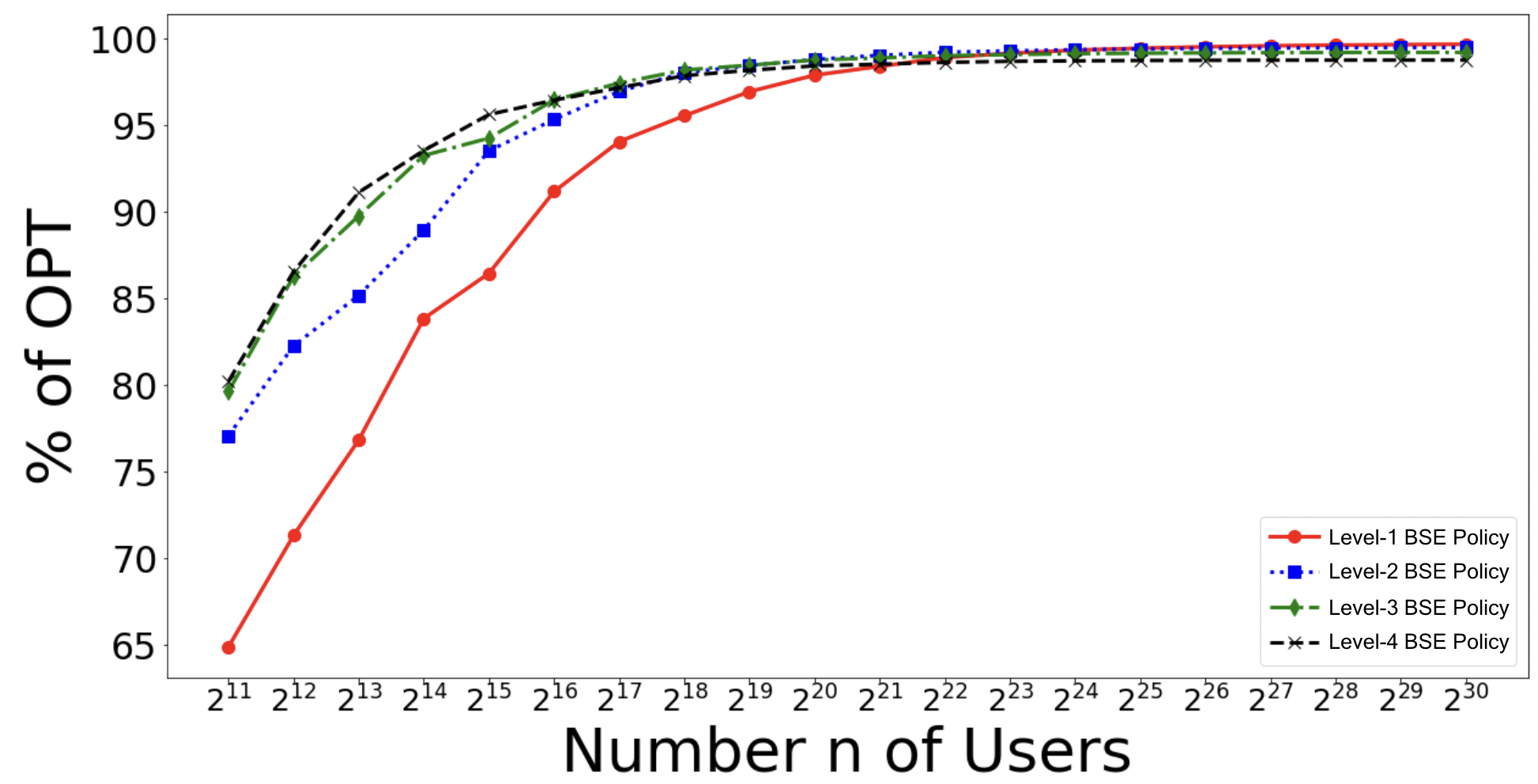}
\captionof{figure}{Synthetic Simulation: $k=200$}
  \label{fig:purely_syn2}
\end{minipage}
\end{figure}

We first validate this relationship via a purely synthetic simulation. 
We fix $k$ to be $100$ or $200$ and let $T$ vary from $2^{11}$ to $2^{30}$.
Specifically, in each round $t=1,\dots,500$, we defined a set of $k$ arms whose rewards are \rand ly generated from the uniform \distr\ on  $[0,1]$.
The lifetime of all arms are chosen to be $5$.
For each pair of $n,k$, we implemented the level-$\ell$ BSE policy where $\ell$ ranges from $1$ to $4$.

We computed the percentage reward of our policies relative to the optimal policy that possesses full knowledge of the reward rates; see Figure \ref{fig:purely_syn1} and Figure \ref{fig:purely_syn2}.
We notice that for every $\ell=1,\ldots,4$, the performance improves as $n$ increases. 
Additionally, in line with our theoretical analysis, the empirical performance of the policy deteriorates as $k$ increases for any fixed $\ell$ and $n$.
The underlying intuition here is that as $k$ increases, each arm is assigned to a smaller number of users, leading to an increase in learning error and consequently a decline in performance.



Most importantly, these experiments underscore the importance of having multiple layers. For small values of $n$ (less than approximately $2^{14}$), the level-$3$ and level-$4$ BSE policies demonstrate a significant performance advantage over their level-$1$ and $2$ counterparts.
The underlying rationale is that when $n$ is small relative to $k$, the level-$3$ and level-$4$ BSE policies are more effective in allocating slots for exploration. 
In contrast, as $n$ increases, the level-$1$ and level-$2$ policies perform slightly better. In such situations, the level-$3$ and level-$4$ policies allocate excessive resources to exploration, thereby missing out on the opportunity for exploitation and potential rewards from well-explored arms.

To complement the synthetic experiment, we replicated this simulation using the platform's real interaction data. 
In this setup, the arrival rates and lifetimes of the arms are no longer uniform. 
Consequently, we consider a variation of the BSE-induced policy, where the number of layers adjusts based on the number of arrivals in recent rounds.
More precisely, suppose that $k_t$ arms arrive in each round $t$ and this is known to the policy. 
Then, at time $t$, the policy selects $\eps_j = \left(\frac{k_{t-j}}n\right)^{\frac \ell{\ell+2}}$ fraction of the slots to surviving arms in $S_j$.
The same observations remain valid in this more realistic setting, as shown in Figure \ref{fig:semi_synthetic}.

\begin{figure}
\centering
\begin{minipage}{.5\textwidth}
  \centering
\includegraphics[width=\linewidth]{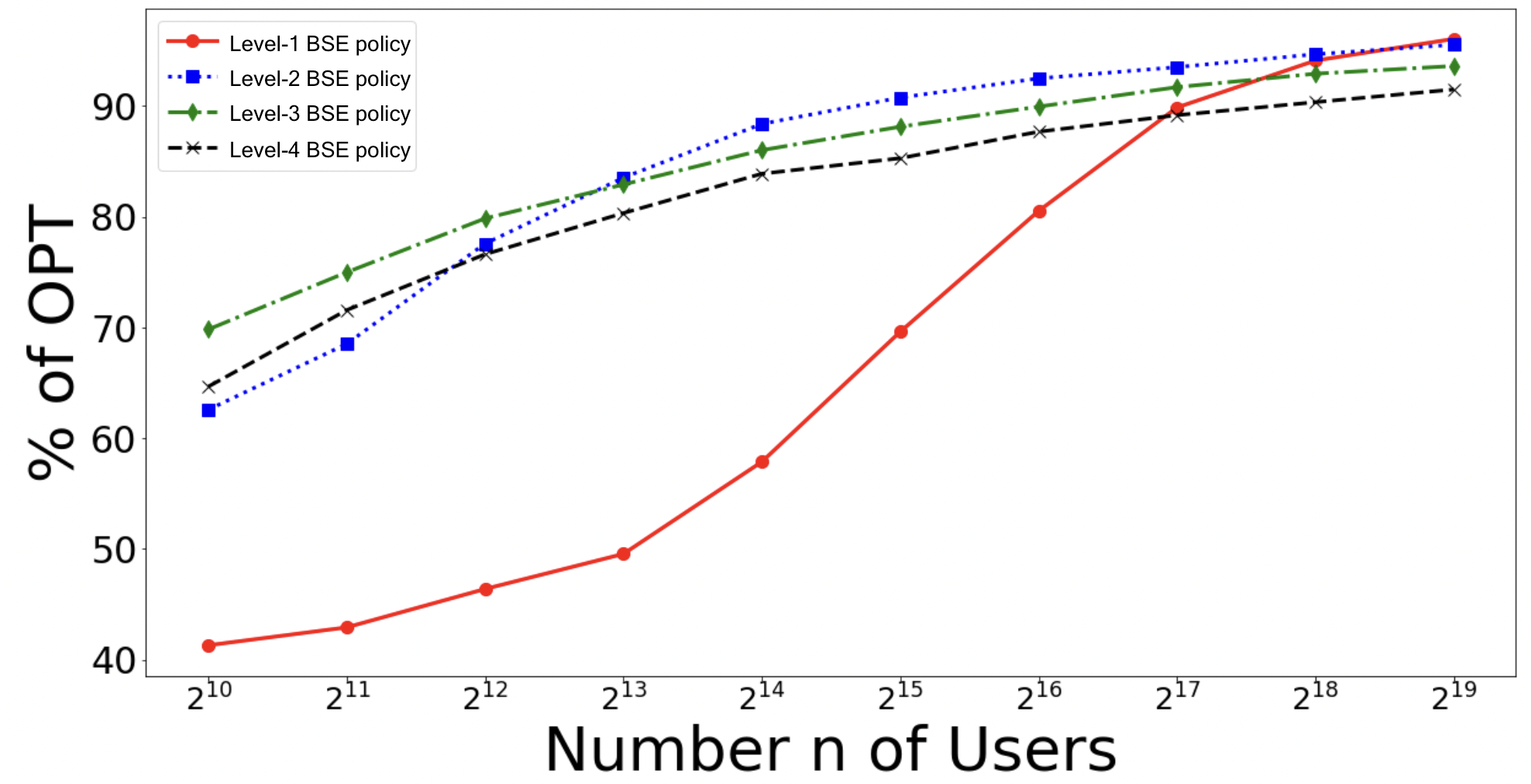}
\captionof{figure}{Performance of Level-$\ell$ BSE-induced Policy}
 \label{fig:semi_synthetic}
\end{minipage}%
\begin{minipage}{.5\textwidth}
 \centering
\includegraphics[width=\linewidth]{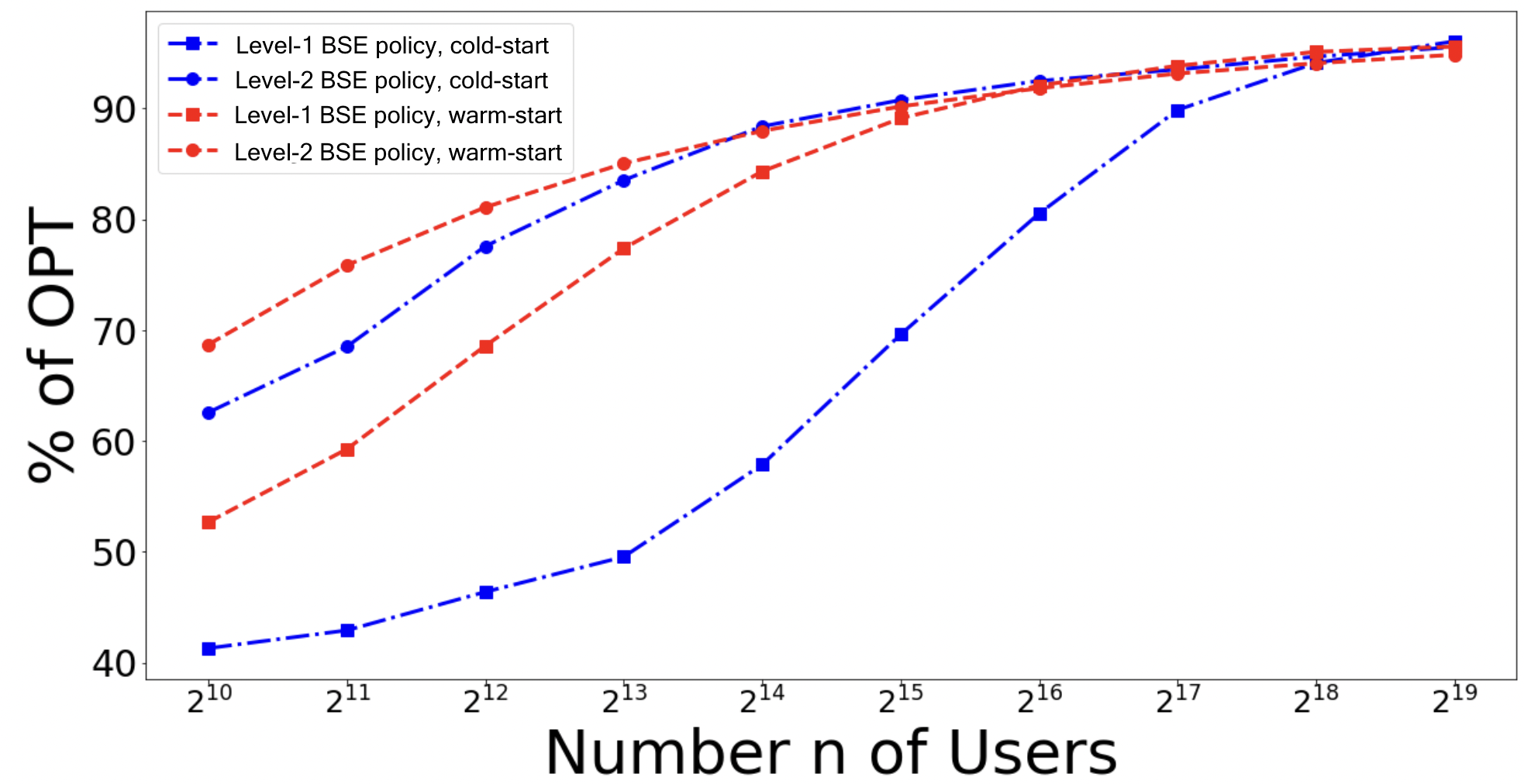}
\captionof{figure}{Cold vs Warm Start}
\label{fig:cold_warm}
\end{minipage}
\end{figure}

\subsection{Cold Start vs. Warm Start}
\label{subsec:warm_start}
This section delves into the impact of prior estimation using the firm's user interaction data. 
In practical scenarios, newly-generated contents are sometimes accompanied by prior information. 
This could involve predicting the expected engagement duration for a particular group of similar users based on factors such as the content's topic and the user's interests.

For long-lived content, previous studies have shown that accurate prior predictions can significantly enhance the performance of bandit algorithms (see, e.g., \citealt{russo2018tutorial}). 
However, we are particularly interested in numerically quantifying the value of prior information in our context, where the content is short-lived and high in volume.

To evaluate the performance of the BSE policies, we conducted offline experiments using the platform's interaction data spanning from May to July 2021. 
For each arm, we assign a Beta prior distribution, whose parameters are determined by applying the method of moments using DNN's predictions, following the same approach as we did in the field experiments.

In Figure \ref{fig:cold_warm}, we compare the performance between the level-$1$ and level-$2$ policies. 
For each $\ell$, the introduction of the prior distribution substantially improved the performance of the level-$\ell$ policy. 
Furthermore, when prior information is available, the level-$1$ BSE policy demonstrates a significant advantage over the level-$2$ policy {\em without} prior information. 

%

\end{document}